\documentclass[lettersize,article,onecolumn]{IEEEtran}
\usepackage{amsmath,amsfonts}
\usepackage{algorithmic}
\usepackage{algorithm}
\usepackage{array}
\usepackage[caption=false,font=normalsize,labelfont=sf,textfont=sf]{subfig}
\usepackage{textcomp}
\usepackage{stfloats}
\usepackage{url}
\usepackage{verbatim}
\usepackage{graphicx}
\usepackage{cite}
\usepackage{graphicx}
\usepackage{amsmath}
\usepackage{amssymb}
\usepackage{float}
\usepackage{xcolor}
\usepackage{tabularx}
\usepackage{booktabs}
\usepackage{bm}
\usepackage[square,numbers]{natbib}
\usepackage{ragged2e}
\usepackage{authblk}
\usepackage{times}
\usepackage{epsfig}
\usepackage{iitem}
\usepackage{dashrule}
\usepackage{float}

\usepackage{amsmath,amsfonts,bm}

\def\eqref#1{equation~\ref{#1}}
\def\1{\bm{1}}

\def\mI{{\bm{I}}}

\def\mO{{\bm{O}}}
\def\mP{{\bm{P}}}

\def\mR{{\bm{R}}}

\DeclareMathAlphabet{\mathsfit}{\encodingdefault}{\sfdefault}{m}{sl}
\SetMathAlphabet{\mathsfit}{bold}{\encodingdefault}{\sfdefault}{bx}{n}

\newcommand{\myparagraph}[1]{\noindent\textbf{#1}}

\begin{document}

\title{Flexible Portrait Image Editing \\ with Fine-Grained Control}

\author{Linlin Liu$^{*}$,
Qian Fu$^{*}$, 
Fei Hou, and Ying He
\thanks{$^{*}$ Equal contribution.}
\thanks{$\bullet$ Linlin Liu is with the Interdisciplinary Graduate School, Nanyang Technological University, Singapore and Alibaba Group. E-mail: linlin001@ntu.edu.sg.}
\thanks{$\bullet$ Qian Fu and Ying He are with the School of Computer Science and Engineering, Nanyang Technological University, Singapore. E-mail: \{qian.fu, yhe\}@ntu.edu.sg.}
\thanks{$\bullet$ Fei Hou is with the Institute of Software, Chinese Academy of Sciences. E-mail: houfei@ios.ac.cn.}}

%\markboth{IEEE TRANSACTIONS ON IMAGE PROCESSING, MARCH~2022}%
%{Liu \MakeLowercase{\textit{et al.}}: Flexible Portrait Image Editing with Fine-Grained Control}


\maketitle

\begin{abstract}
We develop a new method for portrait image editing, which supports fine-grained editing of geometries, colors, lights and shadows using a single neural network model. We adopt a novel asymmetric conditional GAN architecture: the generators take the transformed conditional inputs, such as edge maps, color palette, sliders and masks, that can be directly edited by the user; the discriminators take the conditional inputs in the way that can guide controllable image generation more effectively. Taking color editing as an example, we feed color palettes (which can be edited easily) into the generator, and color maps (which contain positional information of colors) into the discriminator. We also design a region-weighted discriminator so that higher weights are assigned to more important regions, like eyes and skin.
Using a color palette, the user can directly specify the desired colors of hair, skin, eyes, lip and background. Color sliders allow the user to blend colors in an intuitive manner. The user can also edit lights and shadows by modifying the corresponding masks. 
We demonstrate the effectiveness of our method by evaluating it on the CelebAMask-HQ dataset with a wide range of tasks, including geometry/color/shadow/light editing, hand-drawn sketch to image translation, and color transfer. We also present ablation studies to justify our design.
\end{abstract}

%\begin{IEEEkeywords}
%Portrait images, asymmetric conditional GAN, fine-grained control.
%\end{IEEEkeywords}

\section{Introduction}

In the digital age, portrait and self-portrait photographs are extremely popular and have spread to every corner of the world.
Although these images are casual in nature, many users would like to retouch them before sharing online.
The general purpose photo editing tools, such as Adobe Photoshop, provide professional retouching results, however they target only highly skilled users. 

%There are many commercial tools for editing portrait images, spanning from general purpose software to makeover tools. The former, such as The latter provides easy-to-use tools for beautifying one's face, but it does not support fine-grained control.

In research community, the methods for portrait editing can be roughly grouped into two  categories: transfer-based approaches \citep{fu2020poisson,Shu_2017_CVPR,nguyen2021lipstick,Cho_2019_CVPR} and user-guided approaches \citep{xiao2019interactive,zhang2017real}. The transfer-based methods allow the user to transfer colors, lights and shadows from a user-specified reference image to target images. Although these methods can generate visually appealing results, they do not support fine-grained control, since the transfer process is often designed in a fully automated manner. Moreover, it may also be challenging to find suitable reference images to achieve the desired transfer effect.

The user-guided approaches allow the user to control the retouching results by using color strokes \citep{sangkloy2017scribbler}, color and texture patches \citep{xiao2019interactive,Xian_2018_CVPR,zhang2017real}, tags \citep{Kim_2019_ICCV} or even texts \citep{Bahng_2018_ECCV}. The color strokes or patches based methods require moderate user interaction, such as sketching strokes or copying patches to the region of interest. Those methods are intuitive and easy to use, and ideal for the scenarios in which the user only retouches a few images. However, applying them to a large amount of photos is tedious and time consuming. Besides, very sparse user inputs sometimes confuse the neural network that may produce unpredictable colorization results \citep{you2019pi}. The tag/text based methods support batch processing naturally, but their way of control is not intuitive and they cannot support fine tuning. 

\begin{figure*}[t!]
    \centering
    \includegraphics[width=\linewidth]{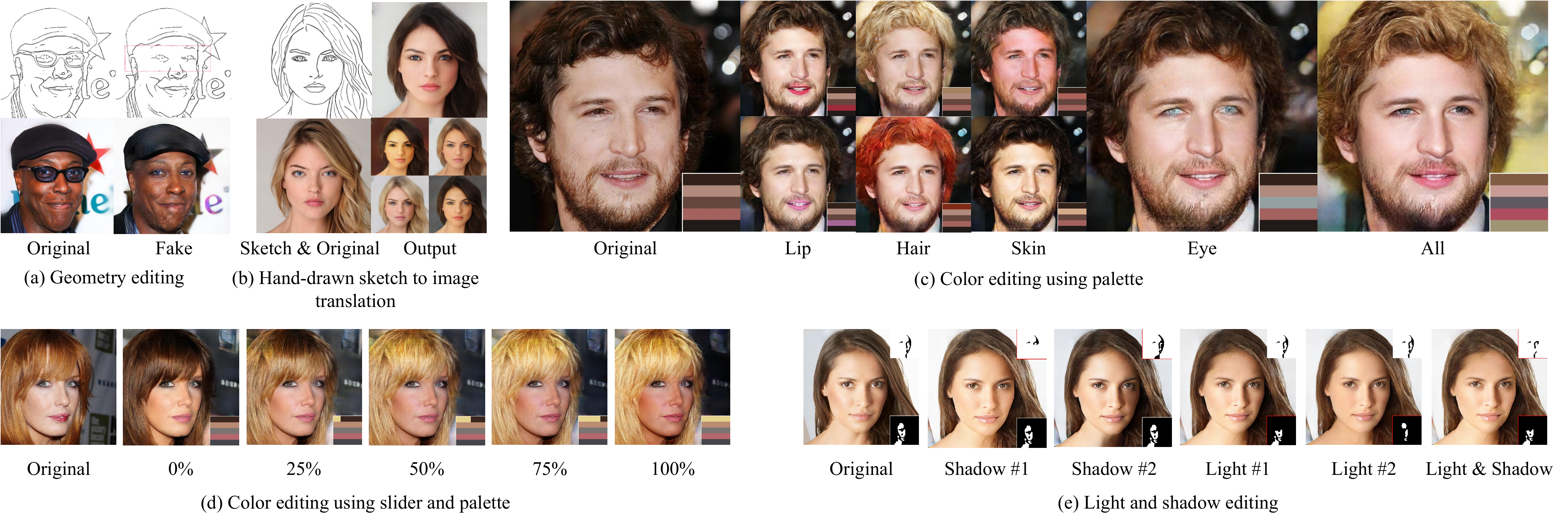}
    \caption{Our method supports fine-grained editing of geometries, colors, lights and shadows using a single neural network model. It allows the user to edit colors via a color palette and/or slider, geometries via edge maps, and lights and shadows via masks. It can also generate images from hand-drawn sketches. The percentages in (d) indicate the yellow ratio in the first row of palette, which controls hair color.}
    \label{fig:teaser}
\end{figure*}

In this paper, we propose an effective framework that adopts conditional GAN \citep{mirza2014conditional,isola2017image} for fine-grained geometry, color, light and shadow editing (see Fig.~\ref{fig:teaser}). Given the training images, We first extract features (e.g., edge maps, color, etc.) as conditional information for controllable image generation. For better visualization to the end users, the features correspond to different editing operations are separated and saved in the form of images. Then we feed the concatenated conditional information into the generator and discriminator of our models, and train the generator to reconstruct the original images. Different from the previous conditional GAN based approaches where the generators and discriminators are usually fed with the same conditional information, we convert generator inputs to the form that are noise robust and easy for hand editing, and convert discriminator inputs to the form that can guide controllable image generation more efficiently. After training, the model can take unseen conditional inputs for image synthesis or editing. Moreover, it is flexible to extend our framework for more editing operations by concatenating additional conditional information with those discussed in this paper.

Our main contributions can be summarized as follows:
\begin{itemize}
    \item We design an all-in-one model for portrait geometry, color, light and shadow editing without compromising performance, which generates photo realistic images and reduces the complexity of deployment and maintenance.
    \item We demonstrate novel and convenient ways for fine-grained portrait image editing, including color editing using palette and slider, and light/shadow editing using masks. These methods are easy to use for both professional and non-professional users.
    \item We propose an asymmetric conditional GAN architecture to improve the performance of fine-grained color editing using palette, and enhance the robustness of geometry editing guided by edge maps.
    \item We design a region-weighted discriminator that enables explicitly assigning higher weight to the more important regions of the generated images, like eyes and face, when computing loss.
    \item Our palette and slider based color editing enables batch image processing, which is not supported in the previous fine-grained editing methods.
\end{itemize}

\section{Related Work}
 
\myparagraph{Conditional GAN.}
Unconditional GANs, proposed by \citet{goodfellow2014generative}, take only a random noise vector as input, thereby is unable to support user control. In contrast, conditional GANs \citep{mirza2014conditional}
train on a labeled data set and allow the user to specify the label to control each generated instance. 
Conditional GANs demonstrate promising performance on various image synthesis tasks \citep{isola2017image,gu2019mask,park2019semantic,shamsolmoali2020image}, where a widely used form for image editing is image-to-image translation \citep{pang2021image}.
However, the existing variants of conditional GANs usually feed the same label/conditional information into generator and discriminator, which is sub-optimal since these two components are designed for different purpose.
Therefore, our proposed asymmetric conditional GAN feeds different conditional information into generator and discriminator to better fit the application needs.

\myparagraph{Portrait Image Editing.}
Portrait image editing has attracted considerable interest due to its wide application, which consists of a long list of tasks, including geometry editing \citep{wu2021coarse}, colorization \citep{zhang2017real}, relighting and shadow manipulation \citep{zhou2019deep,wang2020single,zhang2020portrait}, stylization \citep{li2018closed, yi2019apdrawinggan}, pose and expression control \citep{Tewari_2020_CVPR} and so on. Recently, there are also a few attempts to enable the control of multiple attributes using a single model. For example, MichiGAN \citep{tan2020michigan} is designed for disentangled hair manipulation of structure, shape and so on. However, each attribute in MichiGAN requires a separate condition module, which makes it inflexible to extend. In this work, we conduct extensive experiments to show that our model can be extended for a wide range of tasks by simply concatenating more conditional information to the inputs.

\myparagraph{Colorization and Color Editing.} The existing image colorization and color editing methods can be categorized into two groups, user-guided approach and transfer-based approach. The user guided approaches usually control the results using sparse user inputs, like strokes \citep{levin2004colorization,sangkloy2017scribbler}, points/patches \citep{zhang2017real}, and so on. These methods often do not support batch editing. The transfer-based approach allows automatic color transfer from the reference images \citep{reinhard2001color,he2018deep,afifi2021histogan}, but do not allow fine-grained control. Similar to our approach, the transfer-based image recoloring method proposed by \citet{chang2015palette} also takes palette as input. However, they do not incorporate spatial information into the algorithm, so their method is inflexible for fine-grained editing.

\myparagraph{Light and Shadow Editing.} There is also a long line of research on portrait image relighting or light/shadow editing. Most of the recent approaches \citep{zhang2020portrait,NestmeyerLML20,SunLBXR21,HouZSBT021} use target light for control. For example, \citet{NestmeyerLML20} use physics-based relight, but find noise increases when there is little light. \citet{ShuHSSPS18} also propose method to transfer light from reference image, however, they notice the method fails when adding/removing no-diffuse effects. We adopt the mask-based method, which is more controllable during fine-grained light and shadow editing.

\section{Method}

\subsection{Overview}
We first extract portrait edge maps, colors, and light/shadow masks from the training images. Then the conditional GAN is trained to reconstruct the original images with the extracted constraint features as input. After training, we can modify the model inputs to achieve fine-grained portrait image editing. To ease reading, we list the main notations in Table~\ref{tab:notations}.

\begin{table}[t]
    \centering
    \caption{Summary of the main notations.}
    \begin{tabular}{cl}
    \toprule
    Notation & Description \\
    \midrule
    $\mI_{E}$ & edge map\\
$\mI_{CP}$ & color palette\\
        $\mI_L$ &light mask\\
        $\mI_S$ & shadow mask\\
        $\mI_E$ & original edge map\\
        $\mI_C$ & color map\\
                   $n(\mI_{E})$ & noisy edge map \\
                   \hline

        $G$ & the generator\\
        $\mI_G$ & the input of $G$\\
                        $\mP_F$ & the output of $G$\\
\hline
        $m(\mP_F)$  &  applying the skin and eye masks to $\mP_F$\\        
        $D_L$ & the multi-scale discriminator for local facial components \\
        $\mI_{D-L}$ & the input of $D_L$\\
        $D_G$ & the multi-scale discriminator for global appearance\\
                $\mI_{D-G}$ & the input of $D_G$\\
$\mathcal{L}_{MD-L}$  & the loss function for $D_L$\\
        $\mathcal{L}_{MD-G}$ & the loss function for $D_G$\\
    \bottomrule
    \end{tabular}
    \label{tab:notations}
\end{table}

\subsection{Model Inputs}
\label{sec:model_input_sec}
\subsubsection{Edge Maps}
\label{sec:edge_map_extract}
Edge maps, which provide  salient structural information for images, are often used to control the geometry of the generated images. The DexiNed model~\citep{poma2020dense} shows encouraging edge detection performance on outdoor images. To adapt DexiNed for portrait images, we annotate 27 portrait images manually and then combine them with the Barcelona Images for Perceptual Edge Detection (BIPED) dataset \citep{poma2020dense} that contains 250 hand-annotated outdoor images for model training. Given an input image, the trained model outputs a probability matrix of the same dimension as the input, where each value in the matrix represents the likelihood of the corresponding pixel to be part of an edge. Since the generated edge maps are noisy, we apply the following steps to further reduce noise: 1) apply the Gaussian filter to the input edge map and then set the pixels below a pre-defined threshold $\beta$ to 0 ($\beta=0.35$ in our implementation); 2) move a $5 \times 5$ sliding window through the edge map with stride size 1, and filter out the bottom 20\% values in each window; 3) apply the Gaussian filter in step 1) to generate the final edge map. Comparing with the classic edge detection algorithms \citep{canny1986computational,xie2015holistically,yi2019apdrawinggan}, we observe the portrait-tailored DexiNed model produces visually clean results that facilitate editing.

\subsubsection{Color Palette}
\label{sec:color_palette}

\begin{figure}[t!]
\centering
{\footnotesize
\includegraphics[width=0.19\columnwidth]{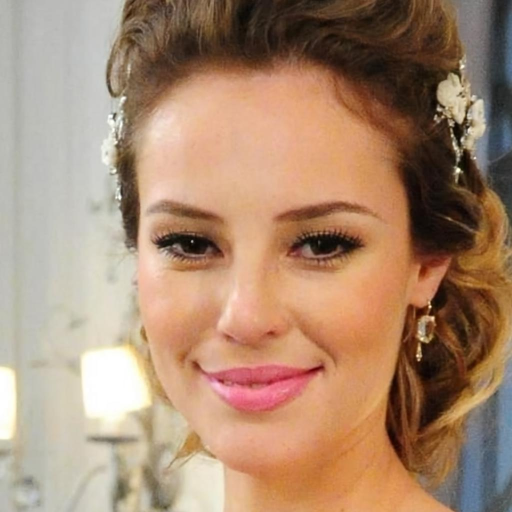}
\includegraphics[width=0.19\columnwidth]{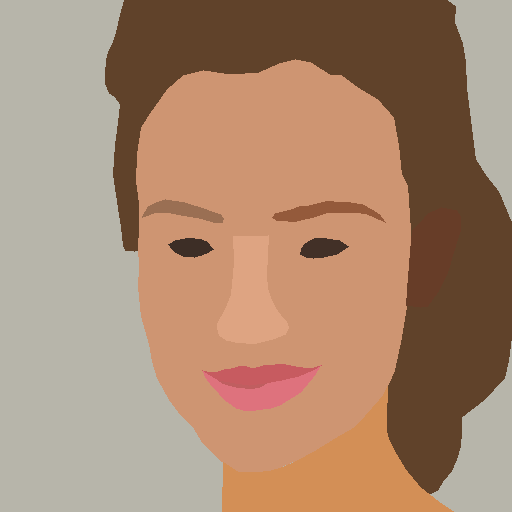}
\includegraphics[width=0.19\columnwidth]{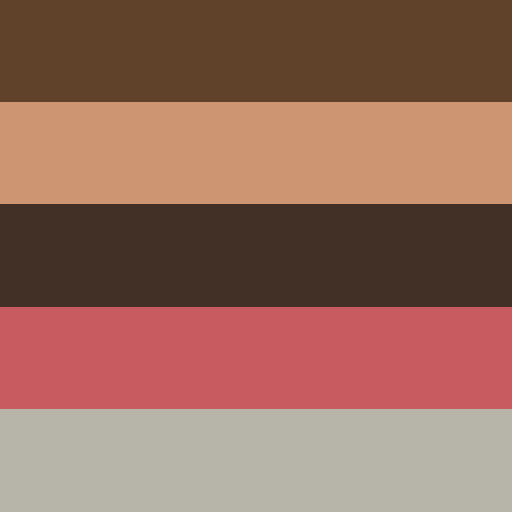}\\
\makebox[0.19\columnwidth]{(a)}
\makebox[0.19\columnwidth]{(b)}
\makebox[0.19\columnwidth]{(c)}\\
}
\caption{Color palette extraction. Given the input image (a), we compute a semantic segmentation, consisting of hair, face skin, eyebrows, eyes, nose, mouth, neck and background. Then the average color of each segment is computed to generate the color map (b). Finally, (b) is transformed to a color palette (c), by taking the average colors of hair, face, eye, lip and background (from top to bottom).
}
\label{fig:color_palette}
\end{figure}

Fine-grained color editing is often a tedious work and usually requires expertise. To help get rid of the tedious work, we design our framework to support automated color editing without compromise of fine grained control. Our framework achieves this objective with the help of color palettes. 

We use the CelebAMask-HQ dataset \citep{DBLPLee0W020} for model training, which contains high-resolution portrait images and manually annotated facial segmentation masks. If there is any additional portrait images from the other sources, the pretrained face parsing model released by \citet{DBLPLee0W020} can also be used for annotation. Then we compute the average RGB pixel value of each facial component as the corresponding color map. An example is shown in Fig.\ref{fig:color_palette}(b). However, editing image with color map still requires users to decide the boundary of each facial component. To further simply the color editing work, we take colors of the 5 most important components from the color map, namely average colors of hair, face, eye, lip and background, to create a color palette for each image. An example is shown in Fig.\ref{fig:color_palette}(c). Order of these components in the color palettes are fixed, if any component is not found we use black as the default color. It is possible to add more facial components for finer grained color control. However, exposing non-professional users to too many choices may increase the editing complexity, so we let the model to help decide colors of the other components based on the context.

\subsubsection{Light and Shadow Masks}
\begin{figure}[t!]
\centering
{\footnotesize
\includegraphics[width=0.19\columnwidth]{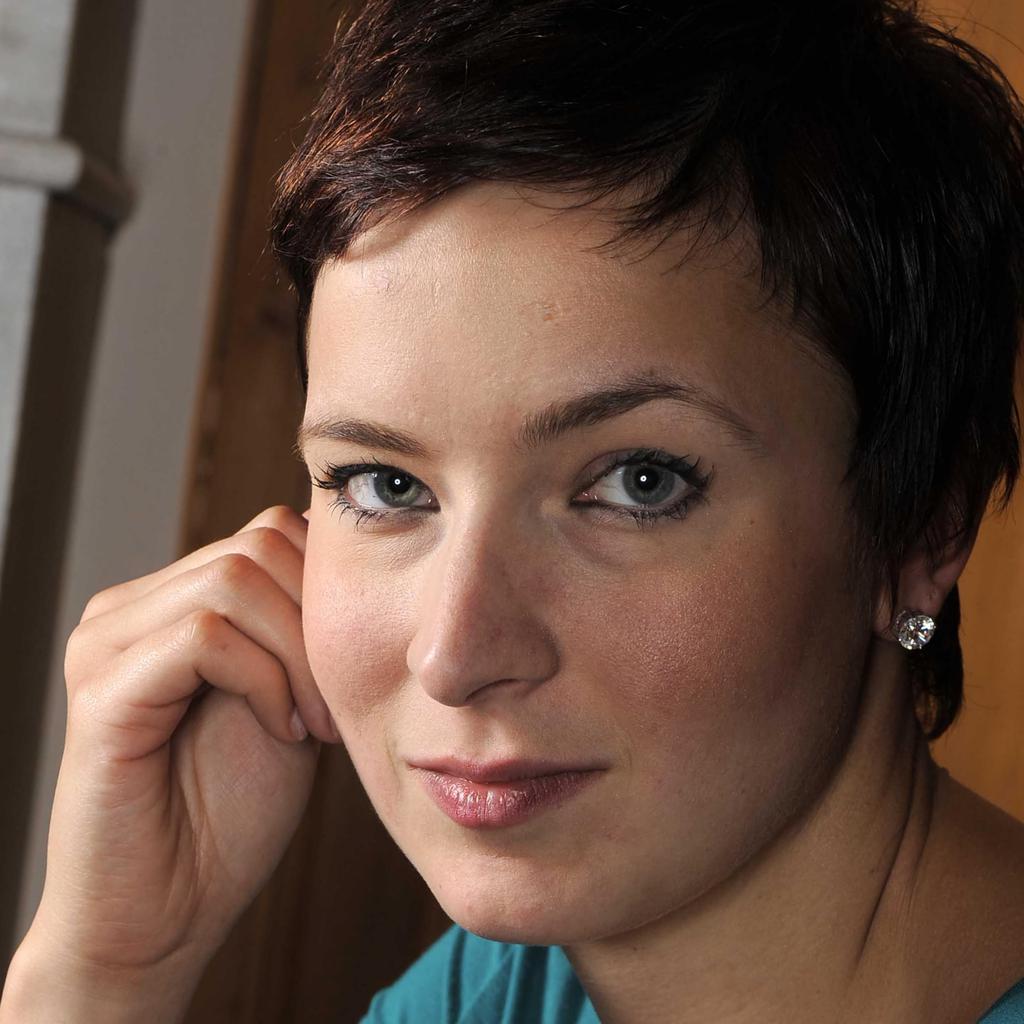}
\includegraphics[width=0.19\columnwidth]{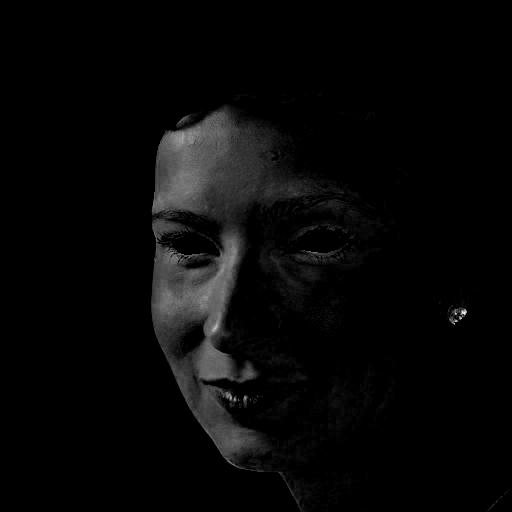}
\includegraphics[width=0.19\columnwidth]{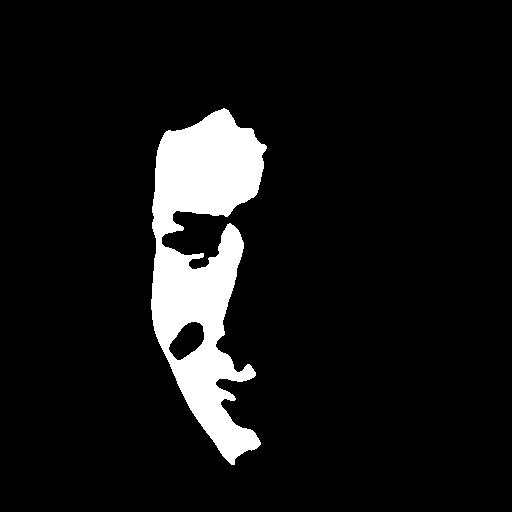}
\includegraphics[width=0.19\columnwidth]{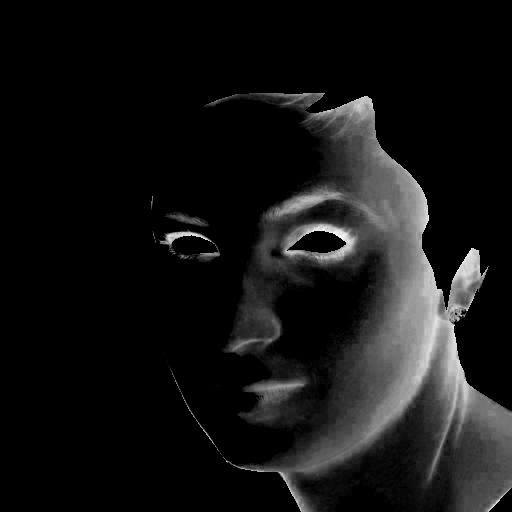}
\includegraphics[width=0.19\columnwidth]{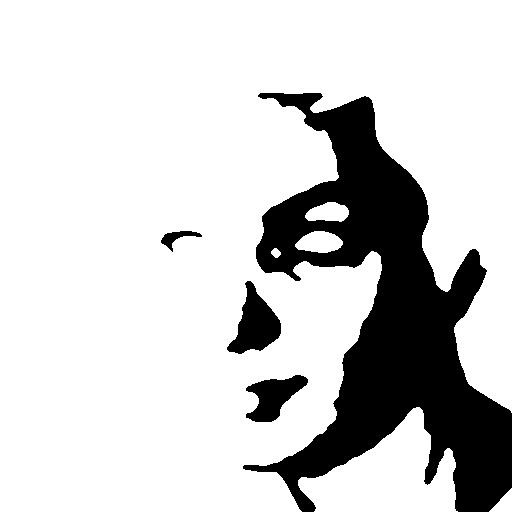}\\
\makebox[0.19\columnwidth]{Original}
\makebox[0.19\columnwidth]{Light}
\makebox[0.19\columnwidth]{Light Mask}
\makebox[0.19\columnwidth]{Shadow}
\makebox[0.19\columnwidth]{Shadow Mask}\\
}
\caption{Light and shadow mask extraction.}
\label{fig:light_removal}
\end{figure}

We apply a simple yet effective light extraction algorithm \citep{shen2013real} to extract facial light from the input images, and facial shadow extraction can be achieved by applying the same algorithm on the inverted image.
Then the extracted light and shadow are binarized with the threshold $0.15$ after being normalized to the range of $[0, 1]$.
Finally, the binary images are processed using a 2D median filter with window size $7\times7$ to generate the light and shadow masks.
Fig.~\ref{fig:light_removal} shows an example of light and shadow mask extraction.

\subsection{Edge Map Noising Methods}
We apply three noising methods, namely random removal, random shift and random lines, to the edge maps to improve model robustness. This allows even beginners to edit extracted edge maps and obtain high quality results. We use $\mathcal{N} = \{n_{RR}(\cdot), n_{RS}(\cdot), n_{RL}(\cdot), n_{N}(\cdot)\}$ to denote the set of noising functions, where $n_{RR}(\cdot), n_{RS}(\cdot), n_{RL}(\cdot)$ denotes random remove, random shift and random line, respectively. And $n_{N}(\cdot)$ denotes \textbf{not} to apply any of the noising functions, i.e., simply returning the original input. Given an edge map $\mI_E$, we apply a random noising function $n(\cdot)$ uniformly selected from $\mathcal{N}$ to generate the noisy edge map $n(\mI_E)$, then $n(\mI_E)$ is used as part of the inputs during model training. During test, we do not apply any noising function to $\mI_E$, that is, we directly feed $\mI_E$ into the model for inference.

\begin{figure}[t!]
\centering
{\footnotesize
\includegraphics[width=0.19\columnwidth]{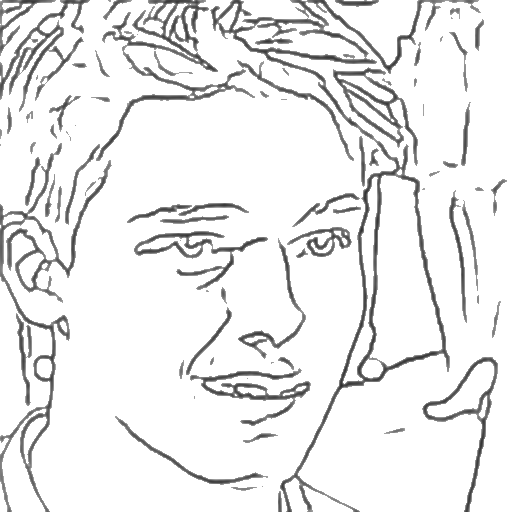}
\includegraphics[width=0.19\columnwidth]{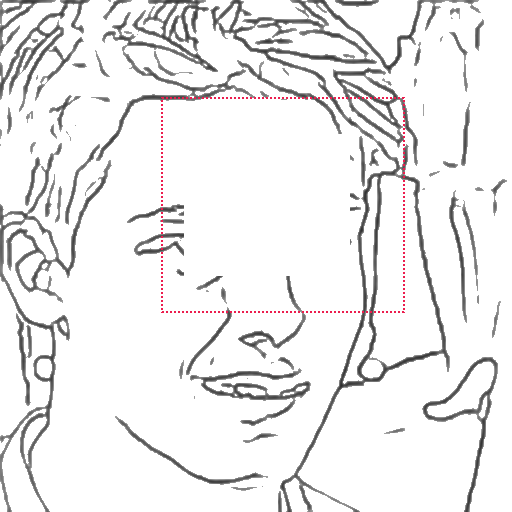}
\includegraphics[width=0.19\columnwidth]{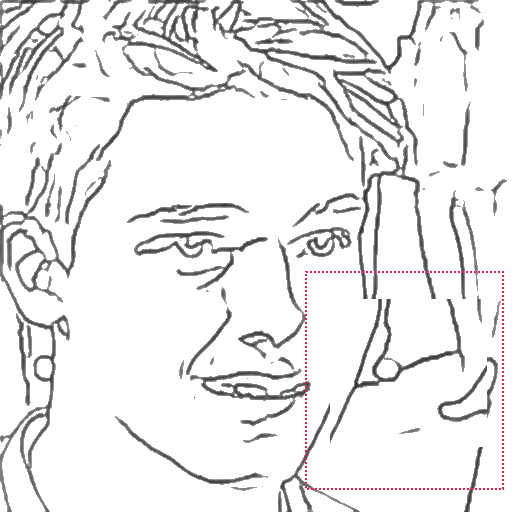}
\includegraphics[width=0.19\columnwidth]{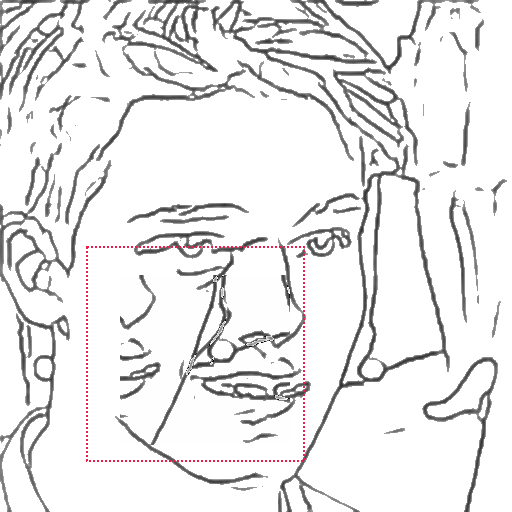}\\
\makebox[0.19\columnwidth]{(a)}
\makebox[0.19\columnwidth]{(b)}
\makebox[0.19\columnwidth]{(c)}
\makebox[0.19\columnwidth]{(d)}\\
}
\caption{Examples of the edge map noising methods. (a) Original edge map; (b) Random removal; (c) Random shift; (d) Random lines.}
\label{fig:edge_noise}
\end{figure}

%\label{sec:noising_random_remove}
\ \\
\noindent\textbf{Random removal.} We remove a randomly selected region from the input in this noising method, which is similar to the masks used in \citep{yang2020deep}, except that we apply it to the edge maps instead of the original images. Let $h$ and $w$ be the height and width of the input image. We select a random rectangle $\mR$ of height $h^\prime$ and width $w^\prime$ from the edge map to remove, where $h^\prime \in (0, \alpha \times h)$ and $w^\prime \in (0, \alpha \times w)$ are randomly selected from the intervals, and $0 < \alpha < 1$ is a hyper-parameter determines the max dimensions of $\mR$. We set $\alpha=0.33$ in our implementation. See Fig.~\ref{fig:edge_noise}(b) for an example of random removal. This noising method improves the robustness of the model so that it can deal with missing lines in edge maps.

%\subsubsection{Random Shift}
%\label{sec:noising_random_shift}
\ \\
\noindent\textbf{Random shift.} We first select a random rectangle $\mR$ of height $h^\prime$ and width $w^\prime$ from the edge map following the same steps in random removal, and then shift the lines in $\mR$ by a random displacement. Fig.~\ref{fig:edge_noise}(c) shows an output generated by this noising method, where the lines on face are shifted towards the right hand side. The introduced noises during training helps model to learn to generate more coherent images when the edge map contains inaccurate lines drawn by user.

\ \\ 
%\subsubsection{Random Lines}
\noindent\textbf{Random lines.} Adding random lines to edge map can be achieved via similar operations as random shift. Assume $\mR$ is the random rectangle selected from the original edge map, and $\mR^{\prime}$ is the target area that lines in $\mR$ will be copied to, where $\mR^{\prime}$ is of the same dimension as $\mR$. In random shift, the lines in $\mR^{\prime}$ are overwritten by white color, which looks like being shifted from $\mR$ to $\mR^{\prime}$. However, the random lines function, Fig.~\ref{fig:edge_noise}(d) for example, keeps the original lines in the target area $\mR^{\prime}$, which looks adding extra lines to the edge map. This function is designed to improve model's robustness to extra noisy lines caused by edge map extraction or hand editing.

\subsection{Asymmetric Conditional GAN}
\begin{figure*}[t!]
\centering
{\footnotesize
\includegraphics[width=1.0\linewidth]{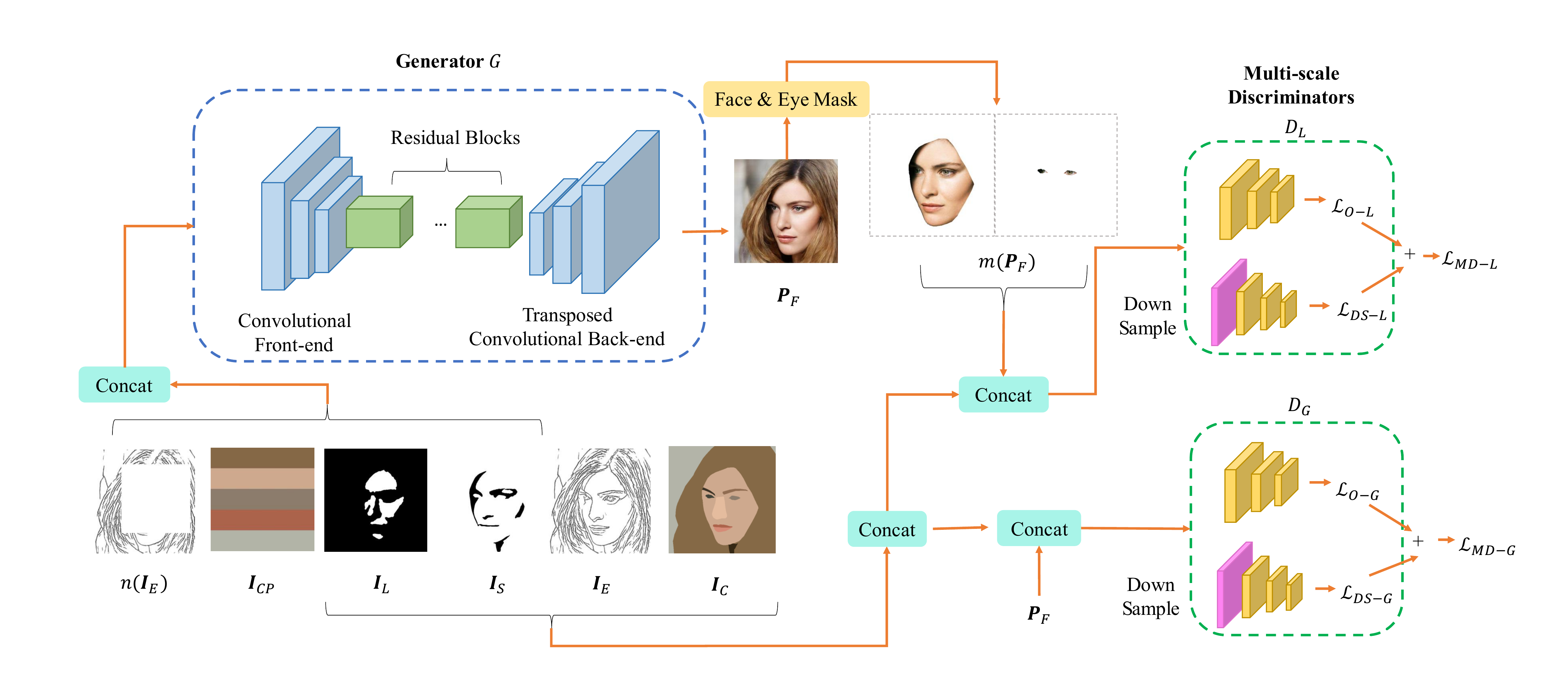}
}
\vspace{-1em}
\caption{Model architecture. In contrast to the existing conditional GAN models, we feed the generator and the discriminators relevant but different types of conditional information. The input to the generator $\mI_G = \mathrm{concat}( n(\mI_{E}), \mI_{CP}, \mI_{L}, \mI_{S})$, consisting of noisy edge map $n(\mI_E)$, color palette $\mI_{CP}$, light mask $\mI_L$ and shadow mask $\mI_S$, can be directly edited by the user. There are two multi-scale discriminators $D_G$ and $D_L$. $D_G$, which takes $\mI_{D-G} = \mathrm{concat}( \mI_{E}, \mI_{C}, \mI_{L}, \mI_{S}, \mP)$ as input, judges global visual appearance. $D_L$, on the other hand, takes $\mI_{D-L} = \mathrm{concat}( \mI_{E}, \mI_{C}, \mI_{L}, \mI_{S}, m(\mP))$ as input and provides feedback of local facial components to $G$.}
\label{fig:model_arch}
\end{figure*}

We build our model upon pix2pixHD \citep{wang2018high}, a conditional GAN based model designed for high-resolution photorealistic image-to-image translation.
%The pix2pixHD model can synthesize portraits from face label maps. 
In the previous conditional GAN based models, both the generators and the discriminators take the same conditional information as input during model training.
To facilitate fine-grained control, we propose a novel asymmetric conditional GAN, where the generator and the discriminators take relevant but different conditional information as input. This feature distinguishes our method and the other conditional GAN based image synthesis methods~\citep{dash2021high,wang2018high,chen2018sketchygan}.
We show the network architecture in  Fig.~\ref{fig:model_arch}. 

\subsubsection{Generator}
The original pix2pixHD model generator, composed of a global generator network and an additional local enhancer network, is able to produce images of resolution $2048 \times 1024$ or even higher. It is observed that even without the local enhancer network the quality of generated images of resolution $1024 \times 512$ or lower are still promising, so we only keep the global generator network in our implementation to reduce computation resource usage. To generate higher resolution images, one could add the local enhancer network back.

The generator $G$ consists of three main components, a convolutional front-end, 9 residual blocks, and a transposed convolutional back-end. We concatenate noisy edges $n(\mI_{E})$, color palette $\mI_{CP}$, light mask $\mI_{L}$, shadow mask $\mI_{S}$ as conditional information to generate the output image. The concatenated input $$\mI_G = \mathrm{concat}( n(\mI_{E}), \mI_{CP}, \mI_{L}, \mI_{S})$$ is passed through the three generator components sequentially, where they are down-sampled in the convolutional front-end and up-sampled in the transposed convolutional back-end. Finally, the generator yields a fake image $\mP_F = G(\mI_G)$. 

\subsubsection{Discriminators}
\label{sec:discriminators}
We design multiple discriminators to provide feedback to the generator $G$ based on more explicit conditions. In the existing conditional GAN models~\citep{dash2021high,wang2018high,chen2018sketchygan}, the generators and the discriminators are fed with the same conditional input. However, such design either constrains the convenience of the control of the generators, or hinders the discriminators from providing better feedback for the generator outputs. Thus, we propose an \textbf{asymmetric conditional GAN}, that allows generators and discriminators to use conditional information in \textit{different} forms, which are more relevant to their objective.

Different from the original multi-scale discriminators used \citet{wang2018high}, we also propose a \textbf{region-weighted discriminator}, which consists of two multi-scale discriminators, namely $D_G$ for the global appearance and $D_L$ for local facial components. As shown in Fig.~\ref{fig:model_arch}, the architectures of these two multi-scale discriminators are similar, but they take different inputs and their parameters are not shared. 
Compared with hair and background, humans are usually more sensitive to the quality of generated faces and eyes, so it is intuitive to give their corresponding regions higher weights when computing loss. Therefore, we use semantic segmentation model to extract face and eye masks from the real portrait images $\mP_R$, and apply masks to $\mP$ to get $m(\mP)$, where $\mP$ denotes $\mP_R$ or $\mP_F$, and $m(\mP)$ denotes the concatenation of face and eye regions extracted from $\mP$.

We feed 
$$\mI_{D-L} = \mathrm{concat}( \mI_{E}, \mI_{C}, \mI_{L}, \mI_{S}, m(\mP))$$
into this region-weighted discriminator to compute loss $\mathcal{L}_{MD-L} = D(\mI_{D-L})$. 

To ensure the coherence of the whole image, we train another multi-scale discriminator with loss function $\mathcal{L}_{MD-G} = D_{L}(\mI_{D-G})$, where 
$$\mI_{D-G} = \mathrm{concat}( \mI_{E}, \mI_{C}, \mI_{L}, \mI_{S}, \mP).$$
We can observe the difference between the generator and discriminator inputs. For geometry control, $n(\mI_{E})$ is used in the generator inputs to simulate the noisy edges, while $\mI_{E}$ is used in the discriminators' inputs, which allows the discriminators to find the possible regions caused by noises easily and thus provide better instructions to the generator on how to handle the noises. Similarly, for color control we use $\mI_{CP}$ in the generator inputs and $\mI_{C}$ in the discriminators' inputs. $\mI_{CP}$ is easier for users to edit, and $\mI_{C}$ is easier for discriminators to learn since it also contains positional information of the color pixels.

Each multi-scale discriminator is also composed of two different discriminator networks. Both of the two discriminator networks take the same input to predict a fake/real label. Then the losses are computed based on their predictions. In multi-scale discriminator $D_{G}$, the first discriminator network passes inputs through several convolutional layers to compute the loss $\mathcal{L}_{O-G}$. The second discriminator down-samples the inputs with an average pool layer first, and then processes the down-sampled inputs with a similar network as the first one to compute the loss $\mathcal{L}_{DS-G}$. The final loss for $D_G$ is computed with $$\mathcal{L}_{MD-G} = \mathcal{L}_{O-G} + \mathcal{L}_{DS-G}.$$
Note that the parameters of the two networks are not shared. Loss $\mathcal{L}_{MD-L}$ is computed in a similar way for $D_{L}$.

\subsubsection{Loss Functions}
As described above, we use $G$ to generate the fake portrait image $\mP_F = G(\mI_G)$. The other conditional information can be concatenated with $\mP_R$ or $\mP_F$ to get the inputs for the two discriminators, $\mI_{D-G}$ for $D_G$ and $\mI_{D-L}$ for $D_L$. In order to differentiate the inputs when the different images $\mP=\mP_R$ and $\mP=\mP_F$ are used, we add additional subscripts $R$ and $F$ to $\mI_{D-G}$ and $\mI_{D-L}$
$$
\mathcal{L}_{MD-G} = \mathbb{E}\left[\log D_G(\mI_{D,R-G})\right] + \mathbb{E}\left[\log (1- D_G(\mI_{D,F-G}))\right],
$$
and 
$$
\mathcal{L}_{MD-L} = \mathbb{E}\left[\log D_L(\mI_{D,R-L})\right] + \mathbb{E}\left[\log (1- D_L(\mI_{D,F-L}))\right].
$$

We define the GAN loss as 
\begin{equation}
    \min_{G} \max_{D_G,D_L} \mathcal{L}_{GAN} = \min_{G} \max_{D_G,D_L} \left(\mathcal{L}_{MD-G} + \mathcal{L}_{MD-L}\right).
    \label{eq:loss_gan}
    \end{equation}
    
In addition to $\mathcal{L}_{GAN}$, we also compute the perceptual loss \citep{johnson2016perceptual} and the feature matching loss \citep{wang2018high} to further improve the quality of the generated images. These two loss functions have proven effective in many image synthesis tasks~\citep{yang2020deep,pang2018visual,dosovitskiy2016generating}. To leverage the knowledge learned by the pretrained models, we feed real images $\mP_R$ and and fake images $\mP_F$ into the pretrained VGGNet \citep{simonyan2014very} to compute the perceptual loss
\begin{gather}
    \mathcal{L}_{VGG} = \mathbb{E}\left[ \sum_{i,j} \lambda_i\left\|\Phi_{i,j}(\mP_R) - \Phi_{i,j}(\mP_F)\right\|_1\right],
    \label{eq:loss_perc}
\end{gather}
where $\Phi_{i}(\cdot)$ is the feature map at the $i$-th VGGNet layer, $\Phi_{i,j}(\cdot)$ is the $j$-th element in the feature map, and $\lambda_{i}$ is the weight for the $i$-th layer. We compute the feature matching loss in a similar way, except that we compute the distance between the feature maps extracted from $D_G$
\begin{gather}
    \mathcal{L}_{FM} = \mathbb{E}\left[\sum_{n,i,j} \lambda^{\prime}_i\left\|\Phi^{\prime}_{n,i,j}(\mI^{\prime}_{D,R}) - \Phi^{\prime}_{n,i,j}(\mI^{\prime}_{D,F})\right\|_1\right],
    \label{eq:loss_fm}
\end{gather}
where $\Phi^{\prime}_{n,j}(\cdot)$ is the feature map extracted from the $i$-th layer of the $n$-th discriminator of $D_G$, since $D_G$ contains two discriminator networks. $\lambda^{\prime}_i$ is the weight of the $i$-th layer. So in addition to Eq.~\ref{eq:loss_gan}, we also adjust $G$ parameters with
\begin{gather}
    \min_{G} (\mathcal{L}_{VGG} + \mathcal{L}_{FM}).
\end{gather}

\section{Experiments}
In this section, we first describe the implementation details of our model. Then, we present the experimental results of various fine-grained editing operations supported by our model. Finally, we perform ablation studies to quantitatively verify the effectiveness of the proposed model.

\subsection{Implementation Details}

\textbf{Dataset.} We use the CelebAMask-HQ dataset \citep{DBLPLee0W020} in our experiments, which contains 30,000 high-resolution portrait images. Each image has a manually-annotated segmentation mask of facial attributes in 19 pre-defined classes. We rescale the images to 512 $\times$ 512, and use 29,490 randomly selected images for model training, and the remaining 510 images for testing.

\ \\ 
\noindent\textbf{Model training.} We use the Adam optimizer \citep{KingmaB14} to train our models for 60 epochs with batch size 64. We set a constant learning rate 0.0002 for the first 30 epochs, and then linearly decrease it to zero for the next 30 epochs. We use VGG19 \citep{simonyan2014very} when computing the perceptual loss.

\subsection{Fine-Grained Editing}

\subsubsection{Geometry}

\begin{figure}[t!]
\centering
{\footnotesize
\includegraphics[width=0.16\columnwidth]{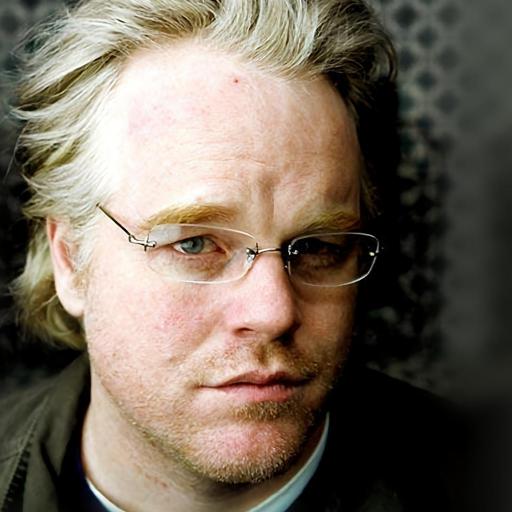}
\includegraphics[width=0.16\columnwidth]{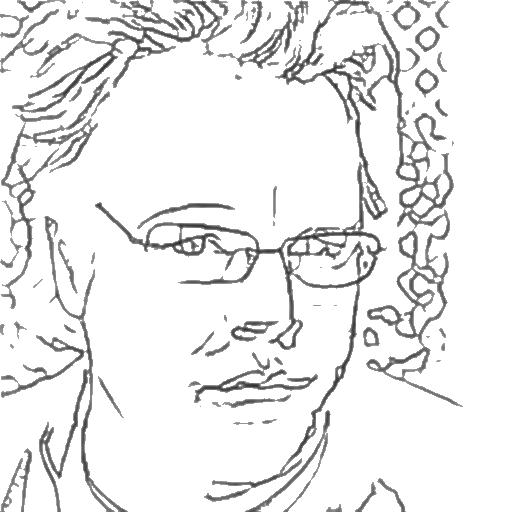}
\includegraphics[width=0.16\columnwidth]{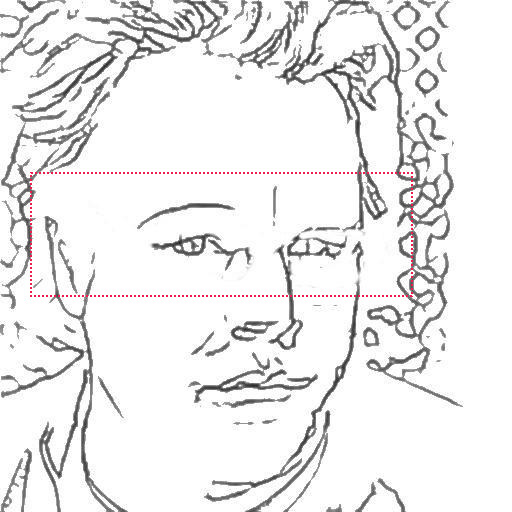}
\includegraphics[width=0.16\columnwidth]{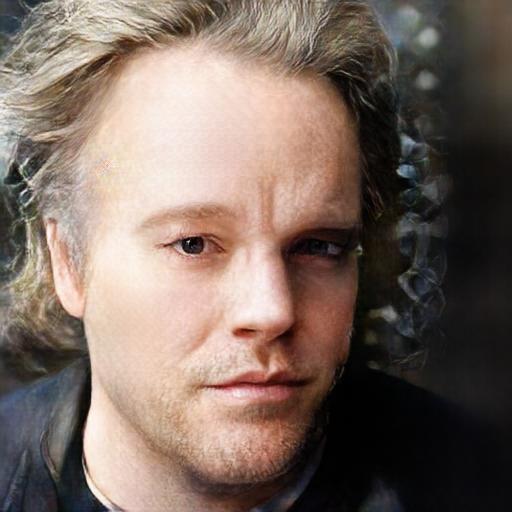} \\
\includegraphics[width=0.16\columnwidth]{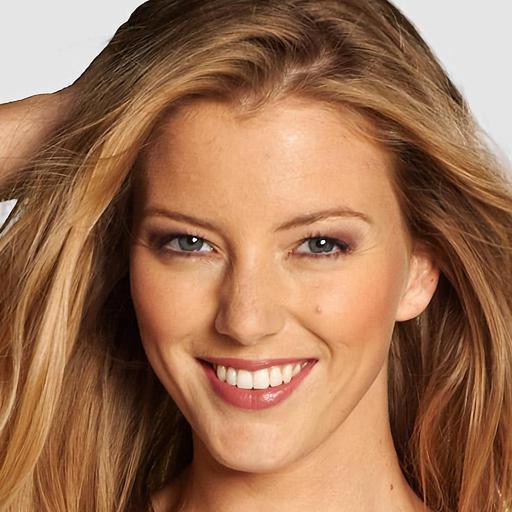}
\includegraphics[width=0.16\columnwidth]{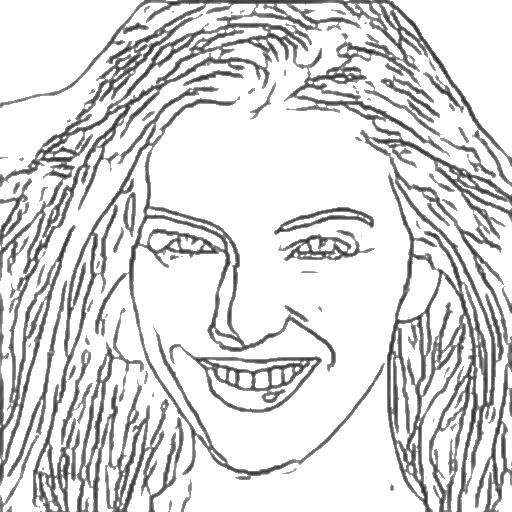}
\includegraphics[width=0.16\columnwidth]{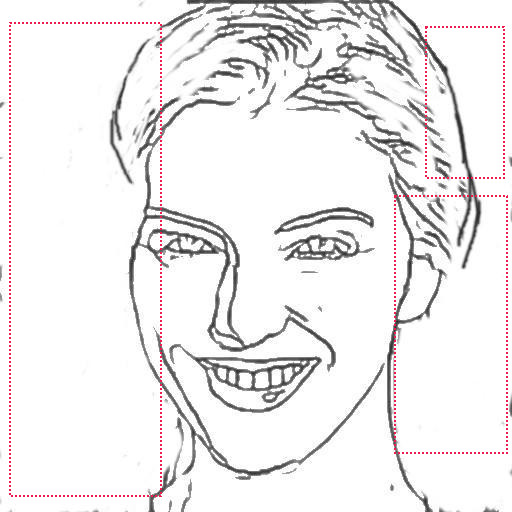}
\includegraphics[width=0.16\columnwidth]{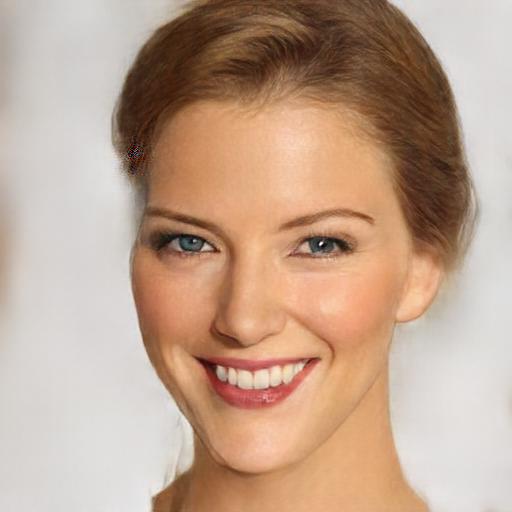} \\
\includegraphics[width=0.16\columnwidth]{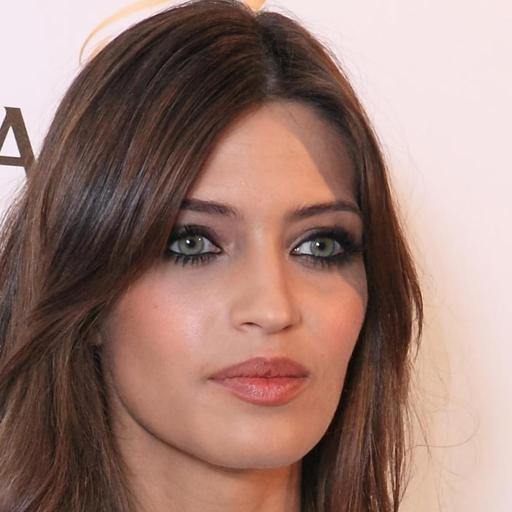}
\includegraphics[width=0.16\columnwidth]{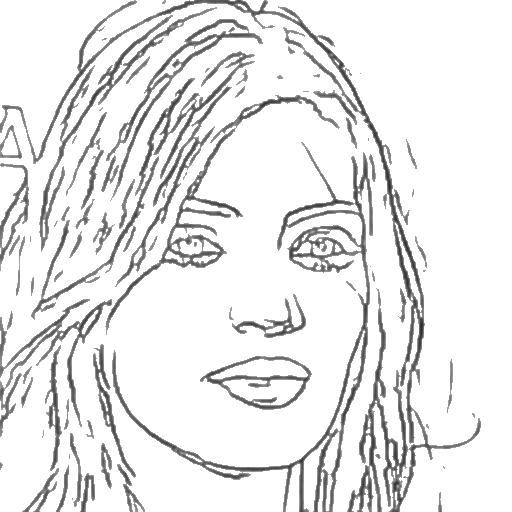}
\includegraphics[width=0.16\columnwidth]{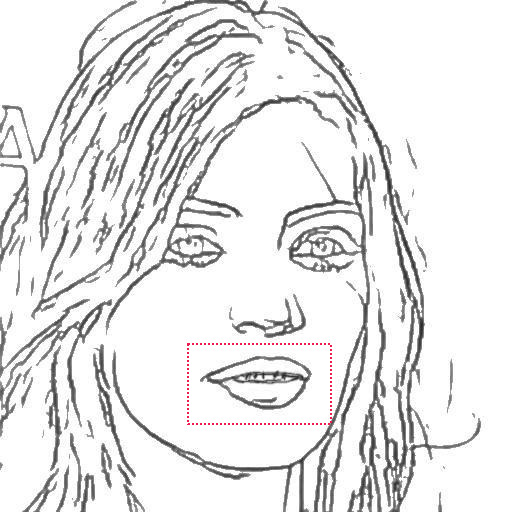}
\includegraphics[width=0.16\columnwidth]{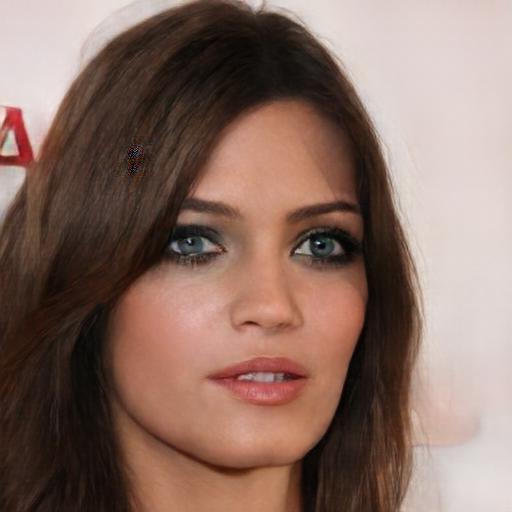} \\
\includegraphics[width=0.16\columnwidth]{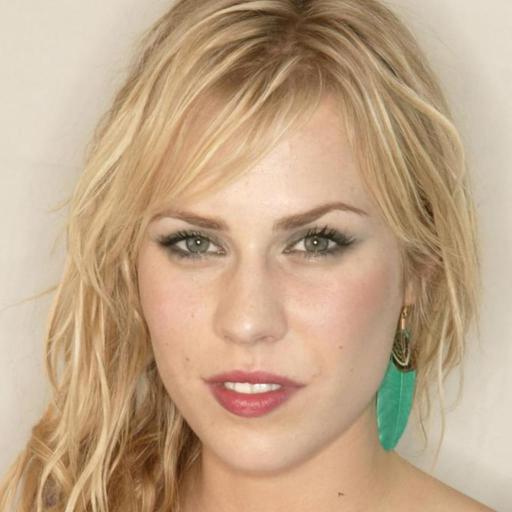}
\includegraphics[width=0.16\columnwidth]{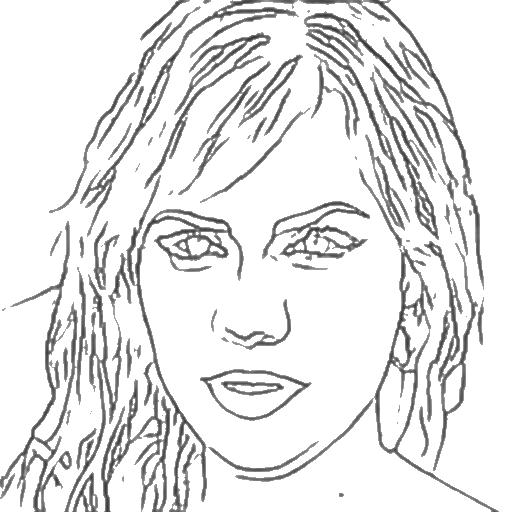}
\includegraphics[width=0.16\columnwidth]{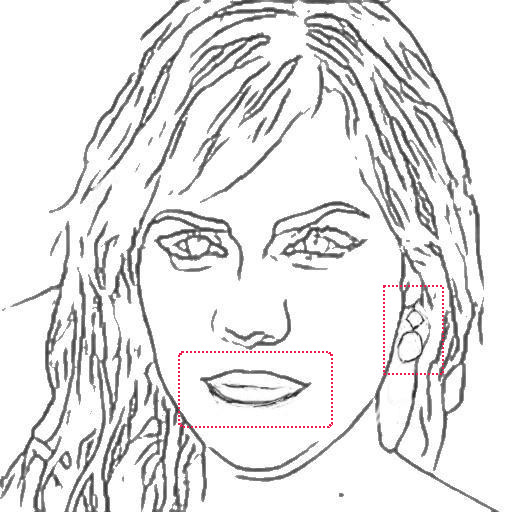}
\includegraphics[width=0.16\columnwidth]{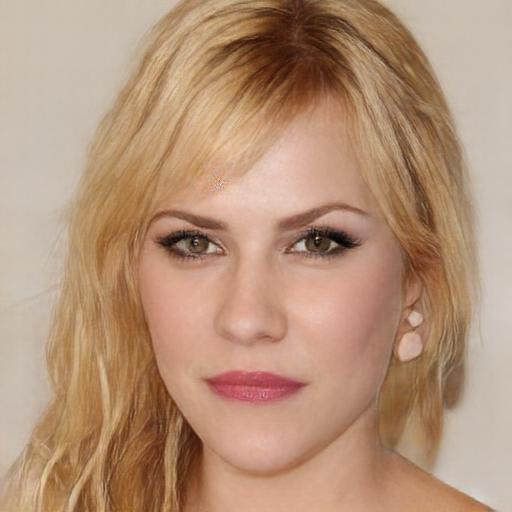} \\
\makebox[0.16\columnwidth]{Original}
\makebox[0.16\columnwidth]{Original edges}
\makebox[0.16\columnwidth]{Edited edges}
\makebox[0.16\columnwidth]{Generated images}\\
}
\caption{Our model allows the user to edit geometry by changing the edge maps. The changes of edges are highlighted in red boxes.}
\label{fig:exp_edge_edit}
\end{figure}

Our proposed framework allows users to perform the common fine-grained portrait image editing operations using a single trained model, which supports geometry, color, light, and shadow editing. It is also flexible to extend the model to support more editing operations, e.g., freckle editing. In the experiments below, we modify the inputs in $\{\mI_{E}, \mI_{CP}, \mI_{L}, \mI_{S}\}$ to demonstrate how they control the synthesised images.

\noindent\textbf{Edge maps.}
The edge maps extracted from the original portrait images provide the most important structural information for image reconstruction, so it is more convenient to directly modify the extracted edge maps for minor geometry editing. As the examples shown in Fig.~\ref{fig:exp_edge_edit}, we modify $\mI_{E}$ and keep the other constraint inputs $\mI_{CP}, \mI_{L}, \mI_{S}$ unchanged to demonstrate many interesting applications, including add or remove accessories like eyeglasses and jewelry, hair style design, adjust facial components, and so on. With the help of our model, the users can conveniently generate photo-realistic images with just a few stroke editing steps, while these editing operations may be very complex and time consuming with the traditional photo editing software.

\begin{figure}[h!]
\centering
{\footnotesize
\makebox[0.16\columnwidth]{}
\makebox[0.16\columnwidth]{Scribbler \citep{sangkloy2017scribbler}}
\makebox[0.16\columnwidth]{}
\makebox[0.16\columnwidth]{} 
\vspace{2pt} \\
\includegraphics[width=0.16\columnwidth]{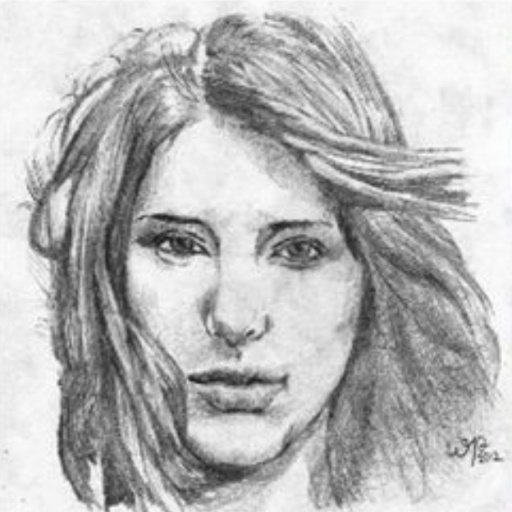}
\includegraphics[width=0.16\columnwidth]{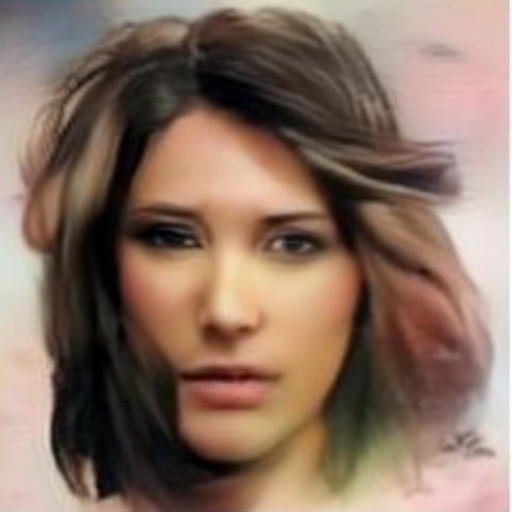}
\includegraphics[width=0.16\columnwidth]{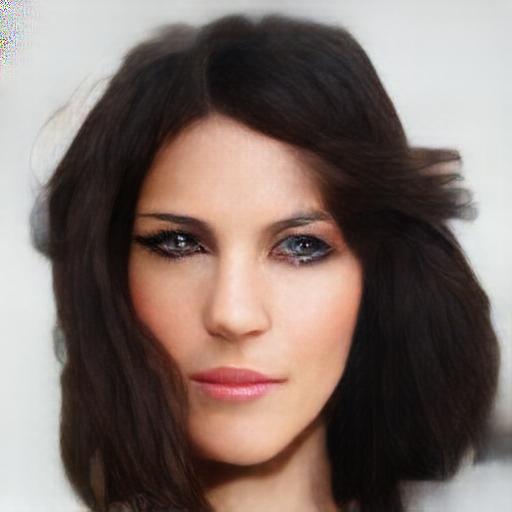}
\includegraphics[width=0.16\columnwidth]{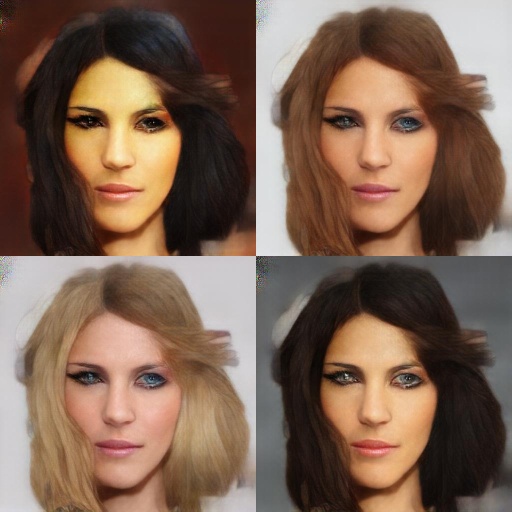} \\
\includegraphics[width=0.16\columnwidth]{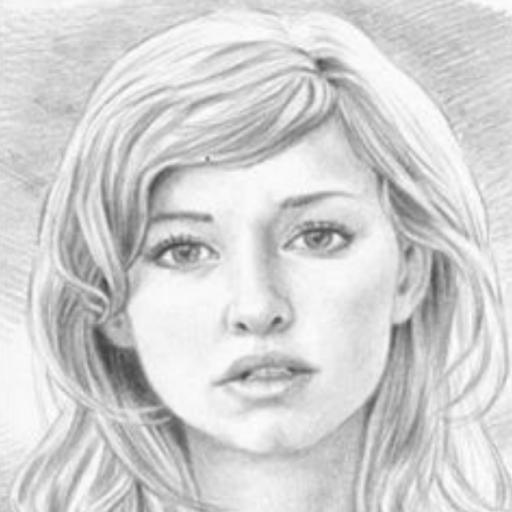}
\includegraphics[width=0.16\columnwidth]{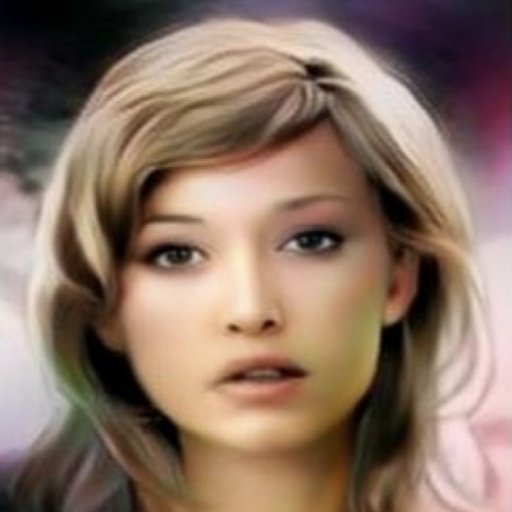}
\includegraphics[width=0.16\columnwidth]{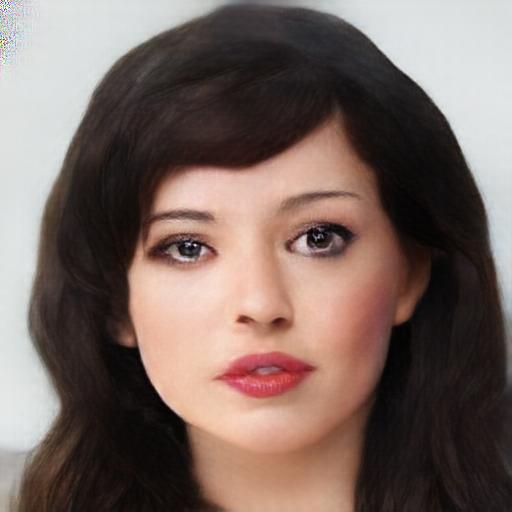}
\includegraphics[width=0.16\columnwidth]{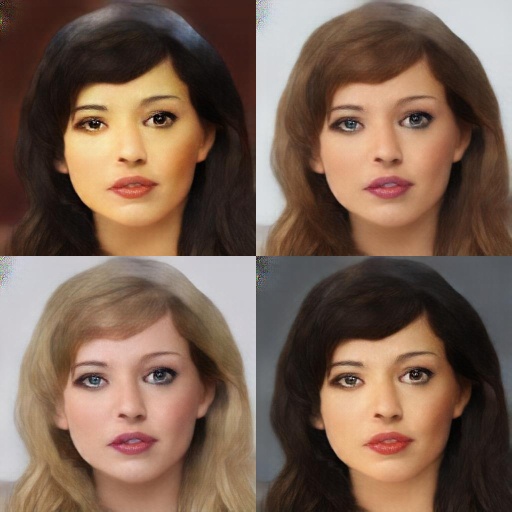} \\
\hdashrule{\columnwidth}{1pt}{1pt} \\
\makebox[0.16\columnwidth]{}
\makebox[0.16\columnwidth]{DeepFaceEditing \citep{chen2021DeepFaceEditing}}
\makebox[0.16\columnwidth]{}
\makebox[0.16\columnwidth]{}
\vspace{2pt}
\\
\includegraphics[width=0.16\columnwidth]{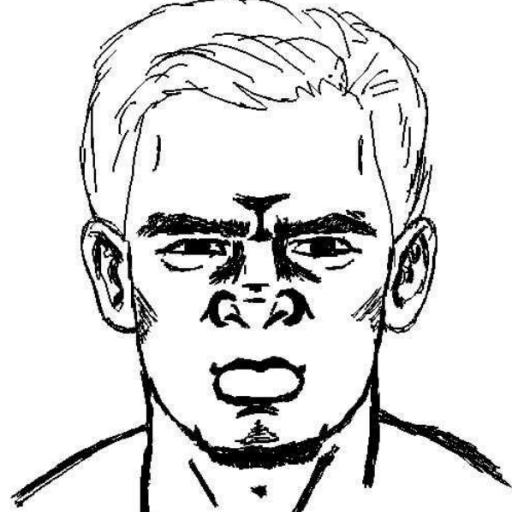}
\includegraphics[width=0.16\columnwidth]{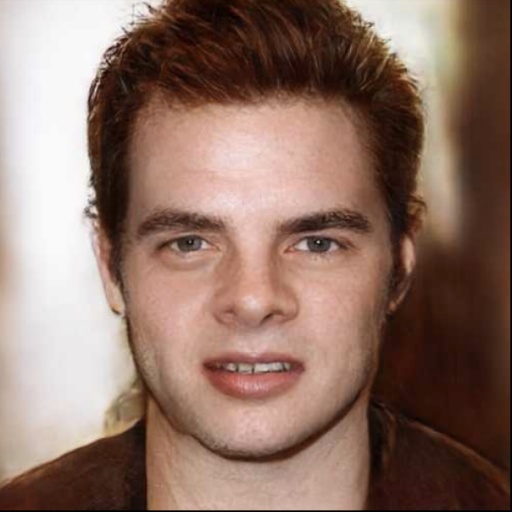}
\includegraphics[width=0.16\columnwidth]{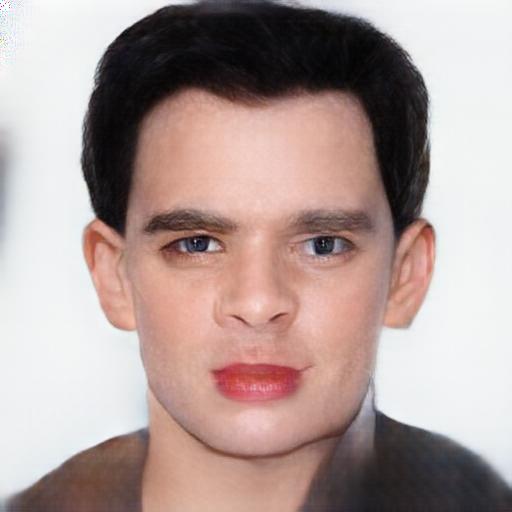} 
\includegraphics[width=0.16\columnwidth]{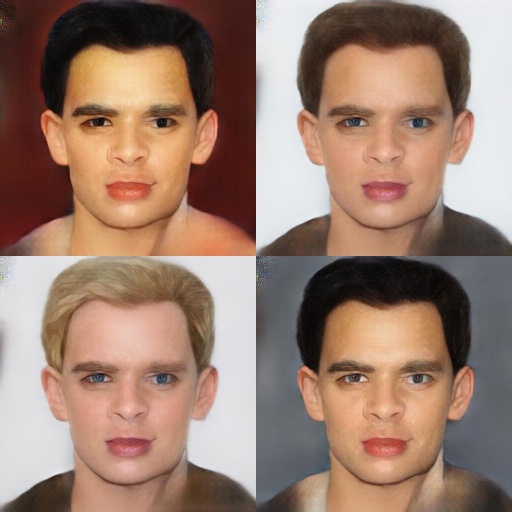} \\
\includegraphics[width=0.16\columnwidth]{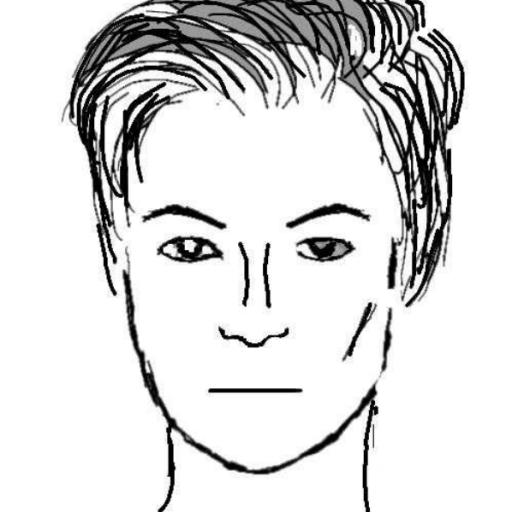}
\includegraphics[width=0.16\columnwidth]{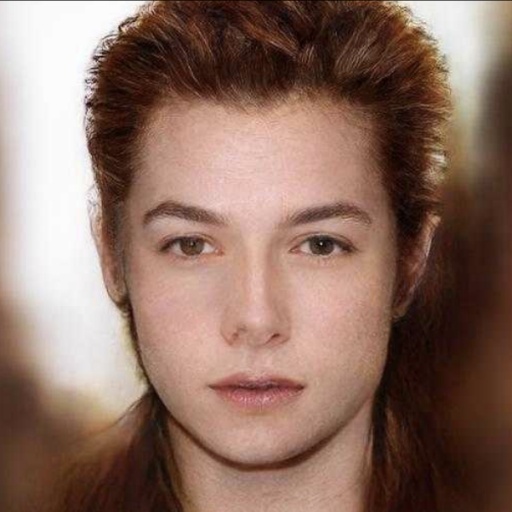}
\includegraphics[width=0.16\columnwidth]{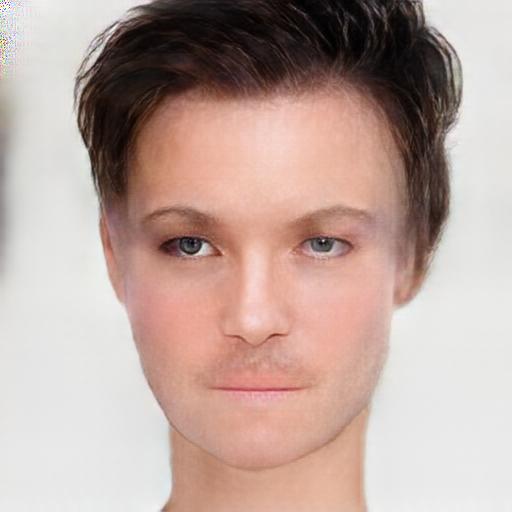}
\includegraphics[width=0.16\columnwidth]{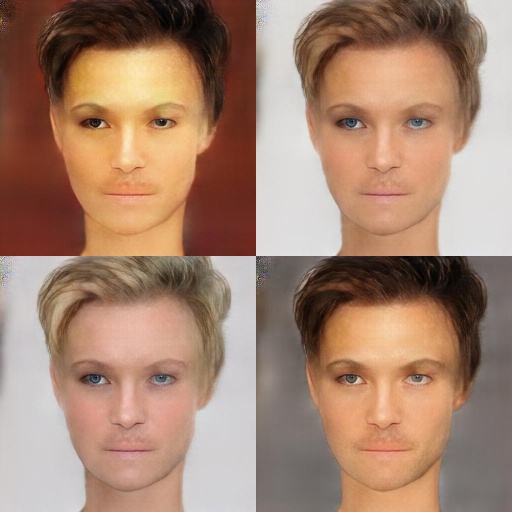} \\
\includegraphics[width=0.16\columnwidth]{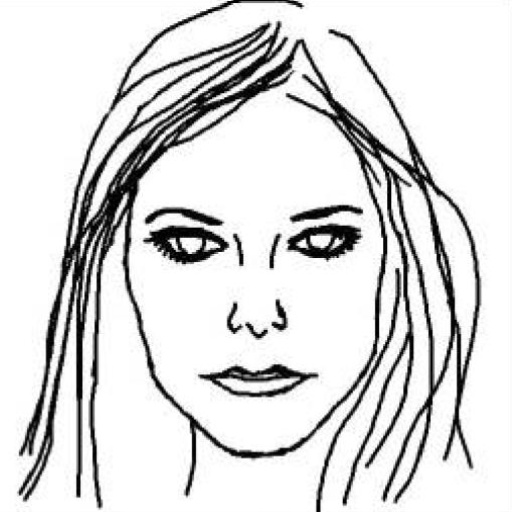}
\includegraphics[width=0.16\columnwidth]{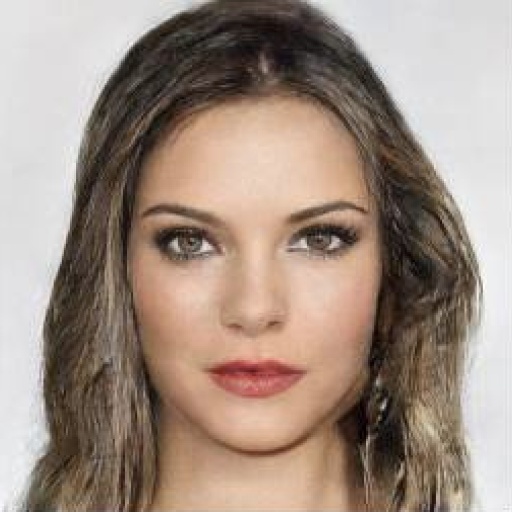}
\includegraphics[width=0.16\columnwidth]{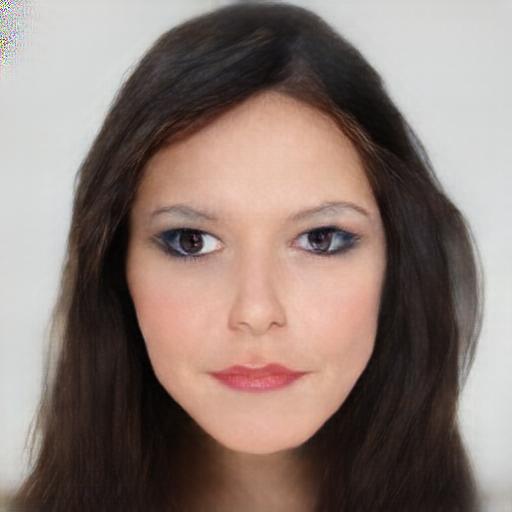} 
\includegraphics[width=0.16\columnwidth]{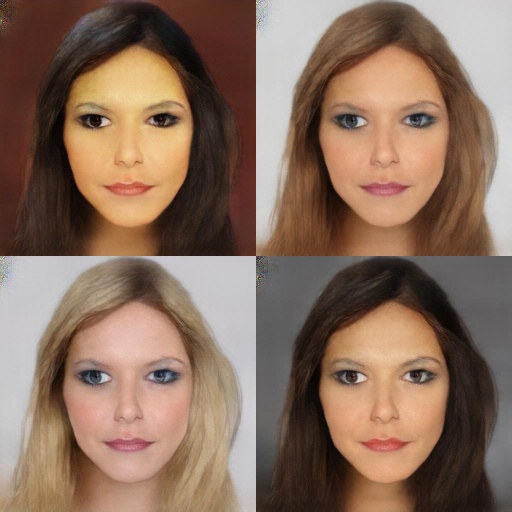} \\
\hdashrule{\columnwidth}{1pt}{1pt} \\
\makebox[0.16\columnwidth]{}
\makebox[0.16\columnwidth]{DeepPlasticSurgery \citep{yang2020deep}}
\makebox[0.16\columnwidth]{}
\makebox[0.16\columnwidth]{}
\vspace{2pt}
\\
\includegraphics[width=0.16\columnwidth]{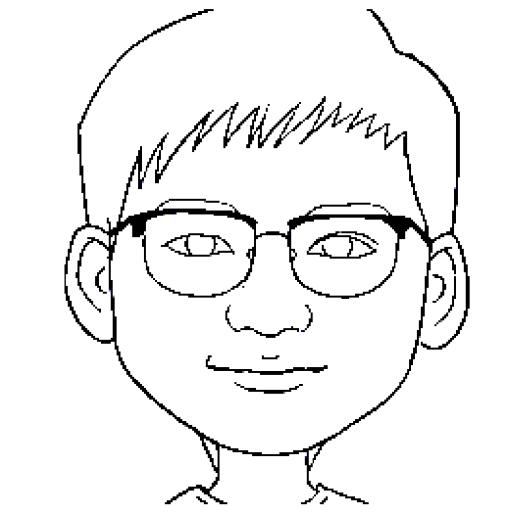}
\includegraphics[width=0.16\columnwidth]{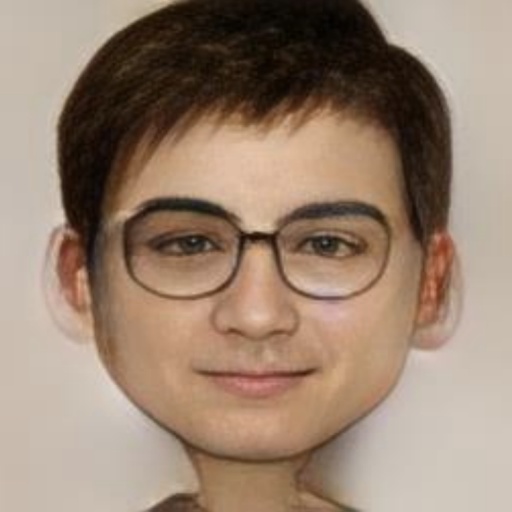}
\includegraphics[width=0.16\columnwidth]{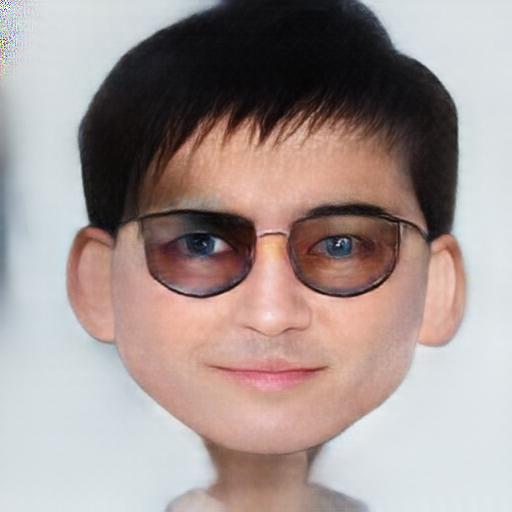}
\includegraphics[width=0.16\columnwidth]{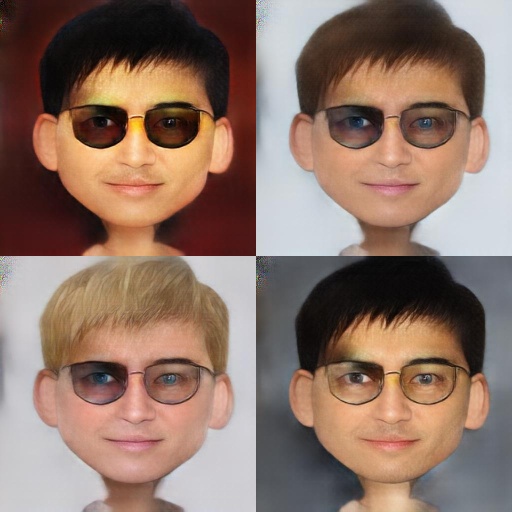} \\
%\hdashrule{\columnwidth}{1pt}{1pt} \\
%\makebox[0.16\columnwidth]{}
%\makebox[0.16\columnwidth]{Reference Image}
%\makebox[0.16\columnwidth]{}
%\makebox[0.16\columnwidth]{}
%\vspace{2pt}
%\\
%\includegraphics[width=0.16\columnwidth]{figs/hd12_r_edge.jpg}
%\includegraphics[width=0.16\columnwidth]{figs/hd12_base.jpg}
%\includegraphics[width=0.16\columnwidth]{figs/hd12_r_fake.jpg}
%\includegraphics[width=0.16\columnwidth]{figs/hd12_r_fake_concat.jpg} \\
\makebox[0.16\columnwidth]{Sketches}
\makebox[0.16\columnwidth]{Baseline}
\makebox[0.16\columnwidth]{Ours}
\makebox[0.16\columnwidth]{More Colors (Ours)} \\
}
\caption{Though our method is not designed for sketches-to-image translation, it can generate visually pleasing images from hand-drawn sketches of various styles.}
%\textcolor{red}{The sketch in the last row is hand drawn by ourselves based on the reference image in the second column. --> put the result of the last row into teaser??}
\label{fig:exp_sketch2img}
\end{figure}

\noindent\textbf{Hand-drawn sketches.}
Although our framework is not designed to handle hand-drawn sketches on purpose, we find it can generate compelling results for hand-drawn sketch to image translation. In Fig.~\ref{fig:exp_sketch2img}, we compare our results with Scribbler \citep{sangkloy2017scribbler}, DeepFaceEditing \citep{chen2021DeepFaceEditing} and DeepPlasticSurgery \citep{yang2020deep}. In addition, we also test on an example drawn by ourselves. We use the model trained without light and shadow masks in this experiment for fair comparison with the baseline methods. As we can see from Fig.~\ref{fig:exp_sketch2img}, the style of these sketches are very different from the edge maps used to train the model, so we apply additional edge processing steps: 1) apply the steps described in \S\ref{sec:edge_map_extract} to extract edges from the Scribbler \citep{sangkloy2017scribbler} sketches, and keep the other sketches unchanged; 2) concatenate the edges and random color palettes as inputs to generate the intermediate images; 3) extract edges (\S\ref{sec:edge_map_extract}) from the intermediate images, and repeat step 2 to generate the final results. Scribbler \citep{sangkloy2017scribbler} is also designed for fine-grained colorization, but compared with their results the color of our method is more photo-realistic. Besides, compared with DeepFaceEditing \citep{chen2021DeepFaceEditing}, our results preserves more identity information about the sketches.

\subsubsection{Colors}

\begin{figure*}[t!]
\centering
{\footnotesize
\includegraphics[width=0.09\columnwidth]{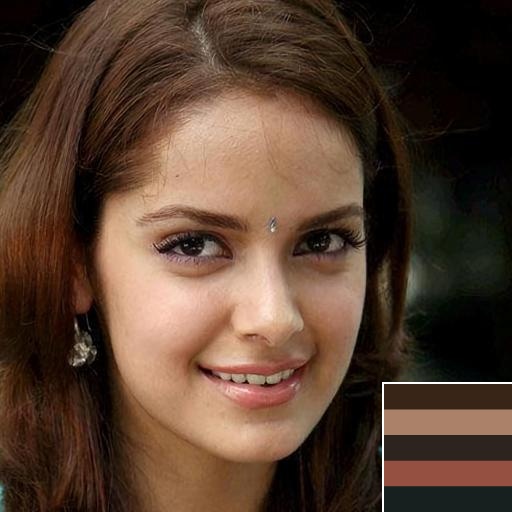}
\includegraphics[width=0.09\columnwidth]{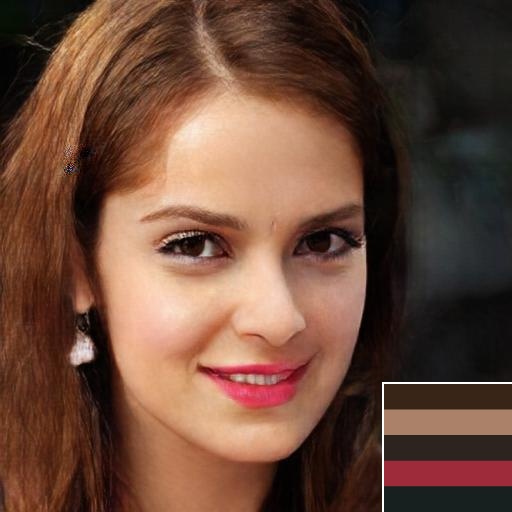}
\includegraphics[width=0.09\columnwidth]{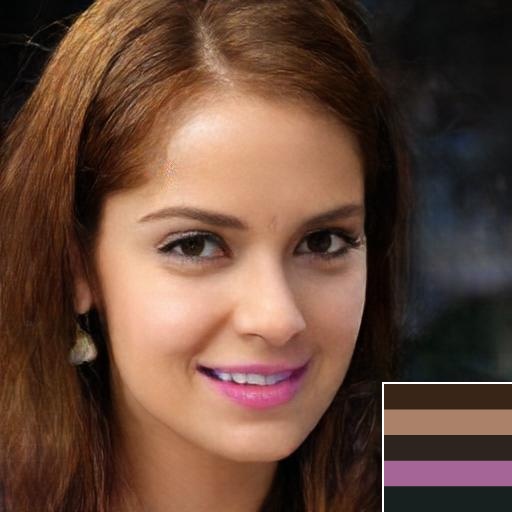}
\includegraphics[width=0.09\columnwidth]{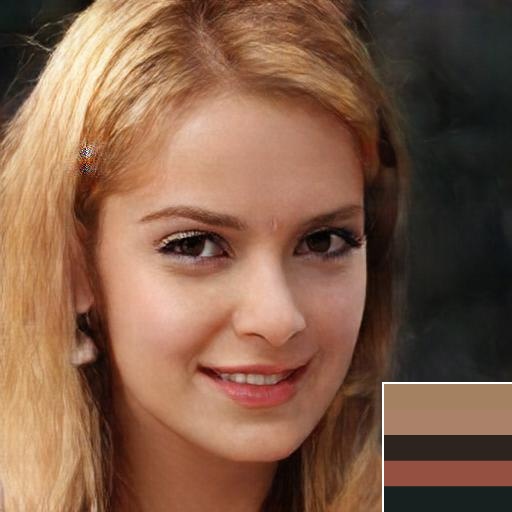}
\includegraphics[width=0.09\columnwidth]{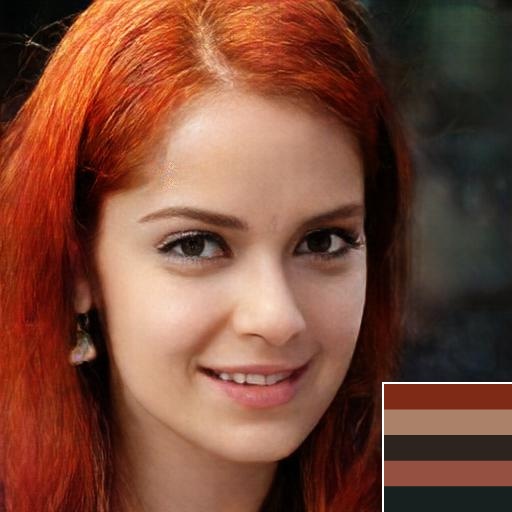}
\includegraphics[width=0.09\columnwidth]{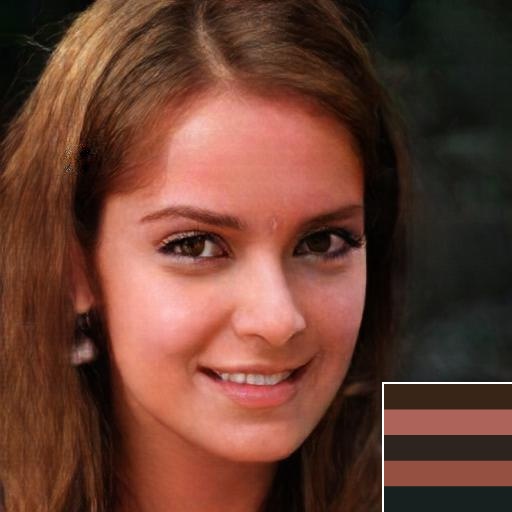}
\includegraphics[width=0.09\columnwidth]{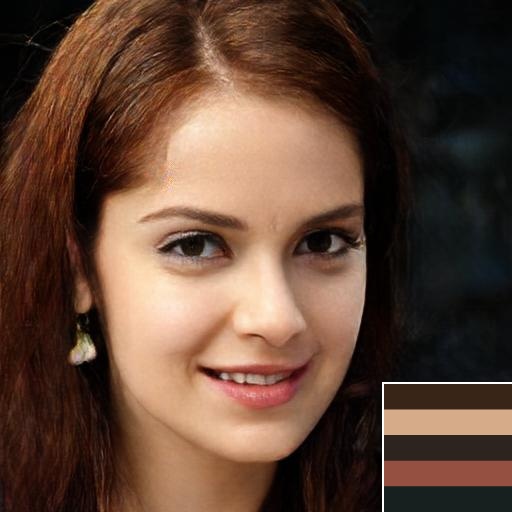}
\includegraphics[width=0.09\columnwidth]{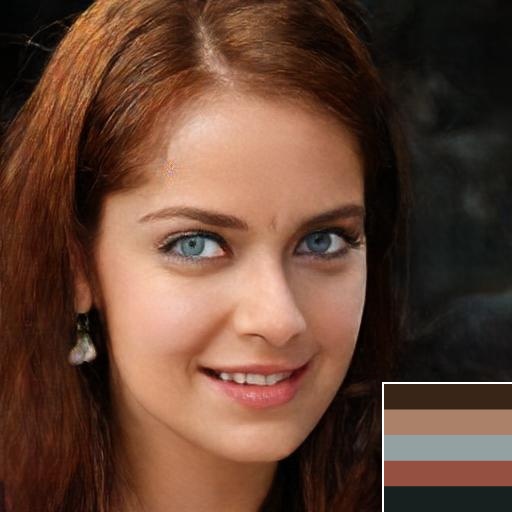} 
\includegraphics[width=0.09\columnwidth]{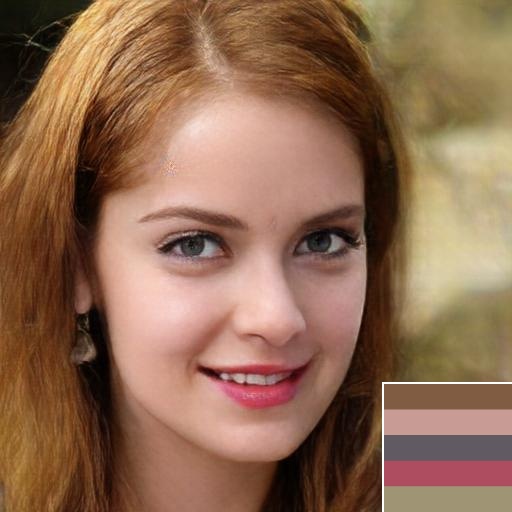} 
\\
\includegraphics[width=0.09\columnwidth]{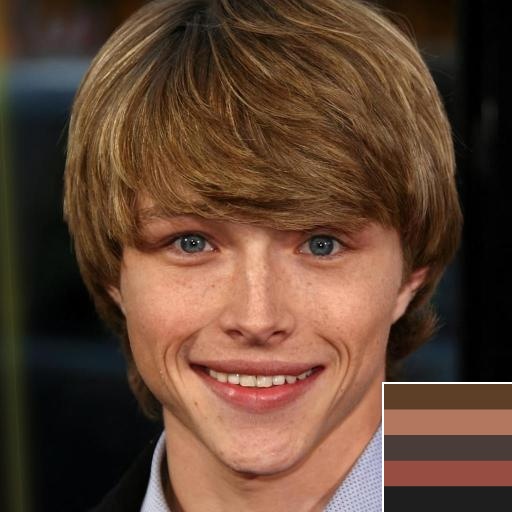}
\includegraphics[width=0.09\columnwidth]{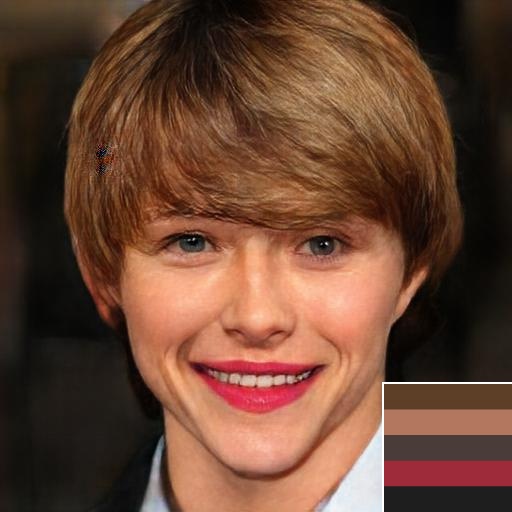}
\includegraphics[width=0.09\columnwidth]{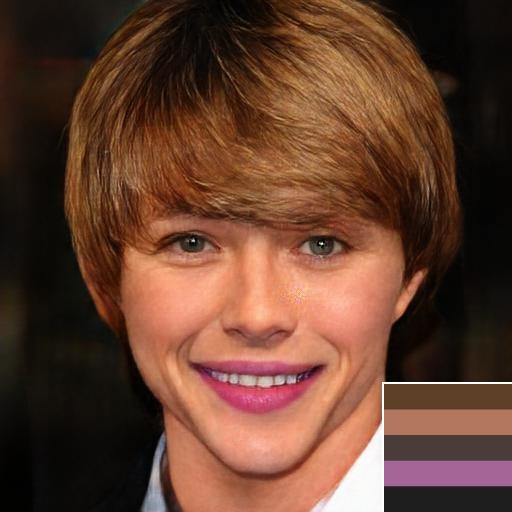}
\includegraphics[width=0.09\columnwidth]{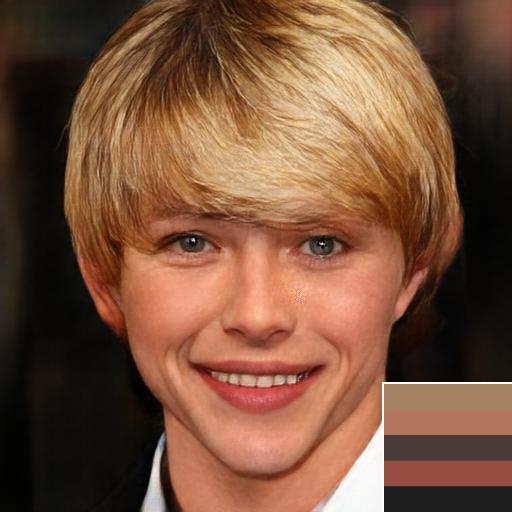}
\includegraphics[width=0.09\columnwidth]{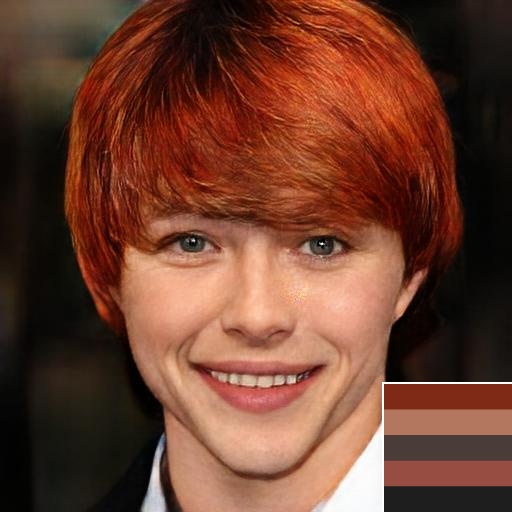}
\includegraphics[width=0.09\columnwidth]{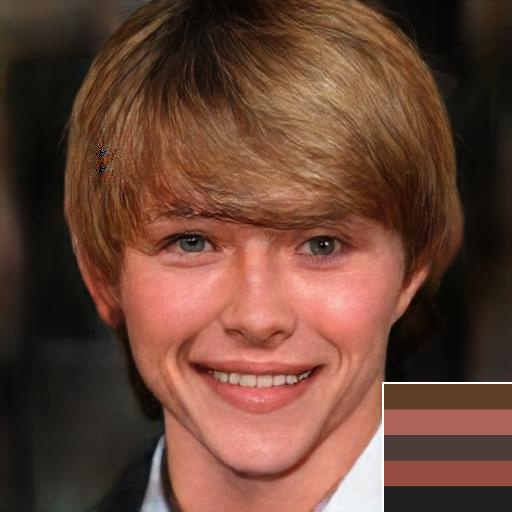}
\includegraphics[width=0.09\columnwidth]{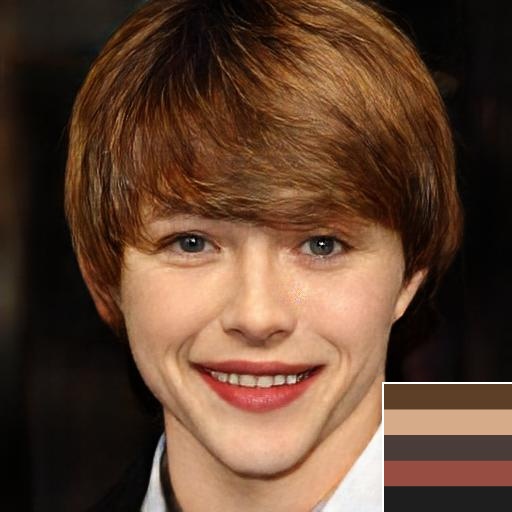}
\includegraphics[width=0.09\columnwidth]{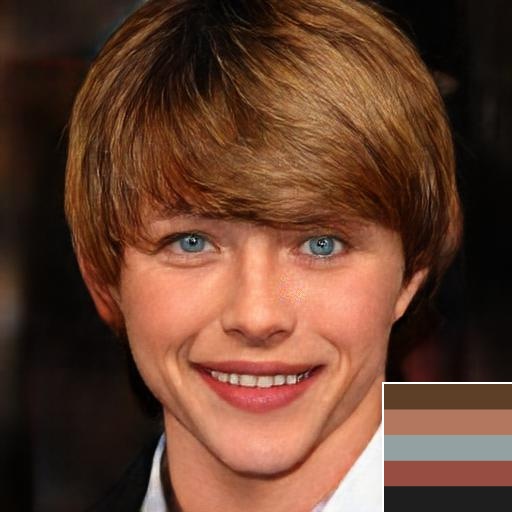}
\includegraphics[width=0.09\columnwidth]{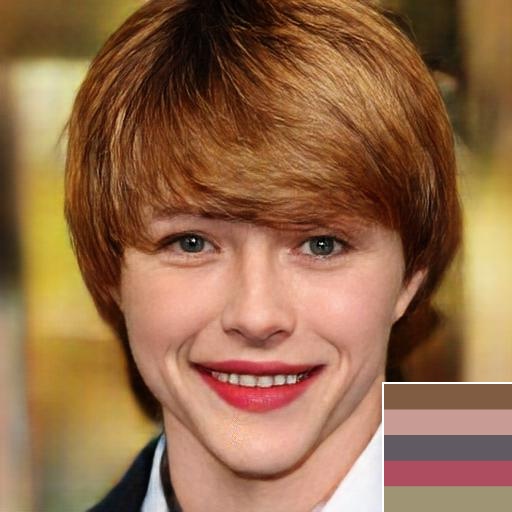} 
\\
\includegraphics[width=0.09\columnwidth]{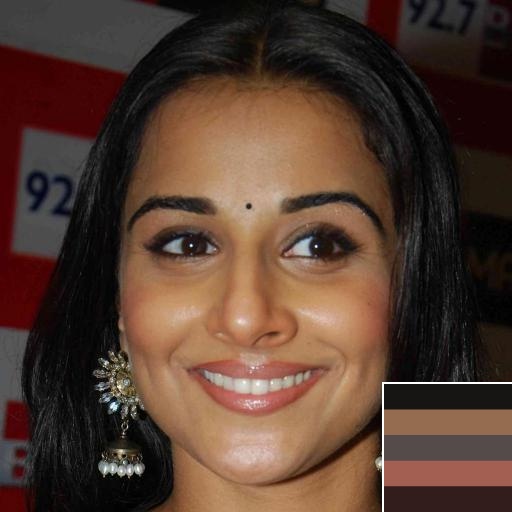}
\includegraphics[width=0.09\columnwidth]{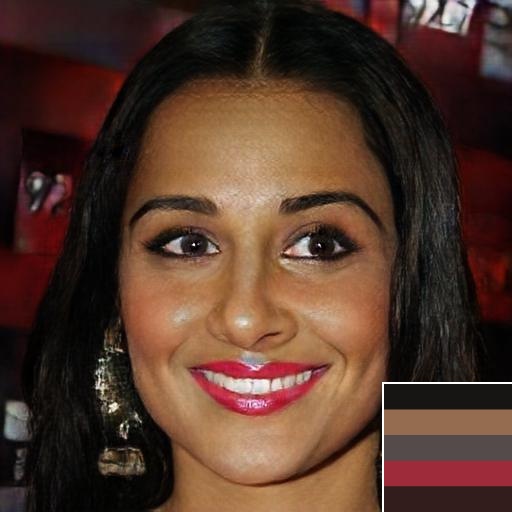}
\includegraphics[width=0.09\columnwidth]{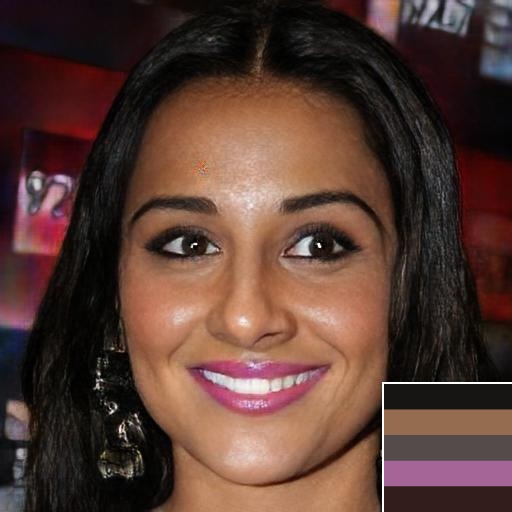}
\includegraphics[width=0.09\columnwidth]{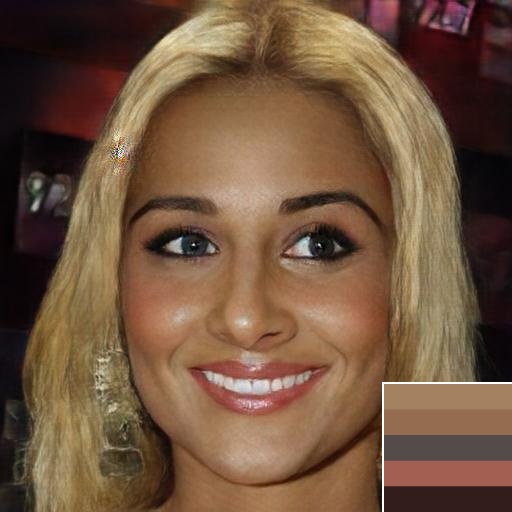}
\includegraphics[width=0.09\columnwidth]{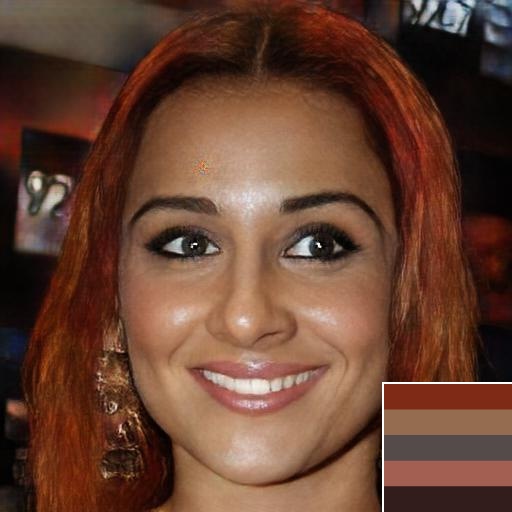}
\includegraphics[width=0.09\columnwidth]{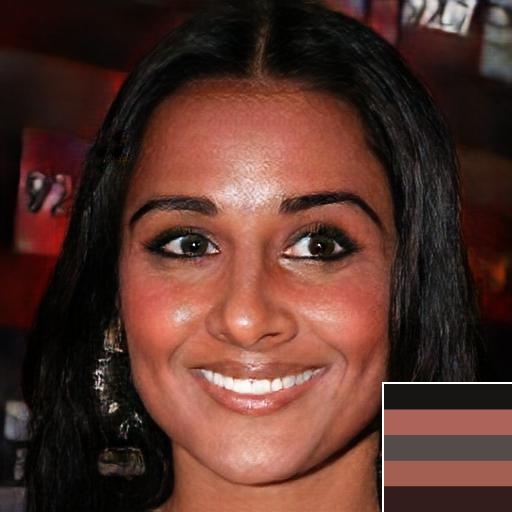} 
\includegraphics[width=0.09\columnwidth]{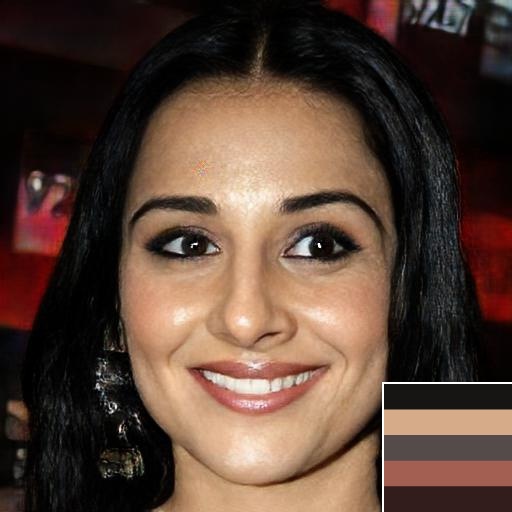}
\includegraphics[width=0.09\columnwidth]{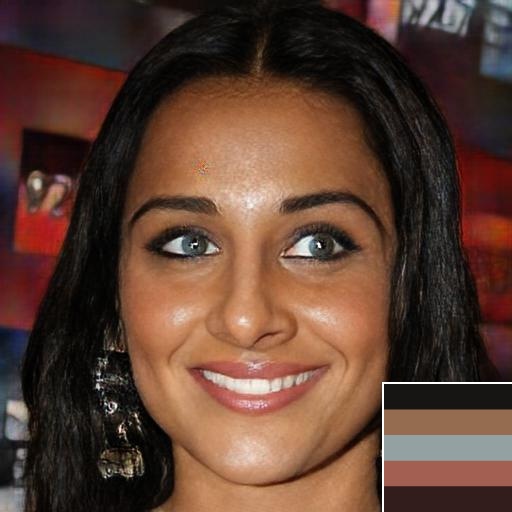} 
\includegraphics[width=0.09\columnwidth]{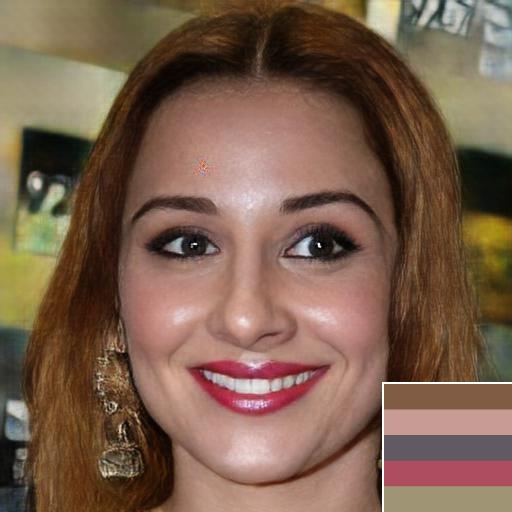} 
\\
\includegraphics[width=0.09\columnwidth]{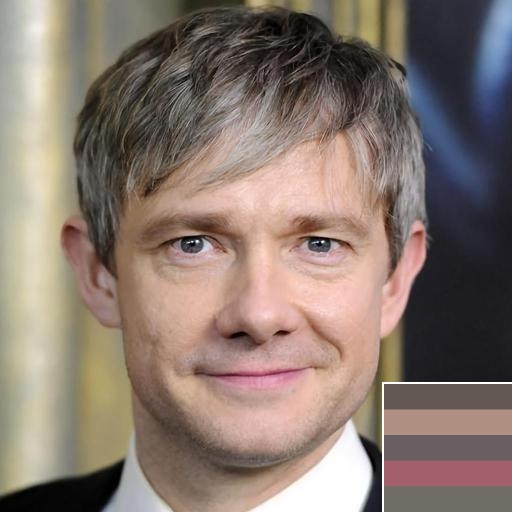}
\includegraphics[width=0.09\columnwidth]{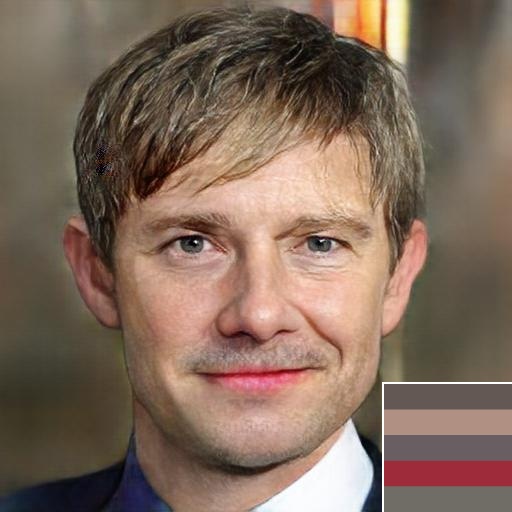}
\includegraphics[width=0.09\columnwidth]{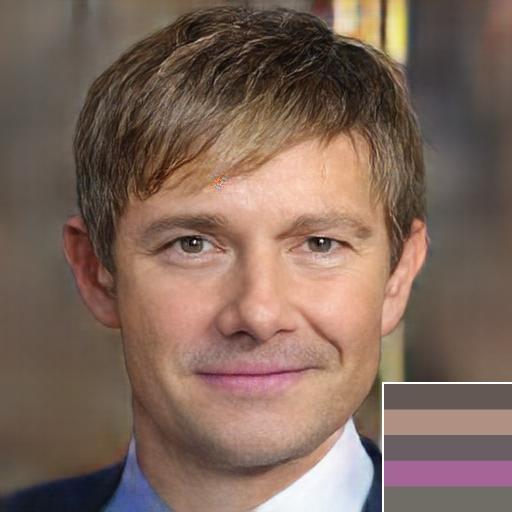}
\includegraphics[width=0.09\columnwidth]{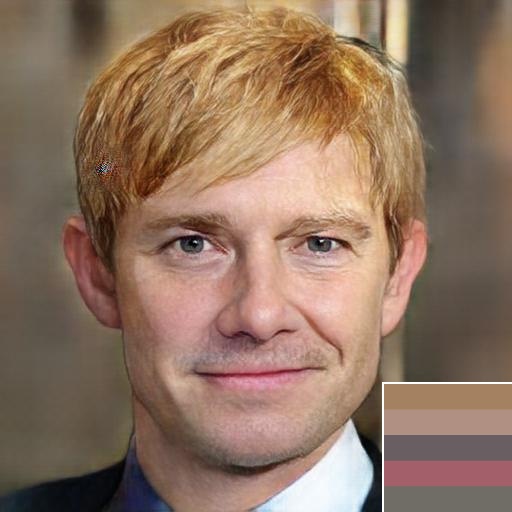}
\includegraphics[width=0.09\columnwidth]{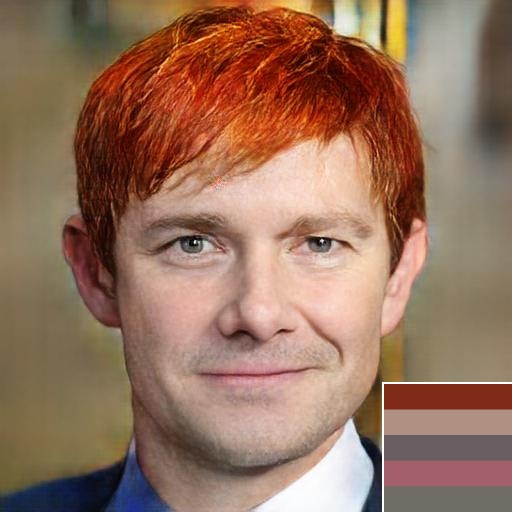}
\includegraphics[width=0.09\columnwidth]{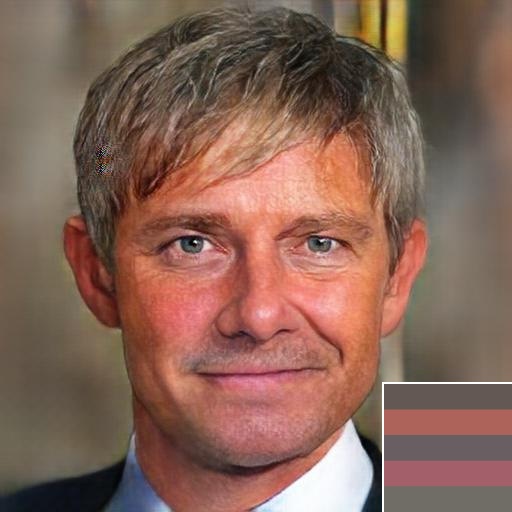}
\includegraphics[width=0.09\columnwidth]{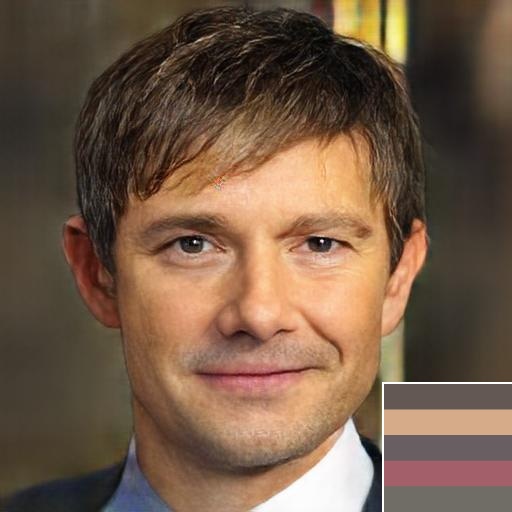}
\includegraphics[width=0.09\columnwidth]{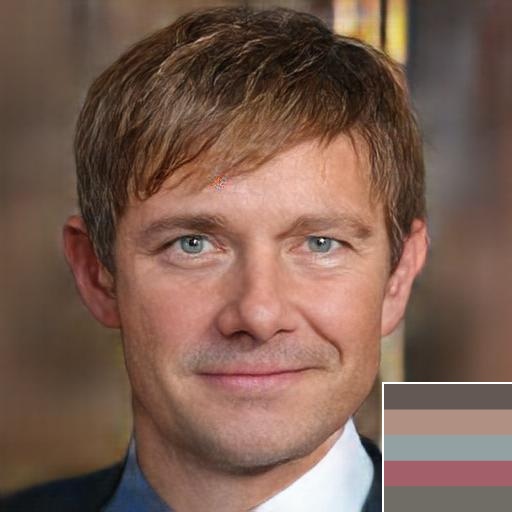}
\includegraphics[width=0.09\columnwidth]{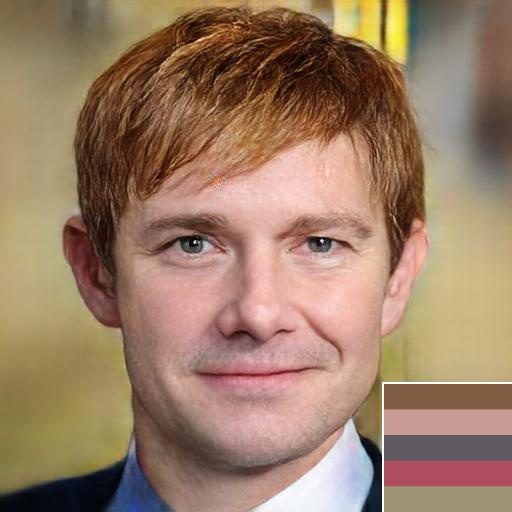}
\\
%\includegraphics[width=0.09\columnwidth]{figs/color_12360_r_real.jpg}
%\includegraphics[width=0.09\columnwidth]{figs/color_12360_r_lip.jpg}
%\includegraphics[width=0.09\columnwidth]{figs/color_12360_r_lip2.jpg}
%\includegraphics[width=0.09\columnwidth]{figs/color_12360_r_hair.jpg}
%\includegraphics[width=0.09\columnwidth]{figs/color_12360_r_hair2.jpg}
%\includegraphics[width=0.09\columnwidth]{figs/color_12360_r_skin3.jpg}
%\includegraphics[width=0.09\columnwidth]{figs/color_12360_r_skin2.jpg}
%\includegraphics[width=0.09\columnwidth]{figs/color_12360_r_eye.jpg}
%\includegraphics[width=0.09\columnwidth]{figs/color_12360_r_all.jpg}
%\\
\makebox[0.09\columnwidth]{Original}
\makebox[0.09\columnwidth]{Lip 1}
\makebox[0.09\columnwidth]{Lip 2}
\makebox[0.09\columnwidth]{Hair 1}
\makebox[0.09\columnwidth]{Hair 2}
\makebox[0.09\columnwidth]{Face Skin 1}
\makebox[0.09\columnwidth]{Face Skin 2}
\makebox[0.09\columnwidth]{Eye}
\makebox[0.09\columnwidth]{All}
\\
}
\caption{Fine-grained color editing. The color palettes are shown in the right bottom corner of the corresponding images. The same edge maps and shadow/light masks extracted from the original images are used as inputs without modification, which are not shown in the figure. Colors from top to bottom in the palette controls hair, skin, eyes, lips and background color, respectively. 
}
\label{fig:exp_color_edit}
\end{figure*}

\begin{figure}[t!]
\centering
{\footnotesize
\includegraphics[width=0.12\columnwidth]{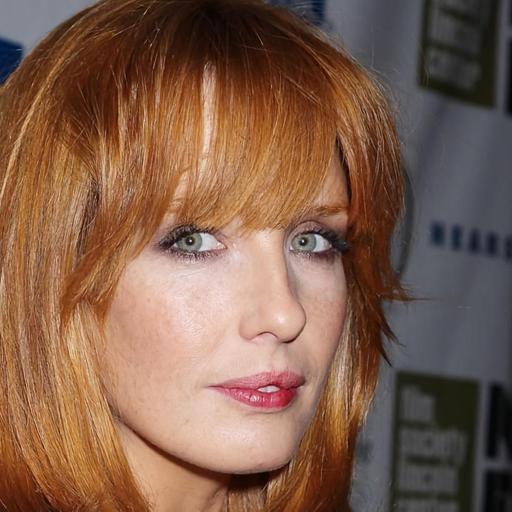}
\includegraphics[width=0.12\columnwidth]{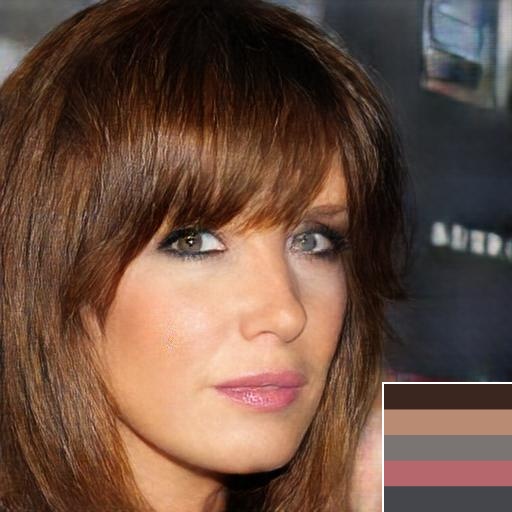}
\includegraphics[width=0.12\columnwidth]{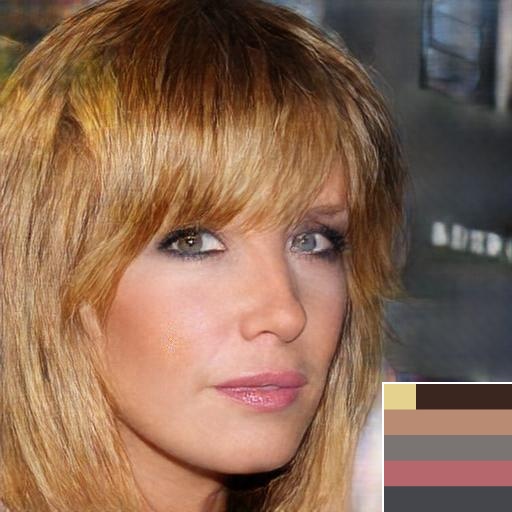}
\includegraphics[width=0.12\columnwidth]{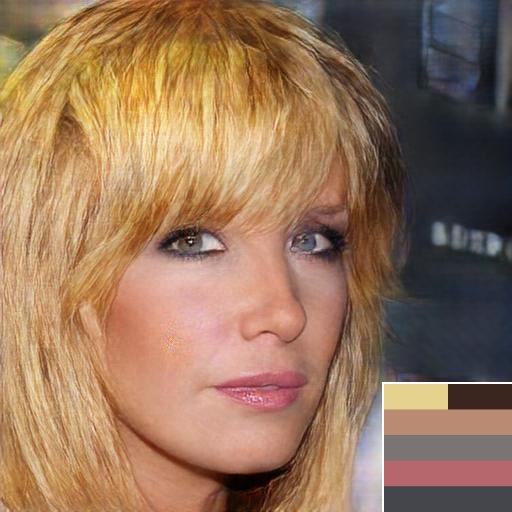}
\includegraphics[width=0.12\columnwidth]{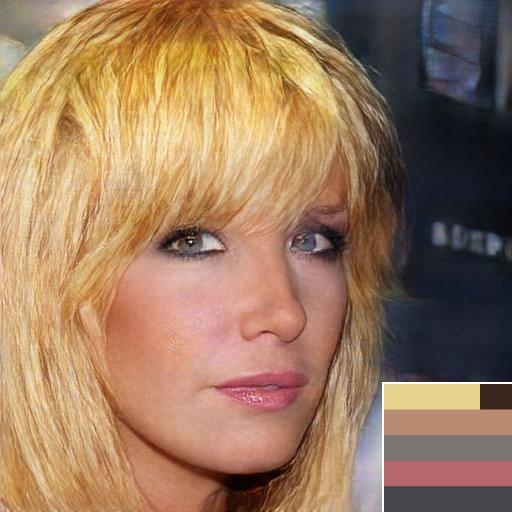}
\includegraphics[width=0.12\columnwidth]{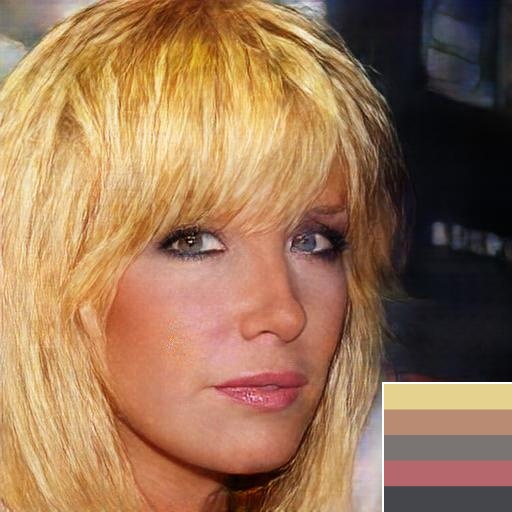}
\\
\includegraphics[width=0.12\columnwidth]{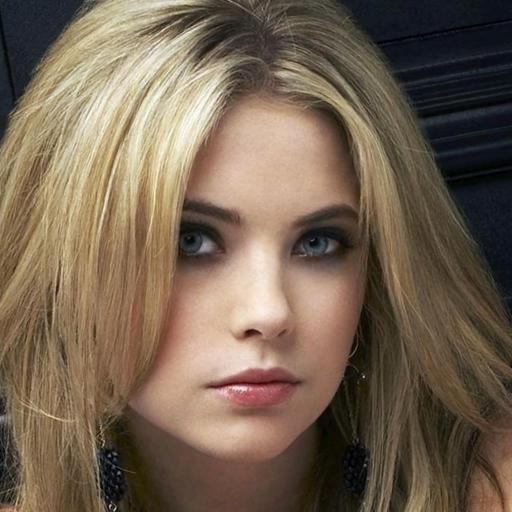}
\includegraphics[width=0.12\columnwidth]{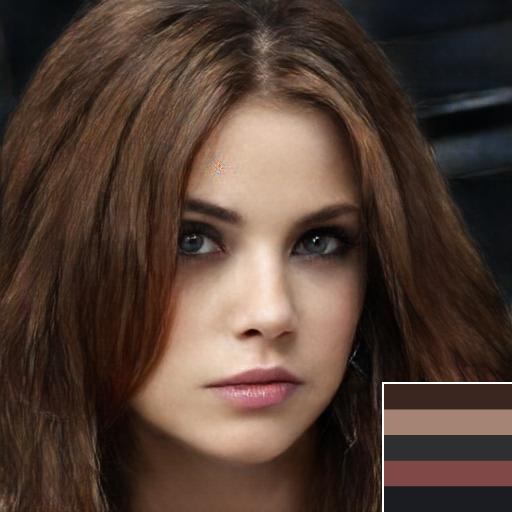}
\includegraphics[width=0.12\columnwidth]{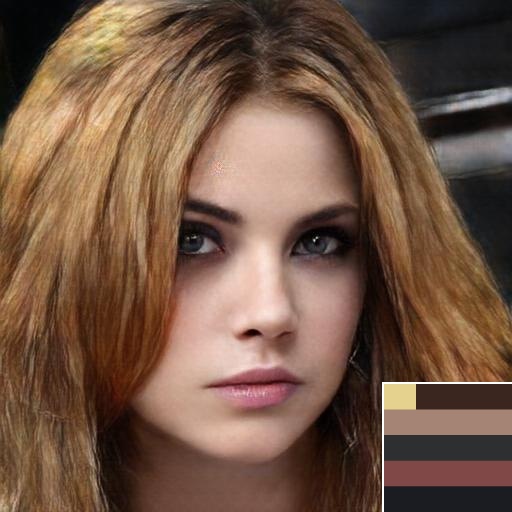}
\includegraphics[width=0.12\columnwidth]{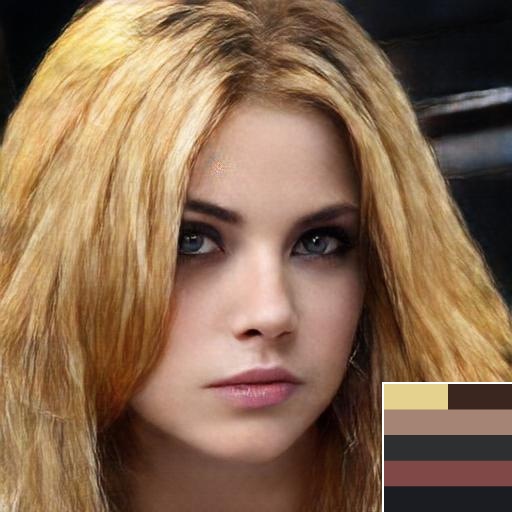}
\includegraphics[width=0.12\columnwidth]{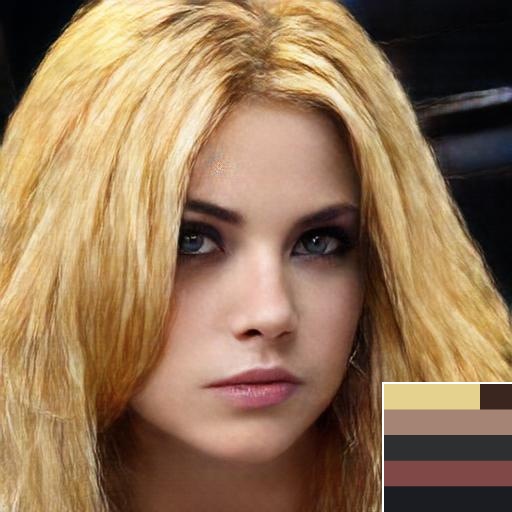}
\includegraphics[width=0.12\columnwidth]{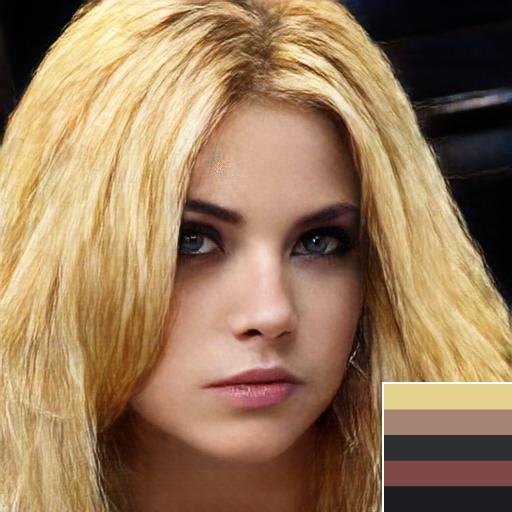}
\\
\includegraphics[width=0.12\columnwidth]{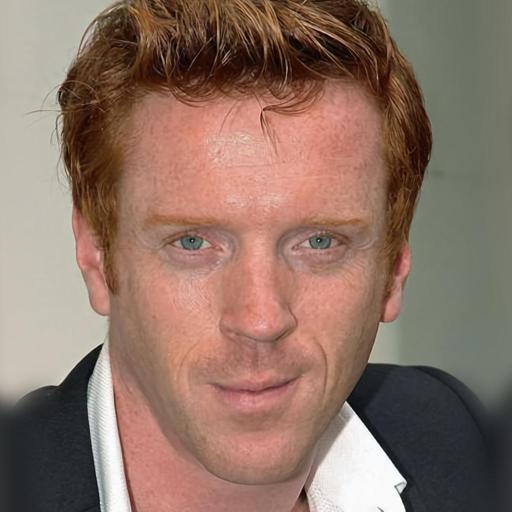}
\includegraphics[width=0.12\columnwidth]{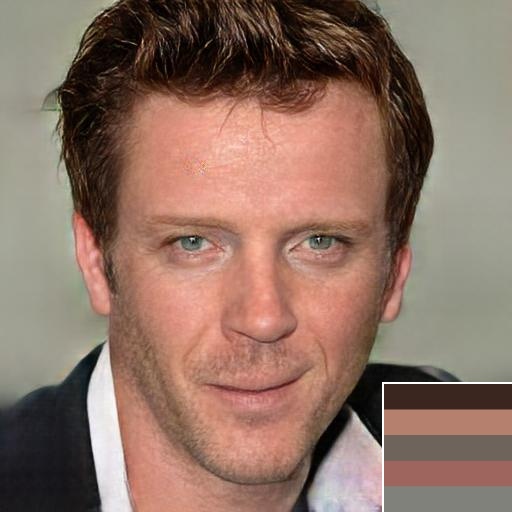}
\includegraphics[width=0.12\columnwidth]{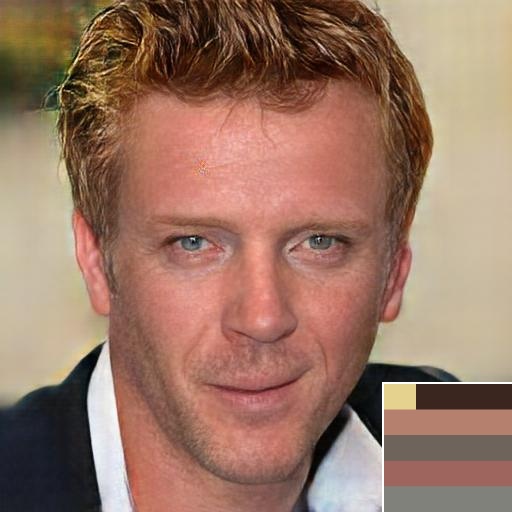}
\includegraphics[width=0.12\columnwidth]{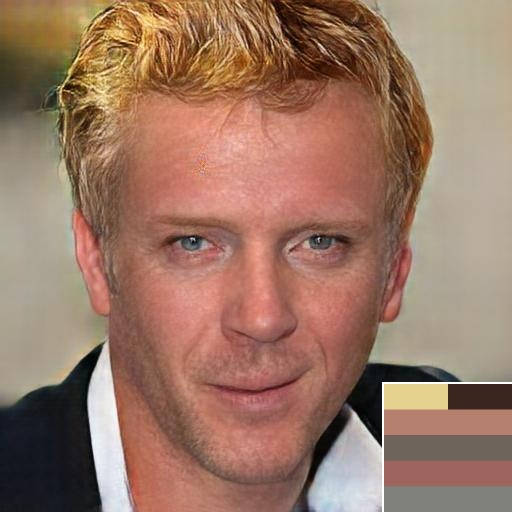}
\includegraphics[width=0.12\columnwidth]{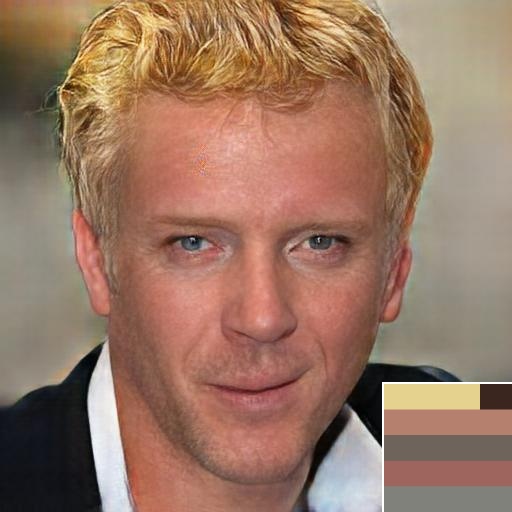}
\includegraphics[width=0.12\columnwidth]{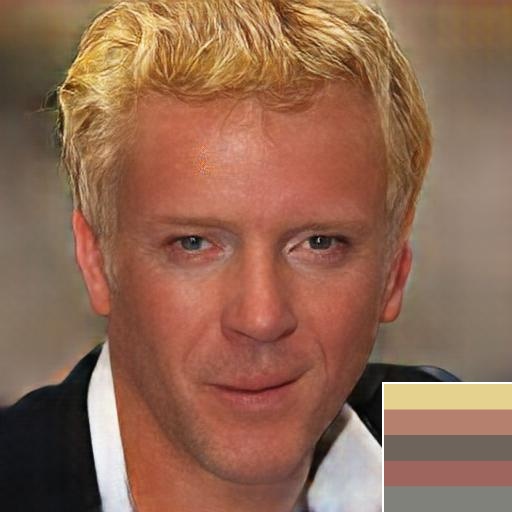}
\\
\includegraphics[width=0.12\columnwidth]{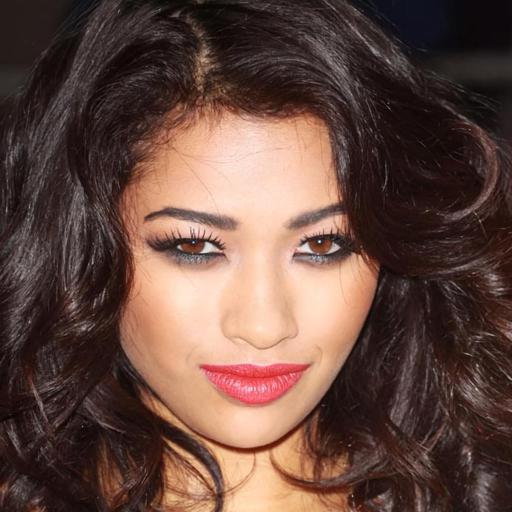}
\includegraphics[width=0.12\columnwidth]{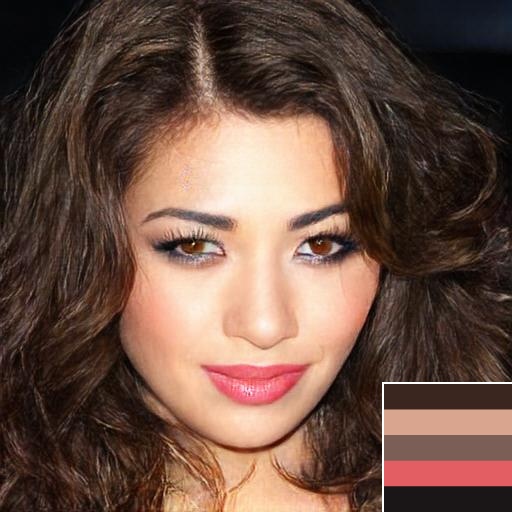}
\includegraphics[width=0.12\columnwidth]{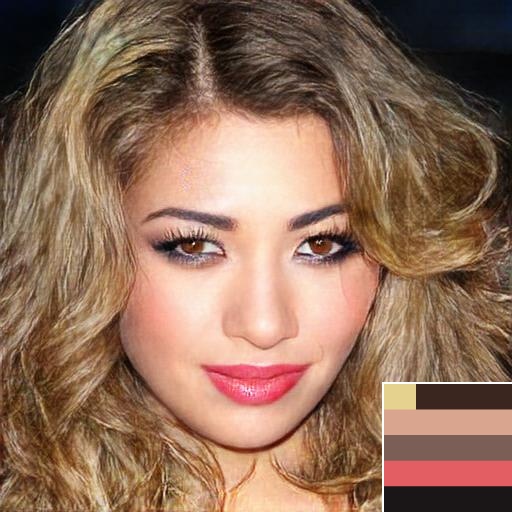}
\includegraphics[width=0.12\columnwidth]{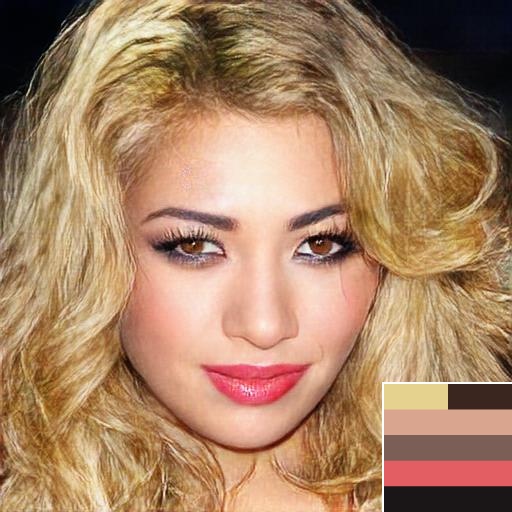}
\includegraphics[width=0.12\columnwidth]{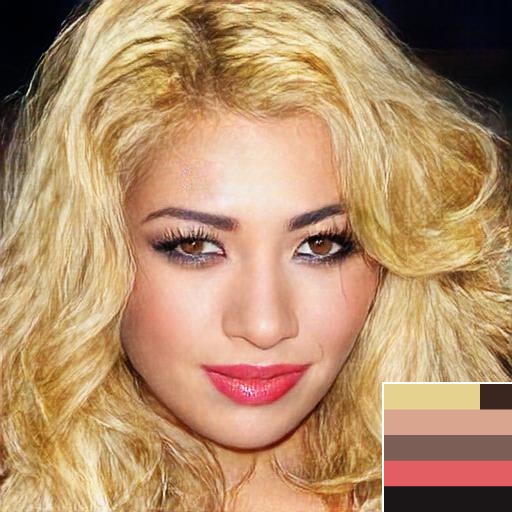}
\includegraphics[width=0.12\columnwidth]{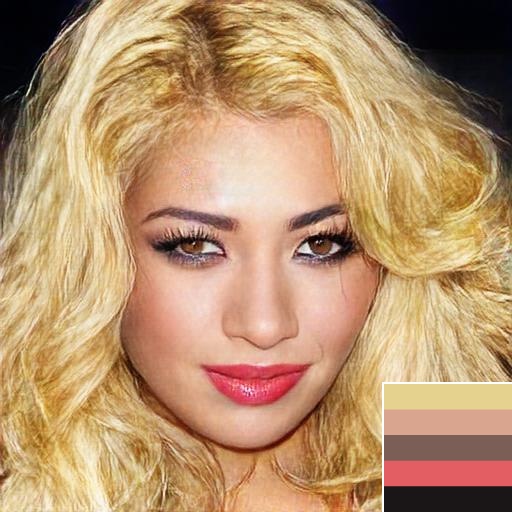}
\\
%\includegraphics[width=0.12\columnwidth]{figs/color_ctr_12333_r_real.jpg}
%\includegraphics[width=0.12\columnwidth]{figs/color_ctr_12333_r_1.jpg}
%\includegraphics[width=0.12\columnwidth]{figs/color_ctr_12333_r_0.25.jpg}
%\includegraphics[width=0.12\columnwidth]{figs/color_ctr_12333_r_0.5.jpg}
%\includegraphics[width=0.12\columnwidth]{figs/color_ctr_12333_r_0.75.jpg}
%\includegraphics[width=0.12\columnwidth]{figs/color_ctr_12333_r_0.jpg}
%\\
\makebox[0.12\columnwidth]{Original}
\makebox[0.12\columnwidth]{0\%}
\makebox[0.12\columnwidth]{25\%}
\makebox[0.12\columnwidth]{50\%}
\makebox[0.12\columnwidth]{75\%}
\makebox[0.12\columnwidth]{100\%}
\\
}
\caption{Color editing using sliders. The user blends two colors, brown and blonde by controlling the mixing ratio. See the first row in the color palettes. }
\label{fig:exp_color_edit_ctr}
\end{figure}

\begin{figure}[t!]
\centering
{\footnotesize
\includegraphics[width=0.12\columnwidth]{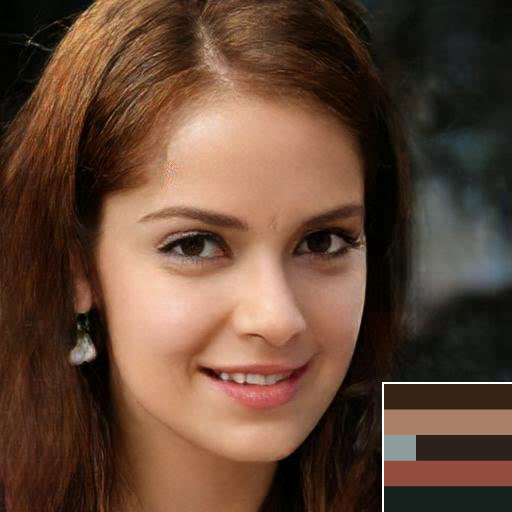}
\includegraphics[width=0.12\columnwidth]{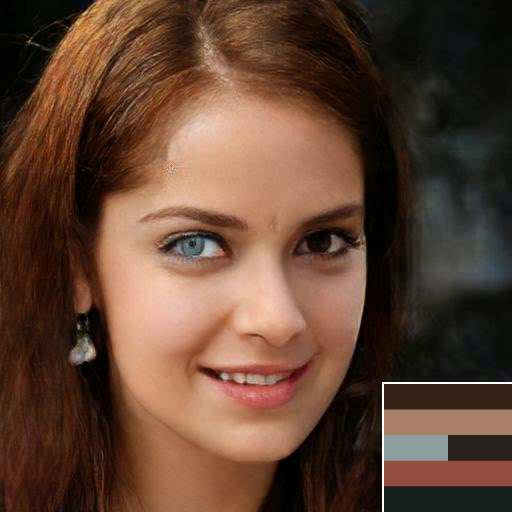}
\includegraphics[width=0.12\columnwidth]{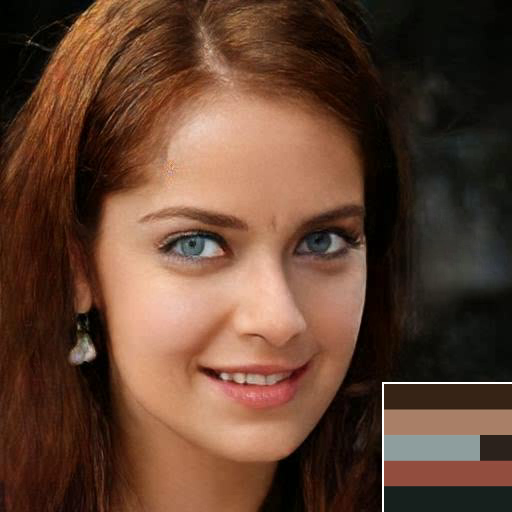}
\includegraphics[width=0.12\columnwidth]{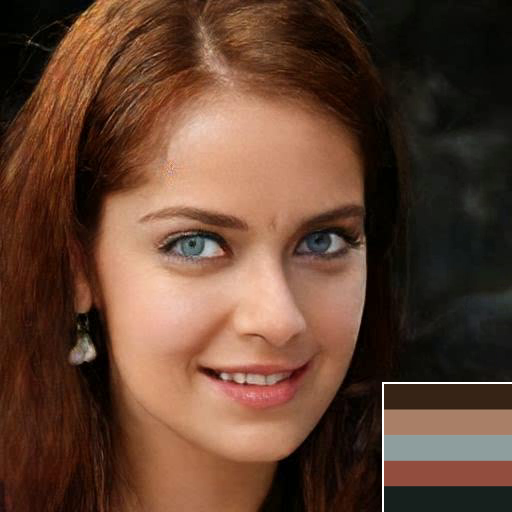}
\\
\includegraphics[width=0.12\columnwidth]{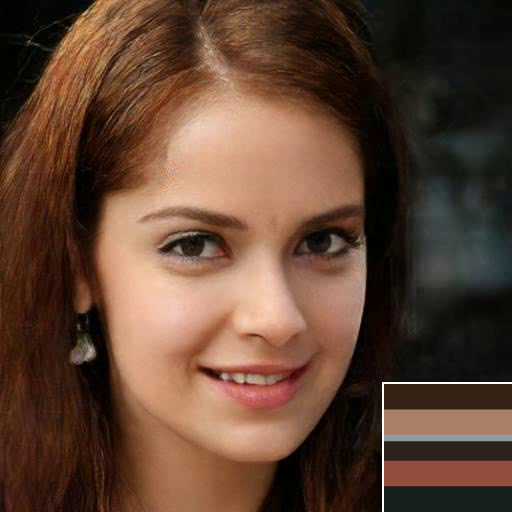}
\includegraphics[width=0.12\columnwidth]{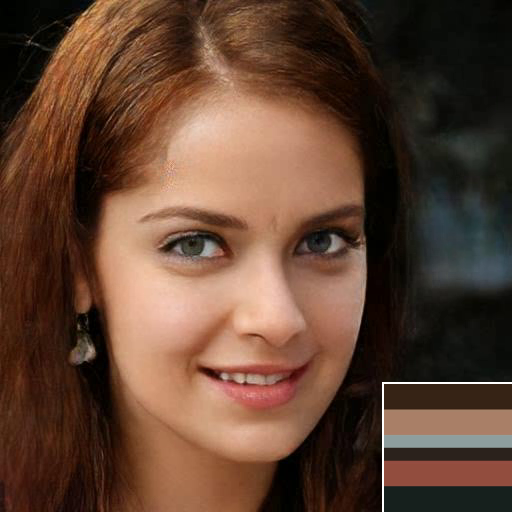}
\includegraphics[width=0.12\columnwidth]{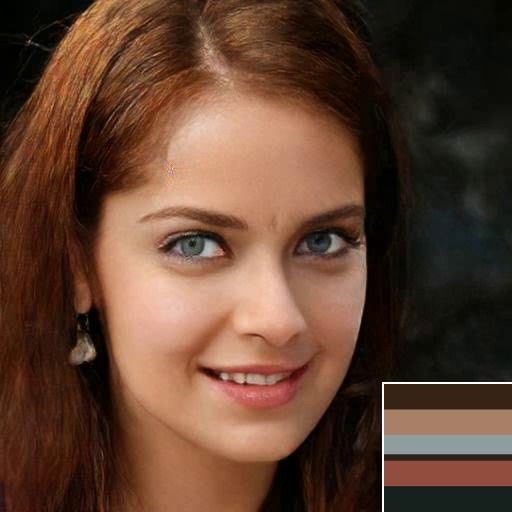}
\includegraphics[width=0.12\columnwidth]{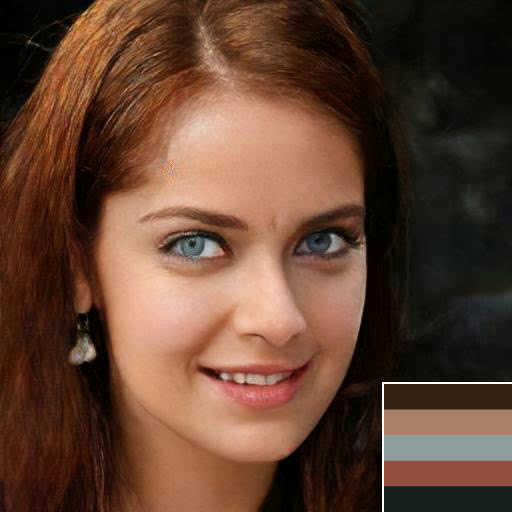}
\\
%\makebox[0.15\columnwidth]{Original}
%\makebox[0.10\columnwidth]{0\%}
\makebox[0.12\columnwidth]{25\%}
\makebox[0.12\columnwidth]{50\%}
\makebox[0.12\columnwidth]{75\%}
\makebox[0.12\columnwidth]{100\%}
\\
}
\caption{Horizontal (top) and vertical (bottom) color sliders for eye color editing. The percentage denotes the ratio of blue in the slider. See also the accompanying video demonstration.}
\label{fig:exp_color_slider_direction}
\end{figure}

In this section, we evaluate the proposed framework for fine-grained color editing. As the examples presented in Fig.~\ref{fig:exp_color_edit}, users only need to edit the color palettes $\mI_{CP}$ shown in the right bottom corner of the images to control the color of each facial components. Compared with the recent work \citep{afifi2021histogan,chen2021DeepFaceEditing}, our approach does not require any reference image, and gives users more freedom to specify any preferred color to edit the target facial components only, while keeping color of the other facial components unchanged. Compared with the other fine-grained color editing approaches \citep{sangkloy2017scribbler,zhang2017real,xiao2019interactive}, which usually require user to specify color points or strokes on the target components, our approach is more convenient for batch editing. In the experiments, we only demonstrated editing the color of hair, skin, eyes, lips and background, but it is convenient to extend the model to support more facial components.

Our model has been proven to support color editing using the combination of colors, although it is trained with a single color for each facial component. In Fig.~\ref{fig:exp_color_edit_ctr}, we show how to perform hair color editing using a slider to control the combination of two colors. As we adjust the ratio of the two colors by move the slider from right to left, the hair color in generated image also changes gradually. While most of the facial components are not very sensitive to the left/right position of the two colors when using slider, like the hair color editing examples shown in Fig.~\ref{fig:exp_color_edit_ctr}, so the horizontal sliders can be used to change their color. However, we find color of eyes changes in a different way. As the examples shown in the first row of Fig.~\ref{fig:exp_color_slider_direction}, color of the left eye changes first when we adjust the slider horizontally from left to right. It is possibly because the eyes occupy two separate segments in the images, while the other facial components, such as hair and face skin, only occupy one single segment. Therefore, we can use a vertical slider for more natural control of the eye colors as shown in the second row of Fig.~\ref{fig:exp_color_slider_direction}.

\begin{figure}[t!]
\centering
{\footnotesize
\makebox[0.12\columnwidth][c]{Original}
\makebox[0.12\columnwidth]{}
\makebox[0.12\columnwidth]{}
\makebox[0.12\columnwidth]{}
\makebox[0.12\columnwidth]{}
\vspace{2pt}
\\
\includegraphics[width=0.12\columnwidth]{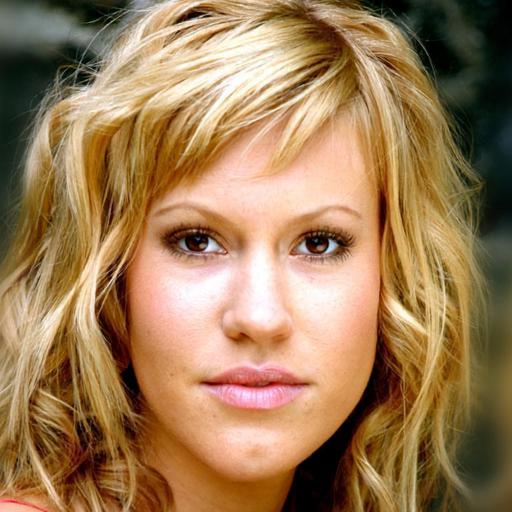}
\includegraphics[width=0.12\columnwidth]{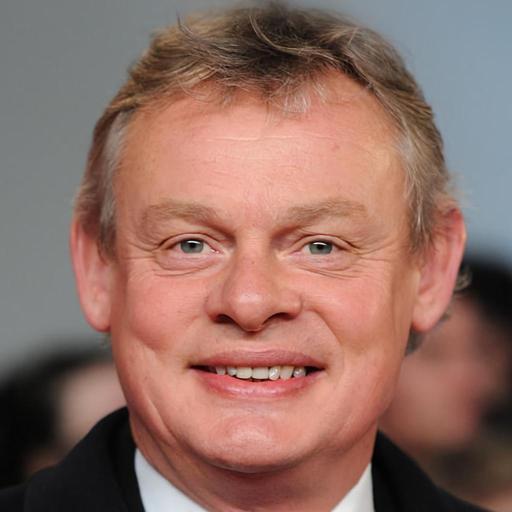}
\includegraphics[width=0.12\columnwidth]{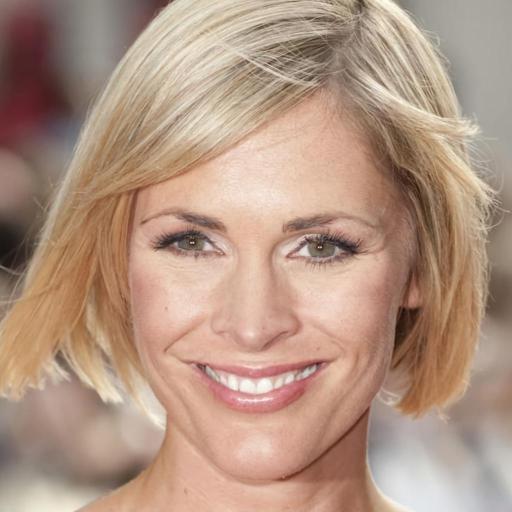}
\includegraphics[width=0.12\columnwidth]{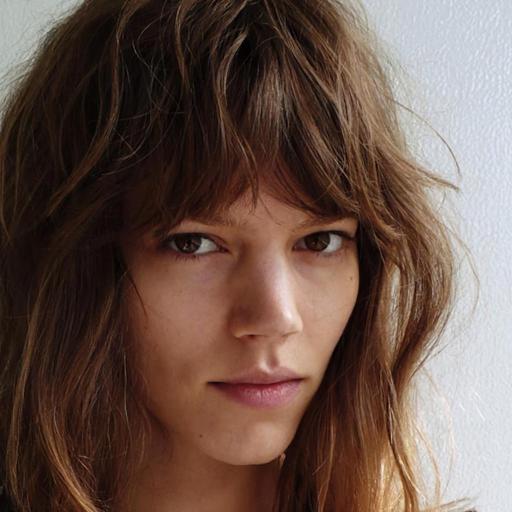}
\includegraphics[width=0.12\columnwidth]{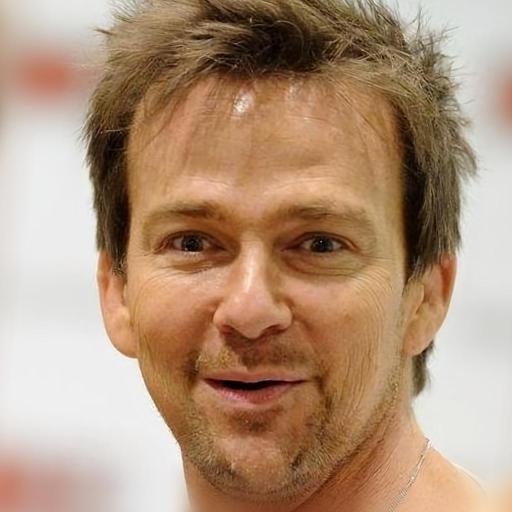}
\\
\makebox[0.12\columnwidth][c]{Editing}
\makebox[0.12\columnwidth]{}
\makebox[0.12\columnwidth]{}
\makebox[0.12\columnwidth]{}
\makebox[0.12\columnwidth]{}
\vspace{2pt}
\\
\includegraphics[width=0.12\columnwidth]{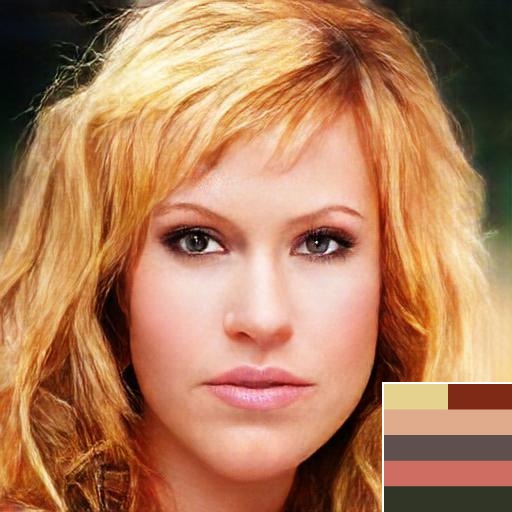}
\includegraphics[width=0.12\columnwidth]{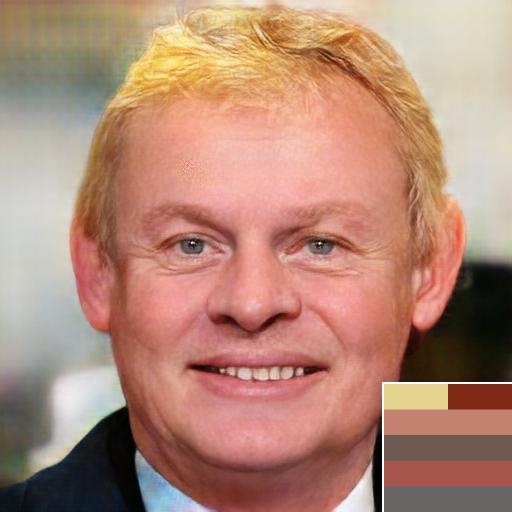}
\includegraphics[width=0.12\columnwidth]{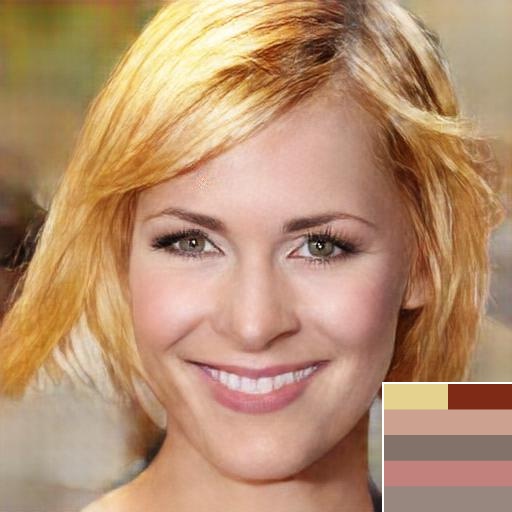}
\includegraphics[width=0.12\columnwidth]{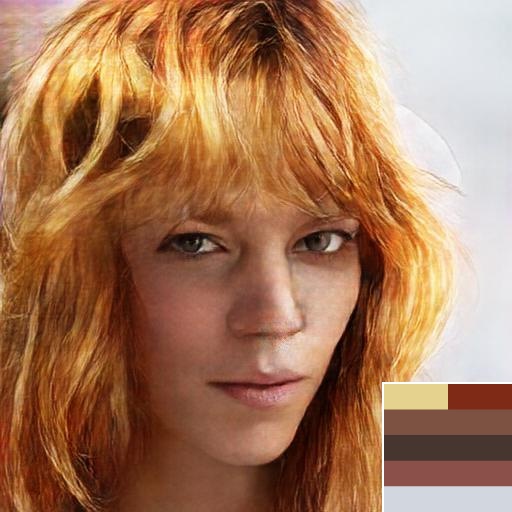}
\includegraphics[width=0.12\columnwidth]{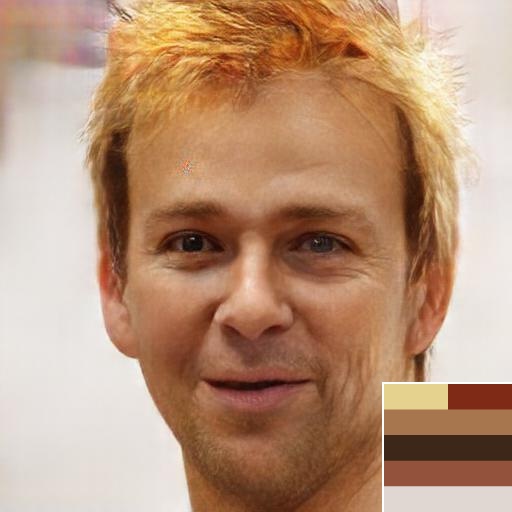}
\\
\includegraphics[width=0.12\columnwidth]{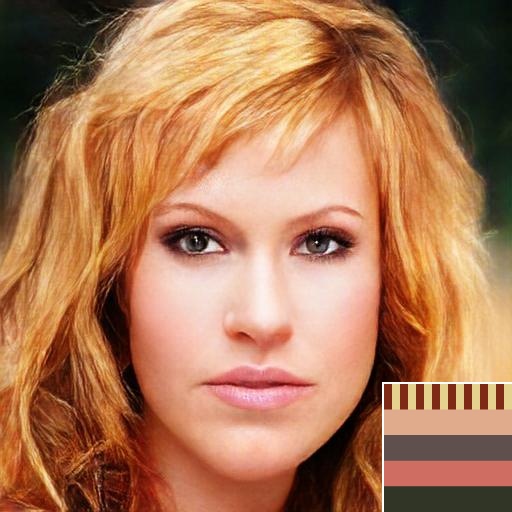}
\includegraphics[width=0.12\columnwidth]{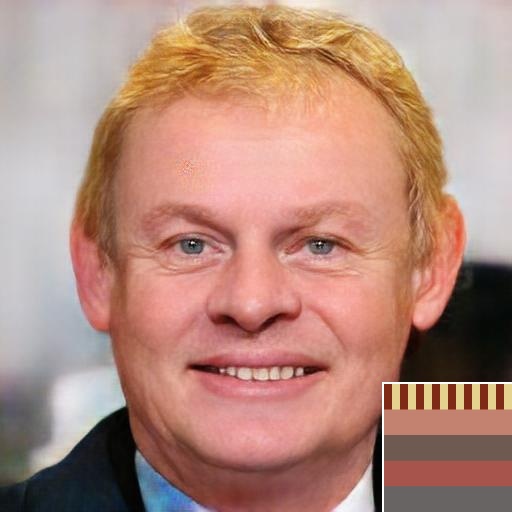}
\includegraphics[width=0.12\columnwidth]{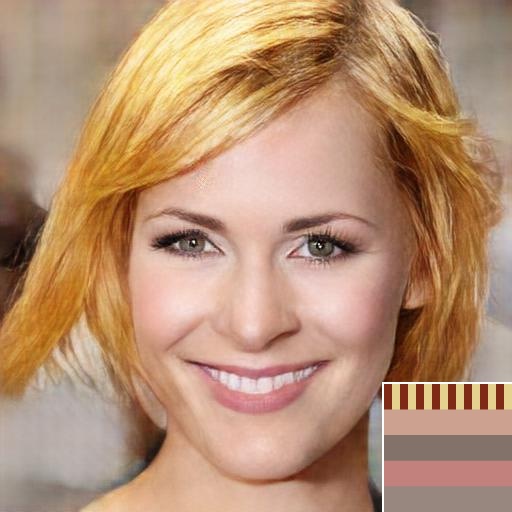}
\includegraphics[width=0.12\columnwidth]{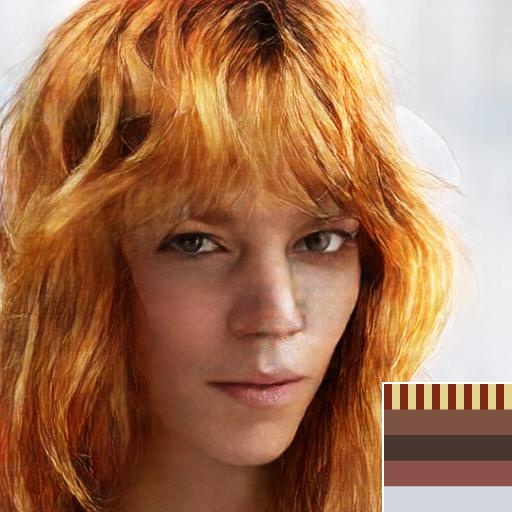}
\includegraphics[width=0.12\columnwidth]{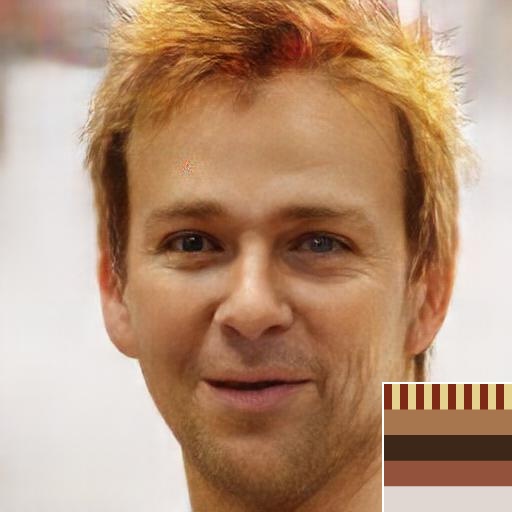}
\\

}
\caption{Blending colors using sliders (top) and vertical stripe patterns (bottom). Both control can yield similar results.
}
\label{fig:exp_color_edit_layout}
\end{figure}

Besides, we also test the color combination methods. As presented in Fig.~\ref{fig:exp_color_edit_layout}, we compare slider and vertical stripe pattern for hair color control. When we adjust the ratio of the two colors to $1:1$, the generated hair colors are very close. Moreover, the results do not demonstrate obvious pattern correlated with the combination method. For example, we do not observe stripe in the generated image when using vertical stripe pattern. Compared with the vertical stripe pattern method, slider is more intuitive and convenient to use for end users.

\begin{figure}[t!]
\centering
{\footnotesize
\makebox[0.12\columnwidth]{}
\includegraphics[width=0.12\columnwidth]{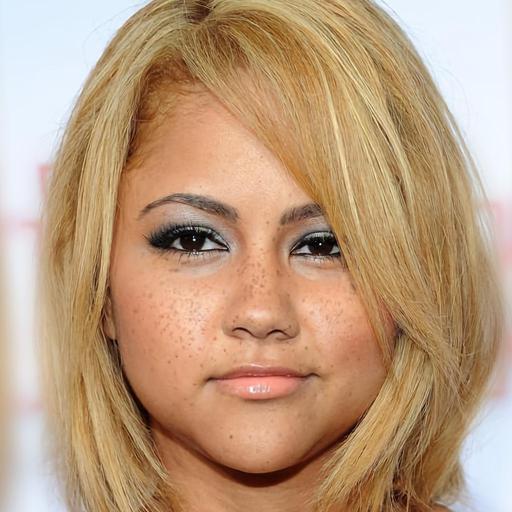}
\includegraphics[width=0.12\columnwidth]{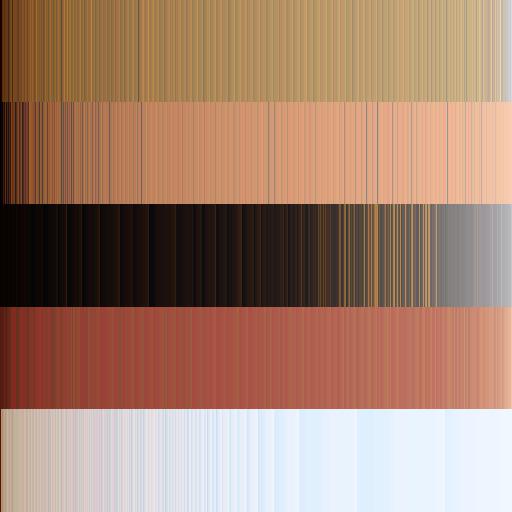}
\includegraphics[width=0.12\columnwidth]{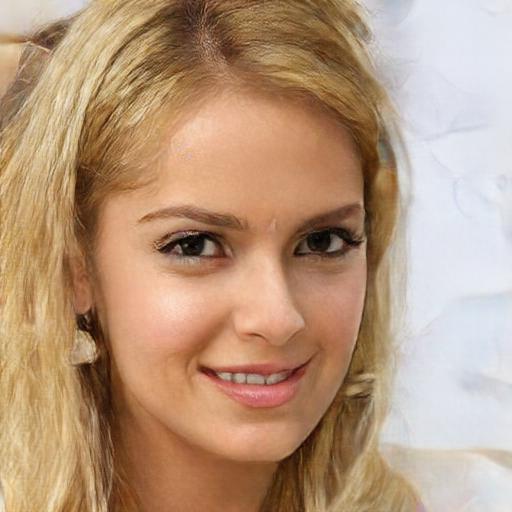}
\\
\includegraphics[width=0.12\columnwidth]{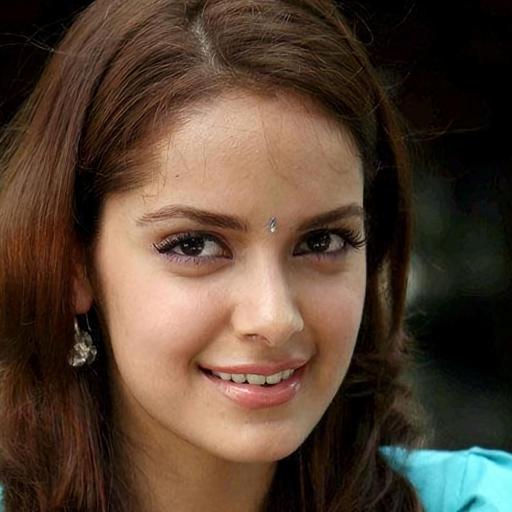}
\includegraphics[width=0.12\columnwidth]{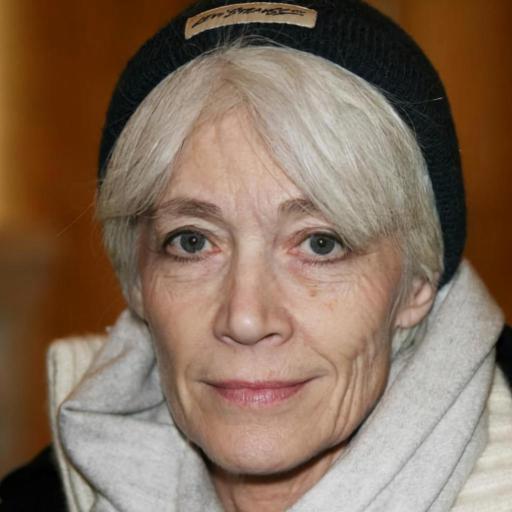}
\includegraphics[width=0.12\columnwidth]{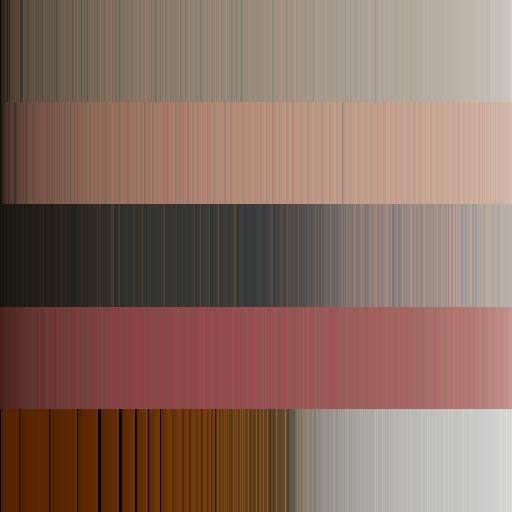}
\includegraphics[width=0.12\columnwidth]{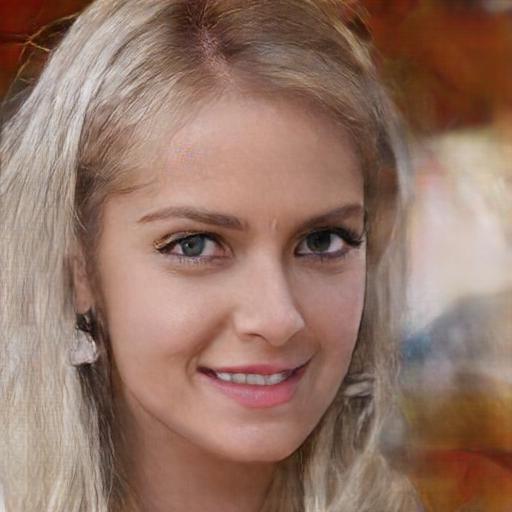}
\\
\makebox[0.12\columnwidth]{}
\includegraphics[width=0.12\columnwidth]{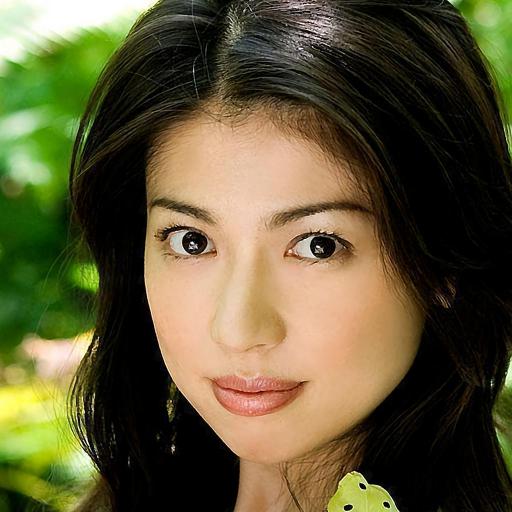}
\includegraphics[width=0.12\columnwidth]{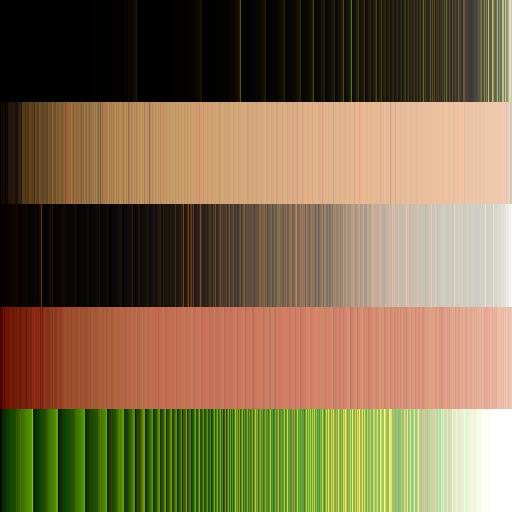}
\includegraphics[width=0.12\columnwidth]{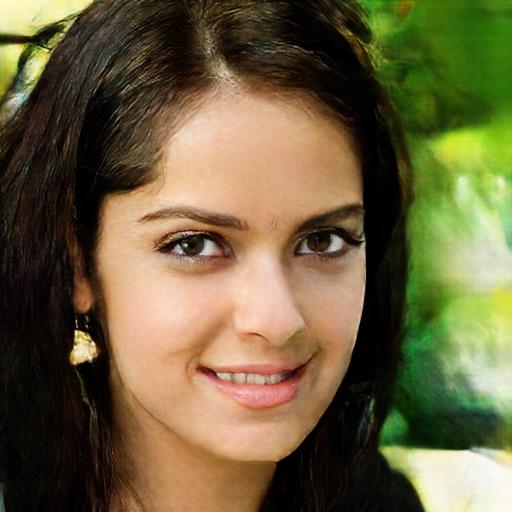}
\\
\makebox[0.12\columnwidth]{Original}
\makebox[0.12\columnwidth]{Reference}
\makebox[0.12\columnwidth]{Color Palette}
\makebox[0.12\columnwidth]{Result}
\\
}
\caption{Color transfer via modifying the color palette. See the main text for details.}
\label{fig:exp_color_transfer_dist}
\end{figure}

In Fig.~\ref{fig:exp_color_transfer_dist}, we demonstrate that our model can also be trained for color transfer. In stead of computing the average color as described in \S\ref{sec:color_palette}, we use color palette to represent the color distributions of each facial component from the reference images. More specifically, we sort the color pixels extracted from each facial component, and uniformly sample colors to create the new color palette, where each 1-pixel-width column represent one sampled color. Then we use the new color palette as conditional information for color transfer after model training. As demonstrated in Fig.~\ref{fig:exp_color_transfer_dist}, our method ensures color transfer from the same facial component of the reference image (i.e., prevents transfer from background to hair, skin, etc.), while most of the previous transfer based methods do not have such explicit restriction and hence are less controllable.

\begin{figure*}[t!]
\centering
{\footnotesize
\includegraphics[width=0.10\columnwidth]{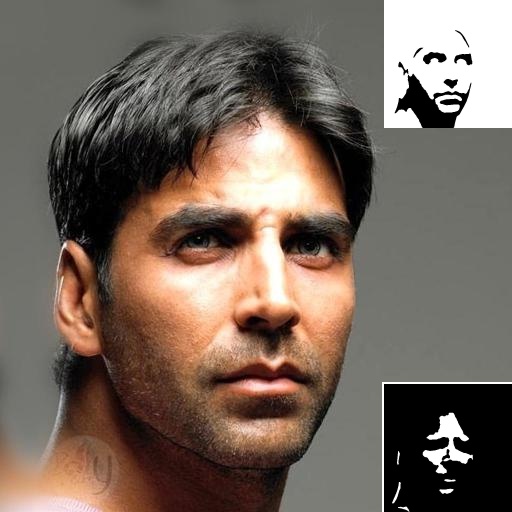}
\hspace{2pt}
\includegraphics[width=0.10\columnwidth]{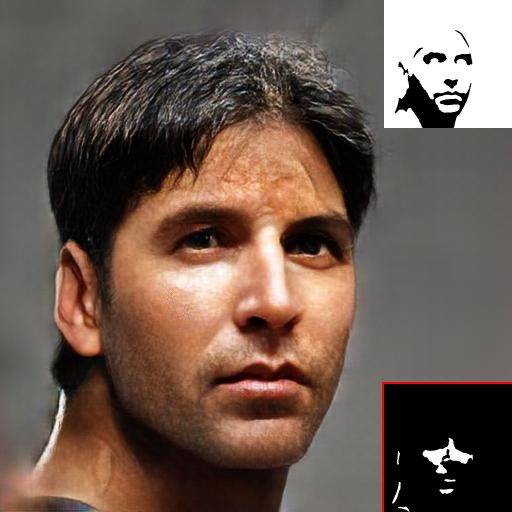} \hspace{-5pt}
\includegraphics[width=0.10\columnwidth]{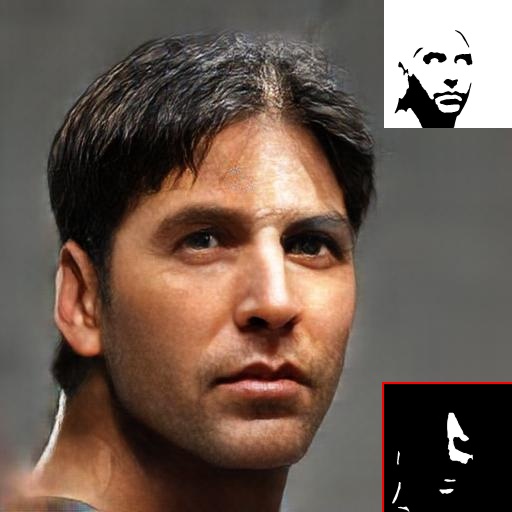} \hspace{-5pt}
\includegraphics[width=0.10\columnwidth]{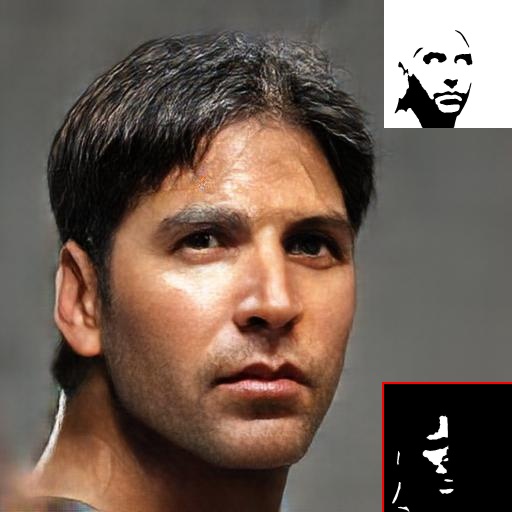}
\hspace{2pt}
\includegraphics[width=0.10\columnwidth]{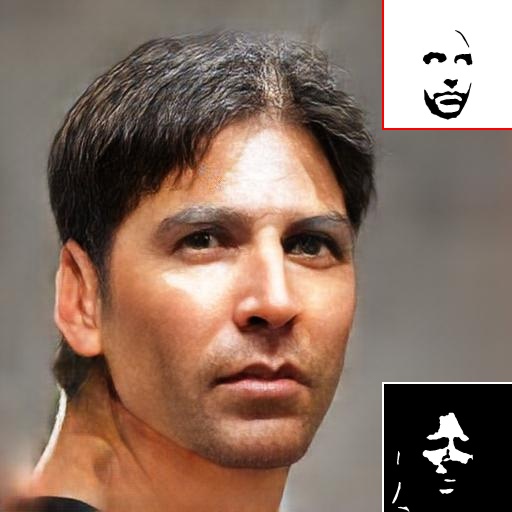} \hspace{-5pt}
\includegraphics[width=0.10\columnwidth]{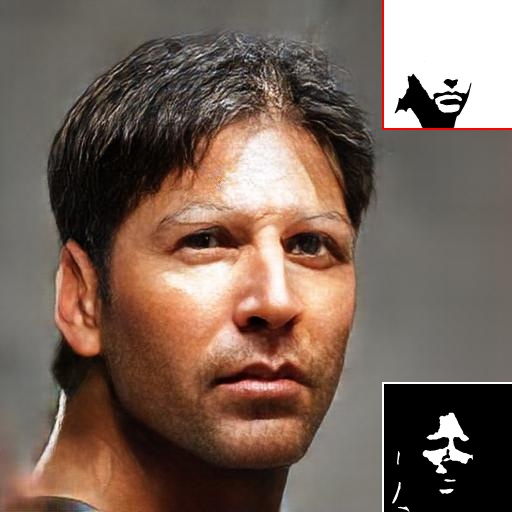}
\hspace{2pt}
\includegraphics[width=0.10\columnwidth]{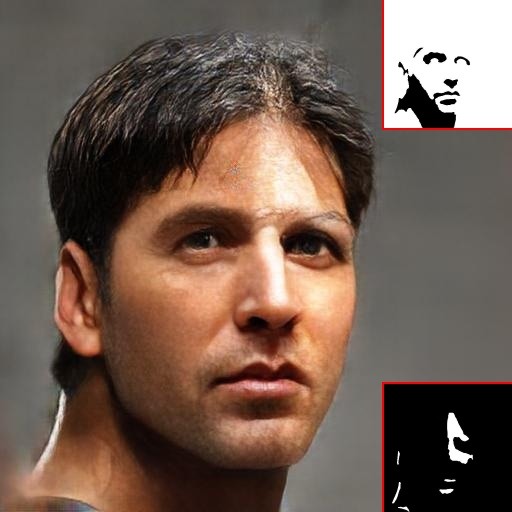} \\
\includegraphics[width=0.10\columnwidth]{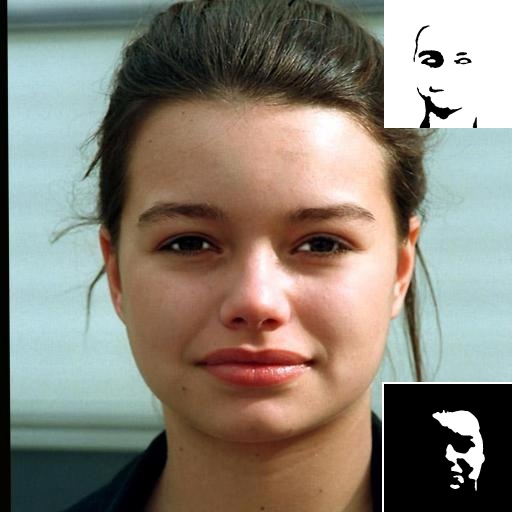}
\hspace{2pt}
\includegraphics[width=0.10\columnwidth]{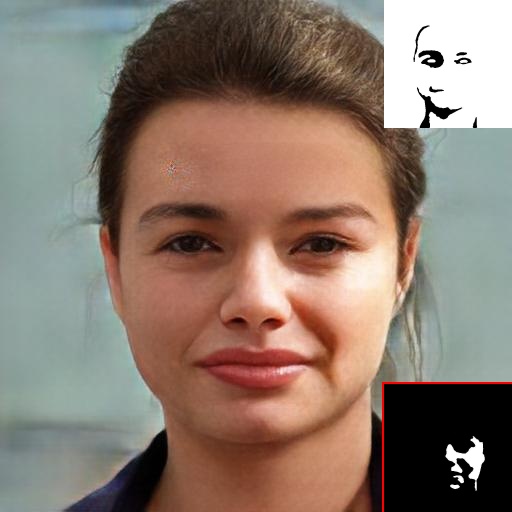} \hspace{-5pt}
\includegraphics[width=0.10\columnwidth]{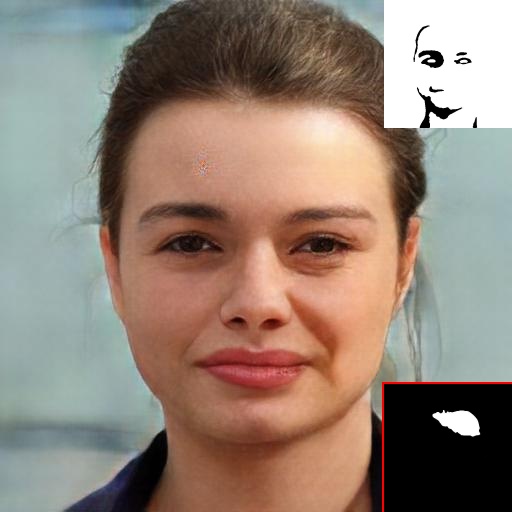} \hspace{-5pt} 
\includegraphics[width=0.10\columnwidth]{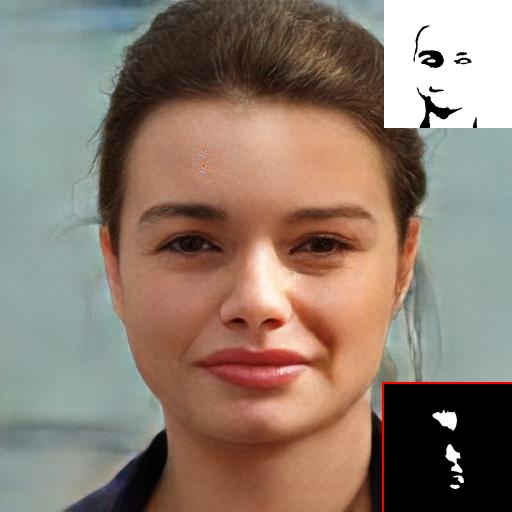}
\hspace{2pt}
\includegraphics[width=0.10\columnwidth]{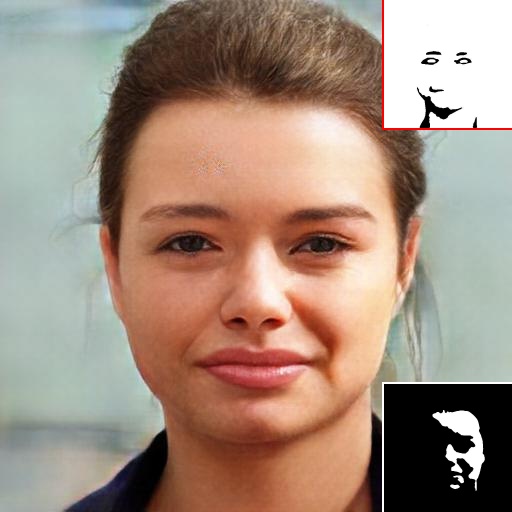} \hspace{-5pt}
\includegraphics[width=0.10\columnwidth]{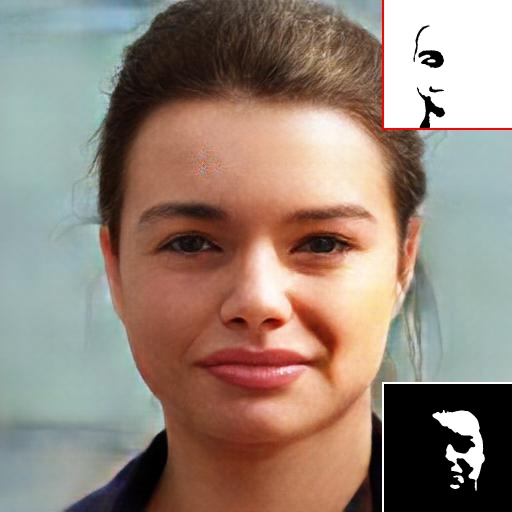}
\hspace{2pt}
\includegraphics[width=0.10\columnwidth]{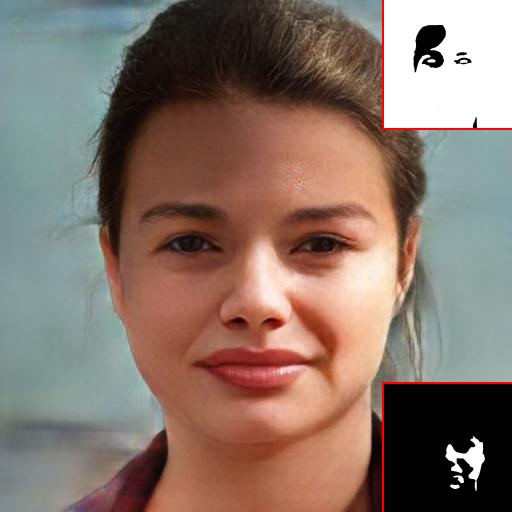} \\
\includegraphics[width=0.10\columnwidth]{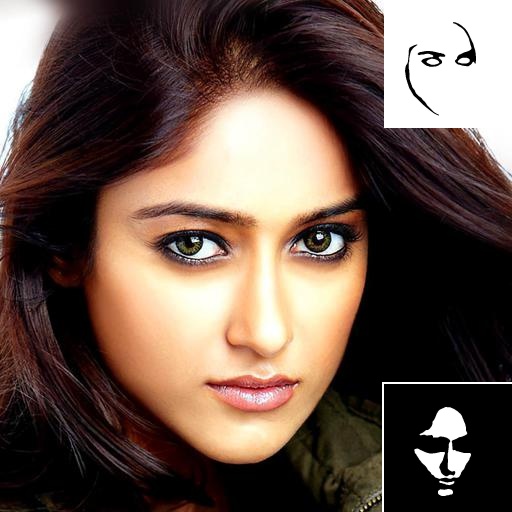}
\hspace{2pt}
\includegraphics[width=0.10\columnwidth]{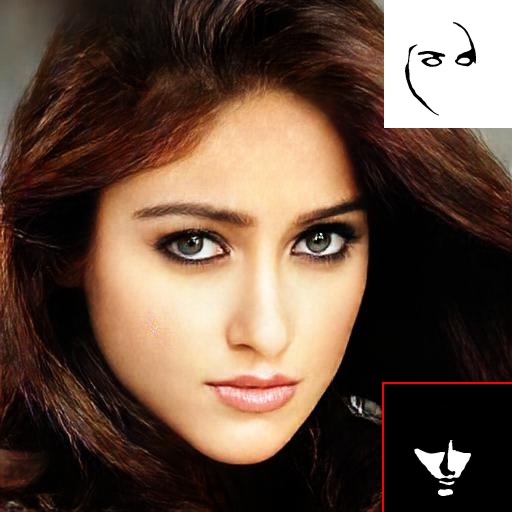} \hspace{-5pt}
\includegraphics[width=0.10\columnwidth]{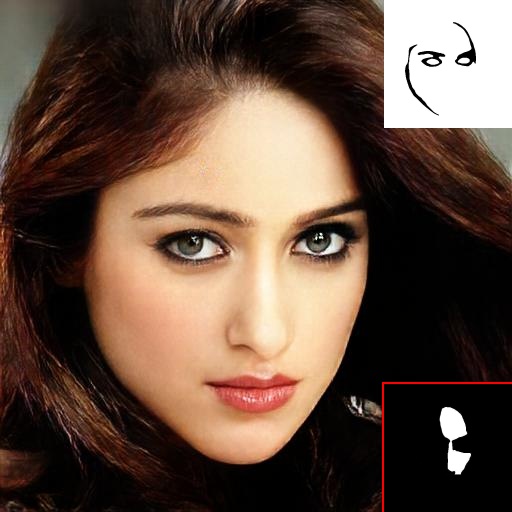} \hspace{-5pt}
\includegraphics[width=0.10\columnwidth]{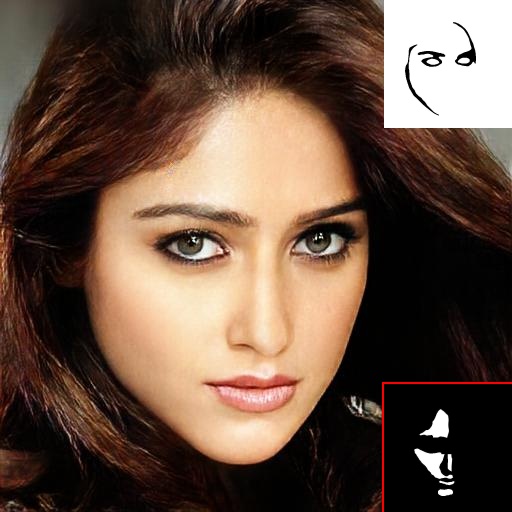}
\hspace{2pt}
\includegraphics[width=0.10\columnwidth]{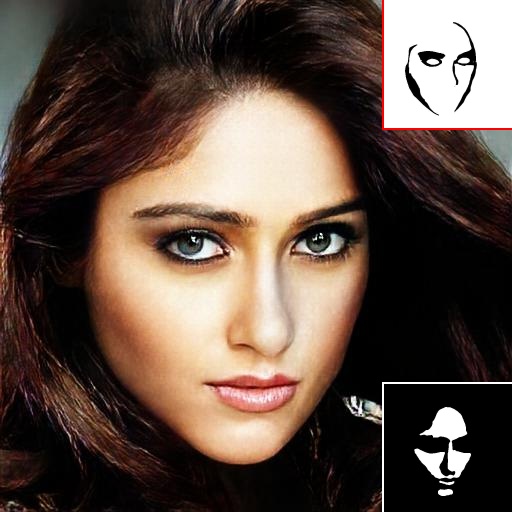} \hspace{-5pt}
\includegraphics[width=0.10\columnwidth]{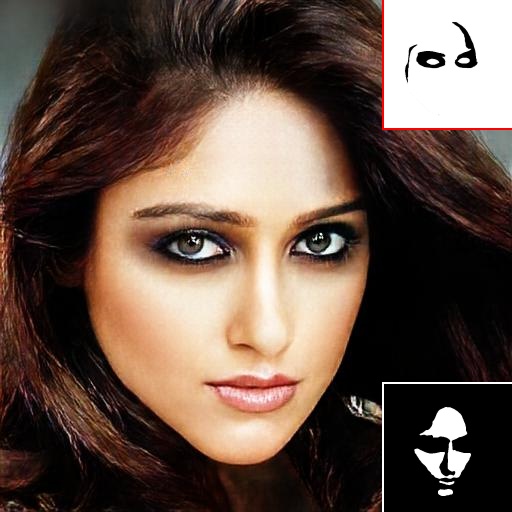}
\hspace{2pt}
\includegraphics[width=0.10\columnwidth]{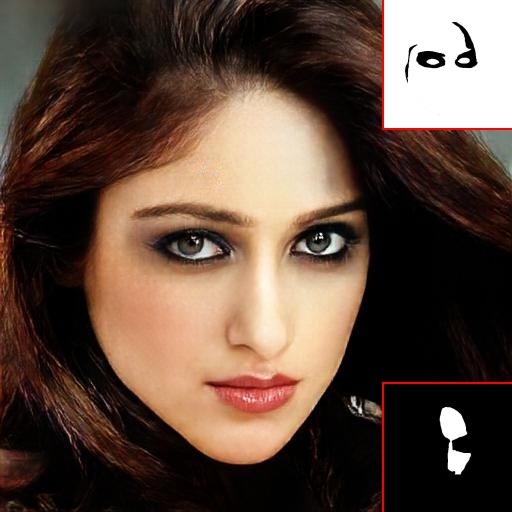} \\
\includegraphics[width=0.10\columnwidth]{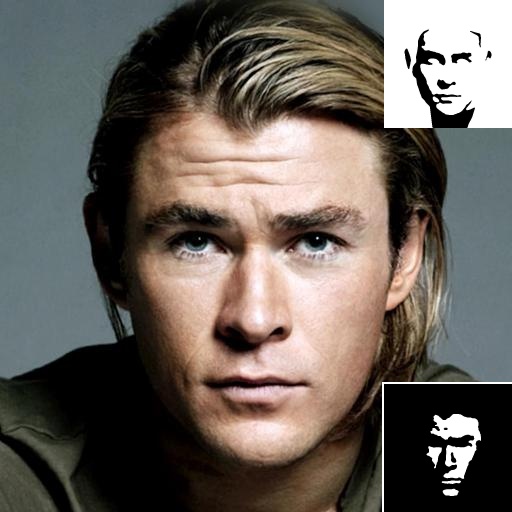}
\hspace{2pt}
\includegraphics[width=0.10\columnwidth]{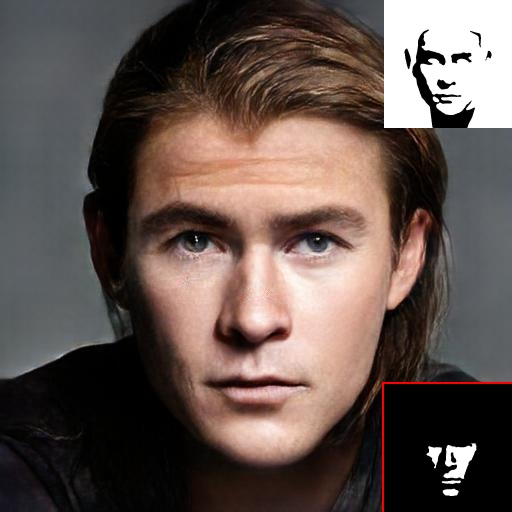} \hspace{-5pt}
\includegraphics[width=0.10\columnwidth]{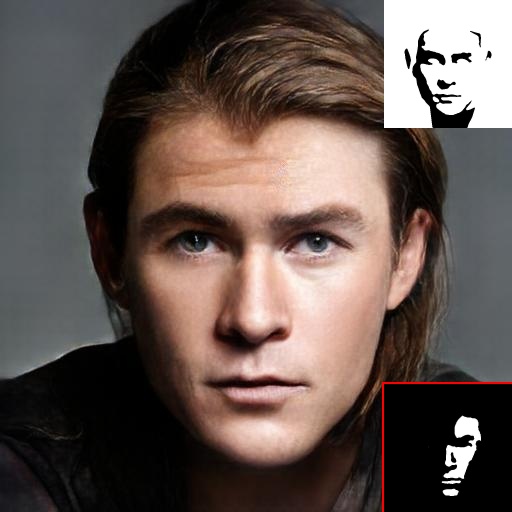} \hspace{-5pt}
\includegraphics[width=0.10\columnwidth]{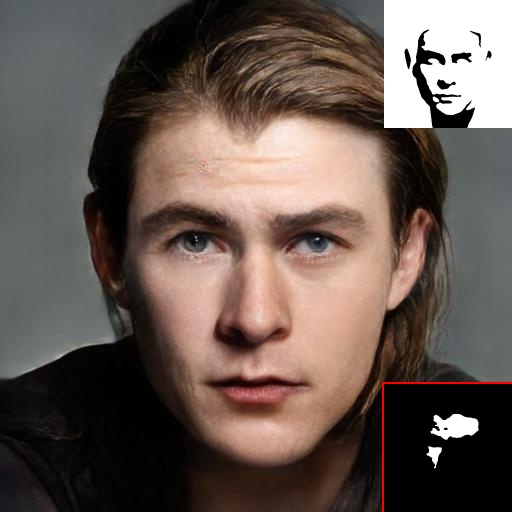}
\hspace{2pt}
\includegraphics[width=0.10\columnwidth]{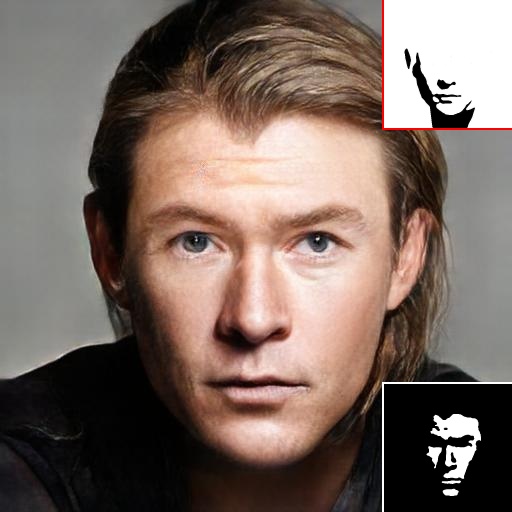} \hspace{-5pt}
\includegraphics[width=0.10\columnwidth]{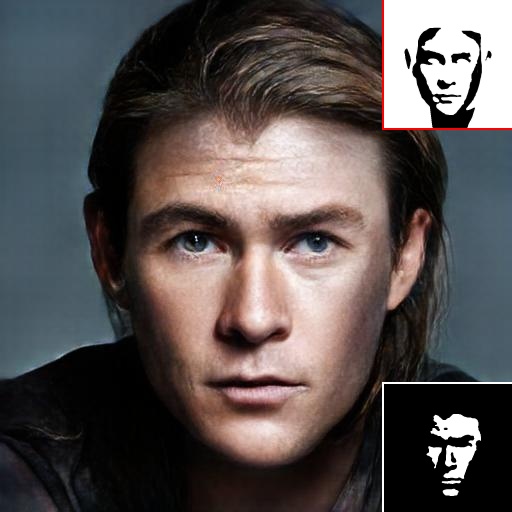}
\hspace{2pt}
\includegraphics[width=0.10\columnwidth]{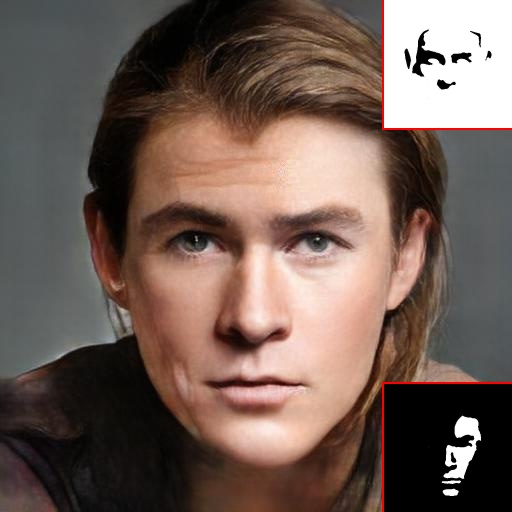} \\
%\includegraphics[width=0.10\columnwidth]{figs/hl_12398_r_real_c.jpg}
%\hspace{2pt}
%\includegraphics[width=0.10\columnwidth]{figs/hl_12398_r_fake_c.jpg} \hspace{-5pt}
%\includegraphics[width=0.10\columnwidth]{figs/hl_12398_r_fake2_c.jpg} \hspace{-5pt}
%\includegraphics[width=0.10\columnwidth]{figs/hl_12398_r_fake3_c.jpg}
%\hspace{2pt}
%\includegraphics[width=0.10\columnwidth]{figs/sd_12398_r_fake_c.jpg} \hspace{-5pt}
%\includegraphics[width=0.10\columnwidth]{figs/sd_12398_r_fake2_c.jpg} 
%\hspace{2pt}
%\includegraphics[width=0.10\columnwidth]{figs/ls_12398_r_fake_c.jpg} \\
\includegraphics[width=0.10\columnwidth]{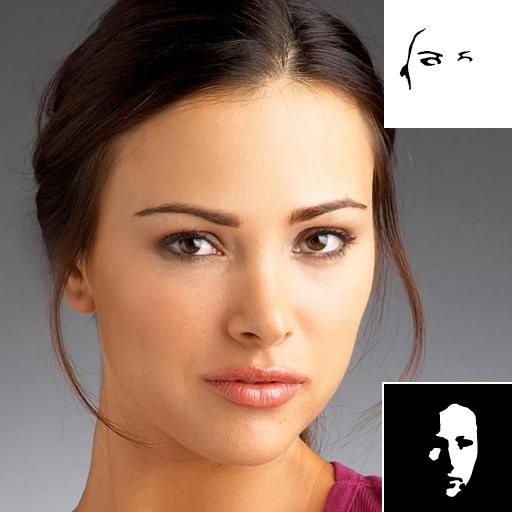}
\hspace{2pt}
\includegraphics[width=0.10\columnwidth]{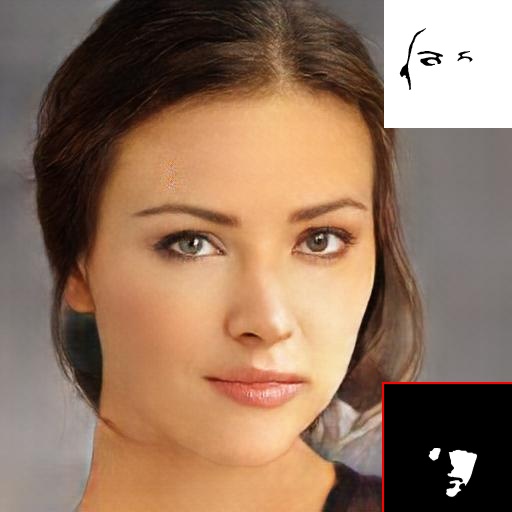} \hspace{-5pt}
\includegraphics[width=0.10\columnwidth]{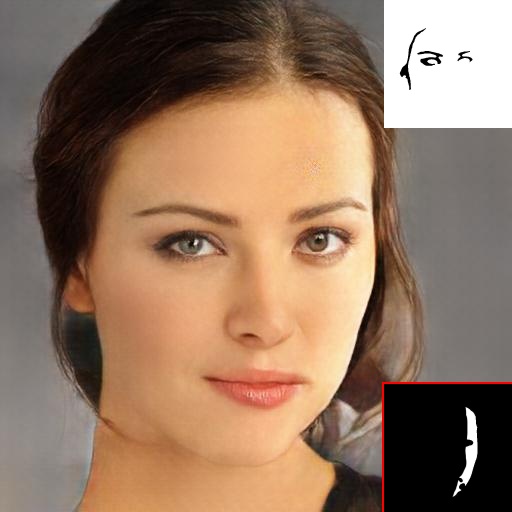} \hspace{-5pt}
\includegraphics[width=0.10\columnwidth]{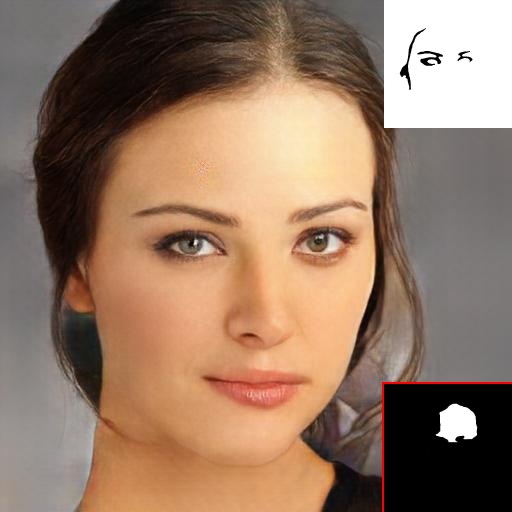}
\hspace{2pt}
\includegraphics[width=0.10\columnwidth]{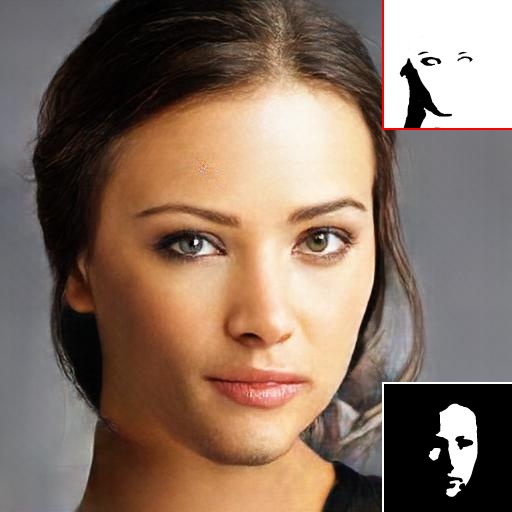} \hspace{-5pt}
\includegraphics[width=0.10\columnwidth]{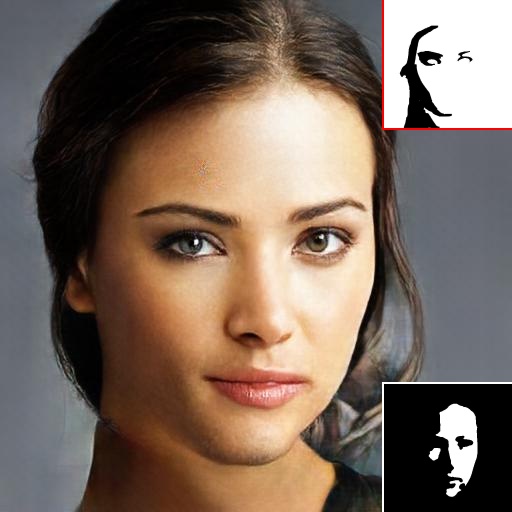}
\hspace{2pt}
\includegraphics[width=0.10\columnwidth]{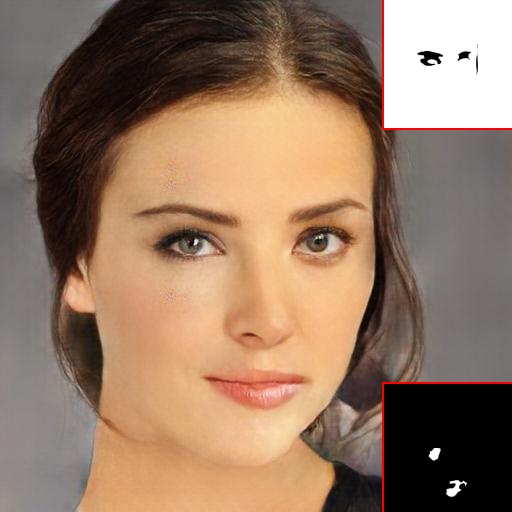} \\
\makebox[0.10\columnwidth]{Original}
\hspace{2pt}
\makebox[0.10\columnwidth]{Light \#1} \hspace{-5pt}
\makebox[0.10\columnwidth]{Light \#2} \hspace{-5pt}
\makebox[0.10\columnwidth]{Light \#3}
\hspace{2pt}
\makebox[0.10\columnwidth]{Shadow \#1} \hspace{-5pt}
\makebox[0.10\columnwidth]{Shadow \#2}
\hspace{2pt}
\makebox[0.10\columnwidth]{Light + shadow}\\
}
\caption{Light and shadow editing via binary masks. We show the light and shadow masks in the bottom- and top-right corners, respectively, and  highlight the modified masks with red boundaries.}
\label{fig:exp_light}
\end{figure*}

\subsubsection{Lights \& Shadows}

\begin{figure*}[htbp!]
\centering
{\footnotesize
\includegraphics[width=0.12\columnwidth]{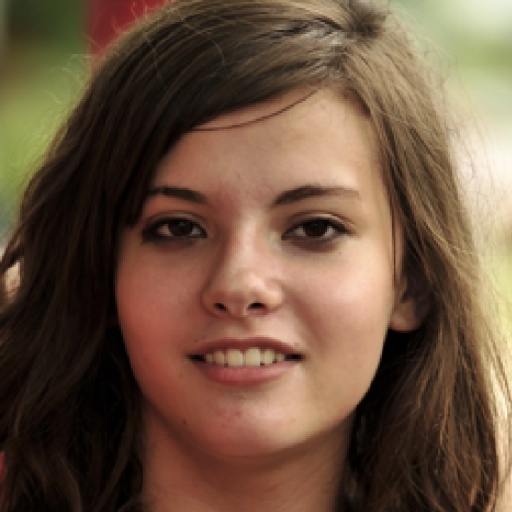}
\includegraphics[width=0.12\columnwidth]{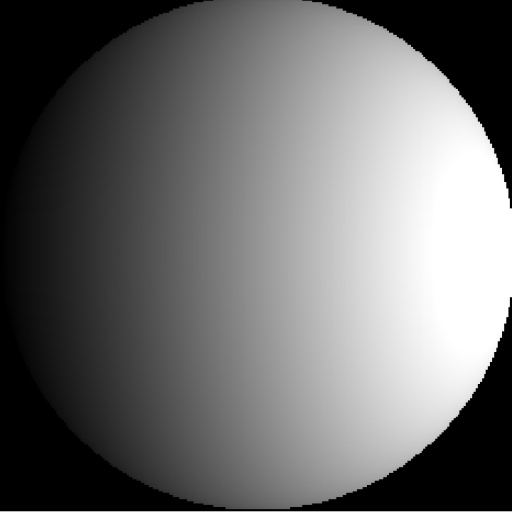}
\hspace{2pt}
\includegraphics[width=0.12\columnwidth]{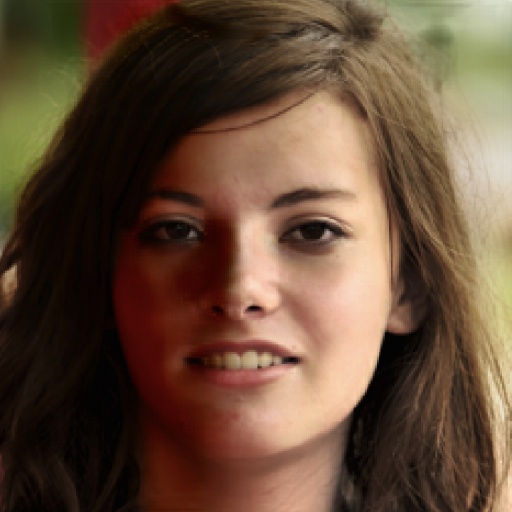}
\includegraphics[width=0.12\columnwidth]{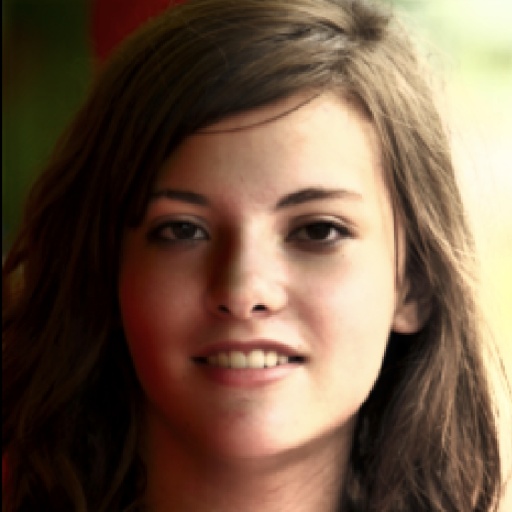}
\includegraphics[width=0.12\columnwidth]{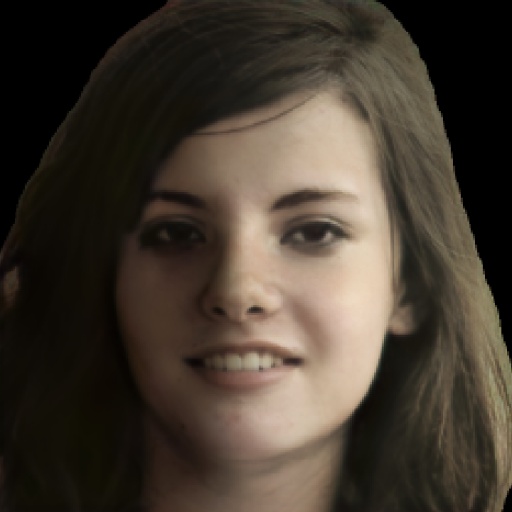}
\includegraphics[width=0.12\columnwidth]{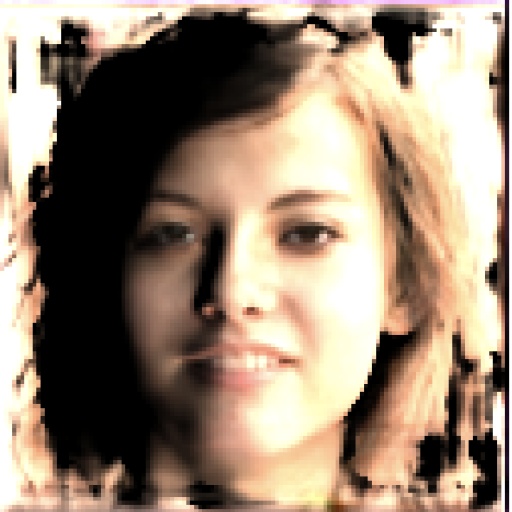}
\hspace{2pt}
\includegraphics[width=0.06\columnwidth]{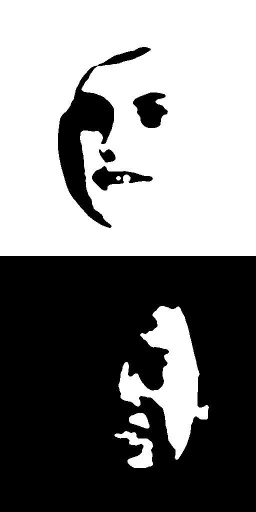}
\includegraphics[width=0.12\columnwidth]{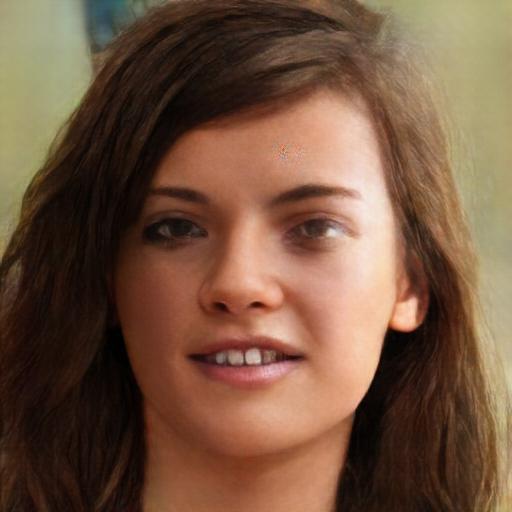}
\\
\includegraphics[width=0.12\columnwidth]{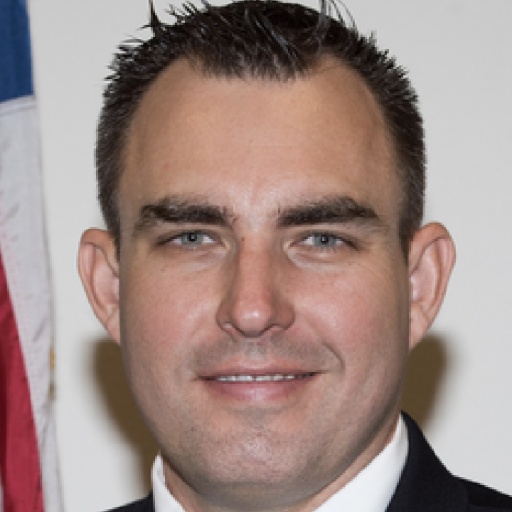}
\includegraphics[width=0.12\columnwidth]{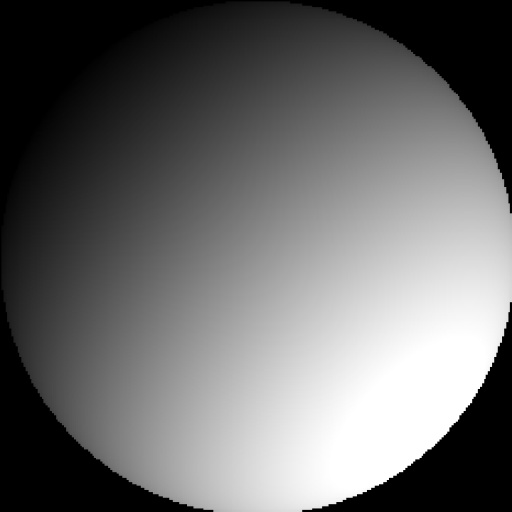}
\hspace{2pt}
\includegraphics[width=0.12\columnwidth]{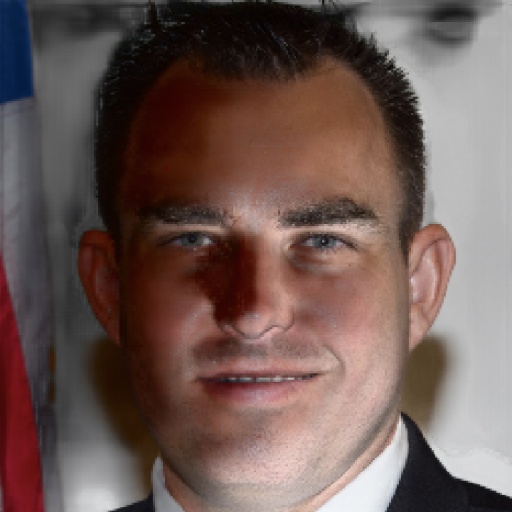}
\includegraphics[width=0.12\columnwidth]{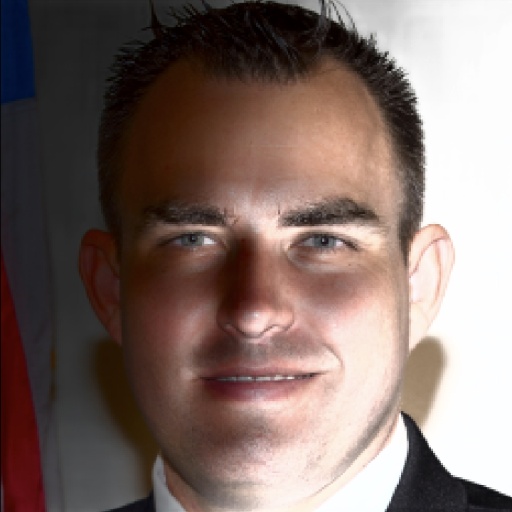}
\includegraphics[width=0.12\columnwidth]{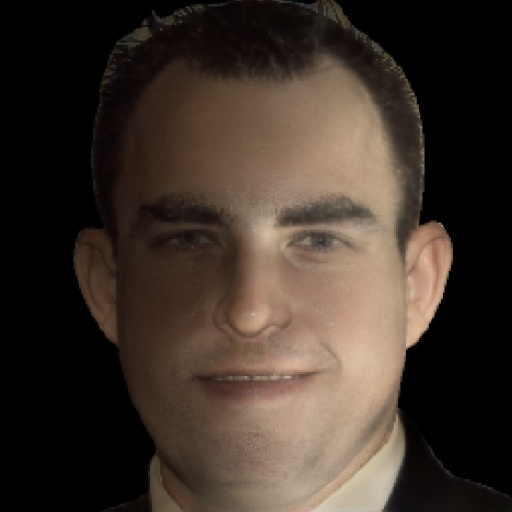}
\includegraphics[width=0.12\columnwidth]{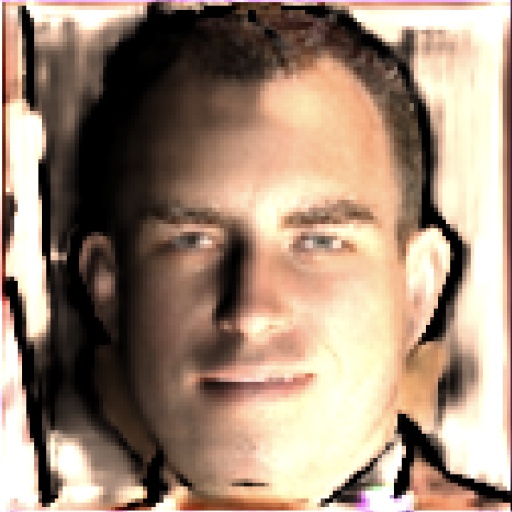}
\hspace{2pt}
\includegraphics[width=0.06\columnwidth]{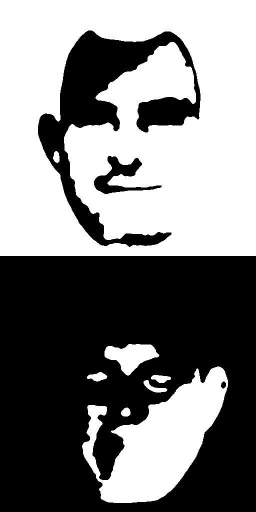}
\includegraphics[width=0.12\columnwidth]{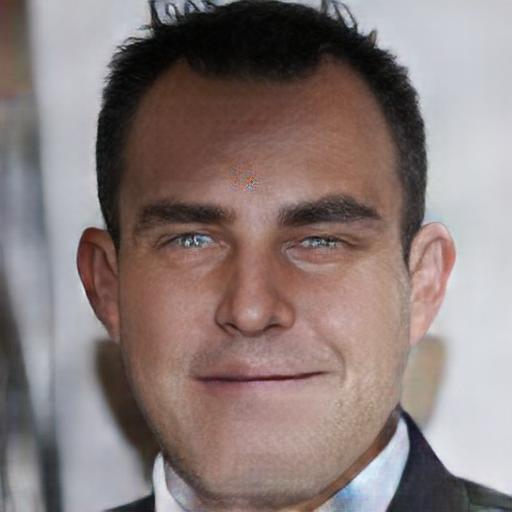}
\\
\includegraphics[width=0.12\columnwidth]{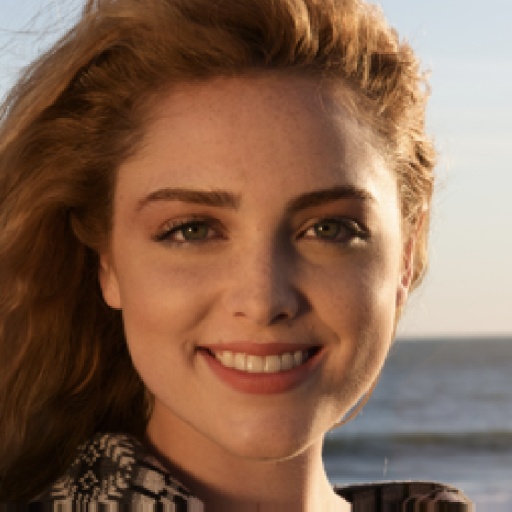}
\includegraphics[width=0.12\columnwidth]{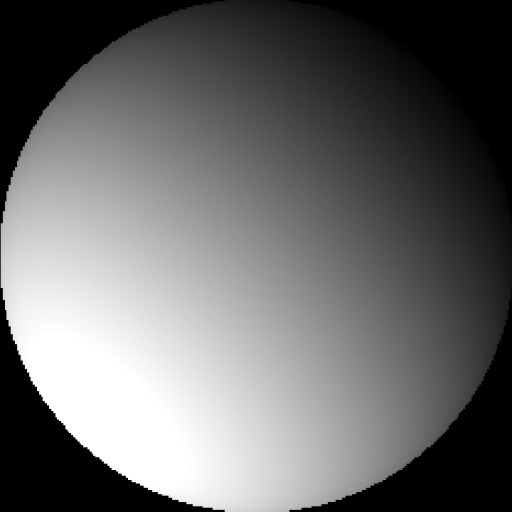}
\hspace{2pt}
\includegraphics[width=0.12\columnwidth]{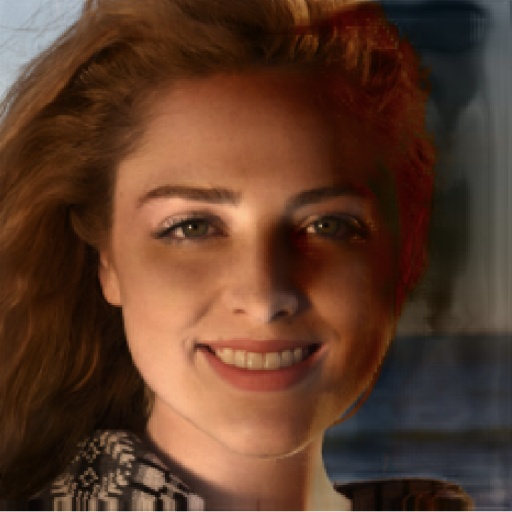}
\includegraphics[width=0.12\columnwidth]{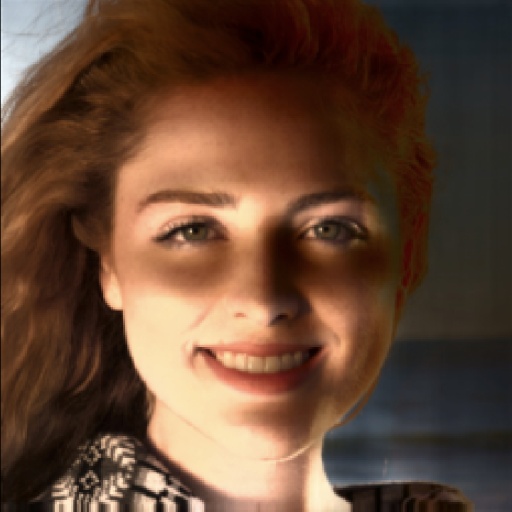}
\includegraphics[width=0.12\columnwidth]{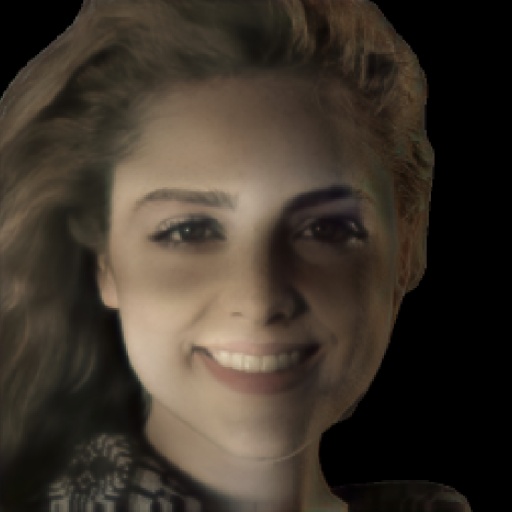}
\includegraphics[width=0.12\columnwidth]{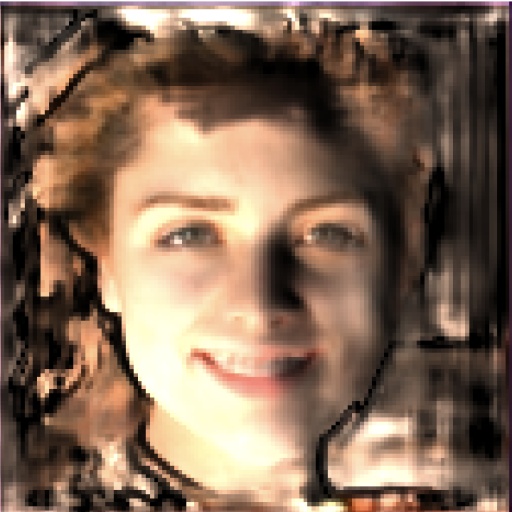}
\hspace{2pt}
\includegraphics[width=0.06\columnwidth]{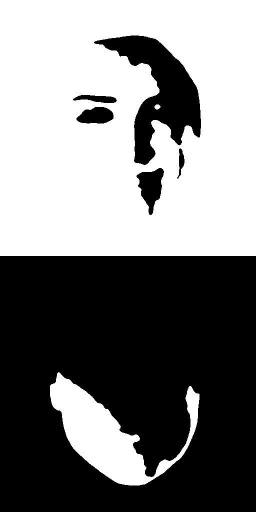}
\includegraphics[width=0.12\columnwidth]{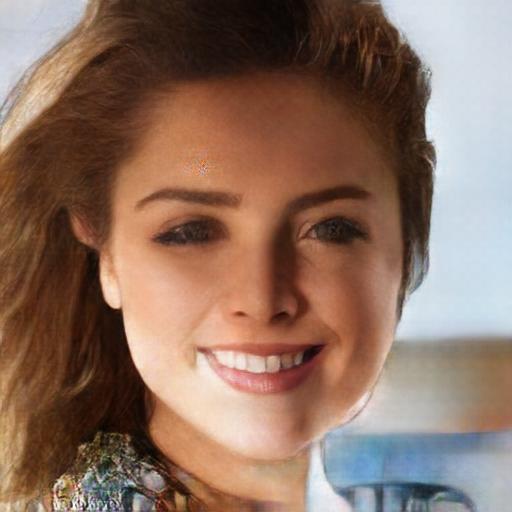}
\\
\includegraphics[width=0.12\columnwidth]{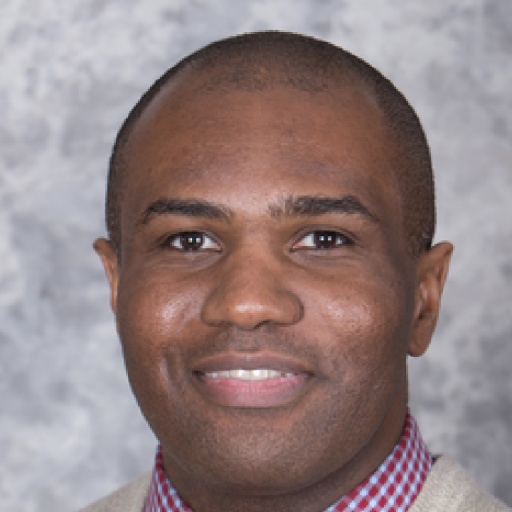}
\includegraphics[width=0.12\columnwidth]{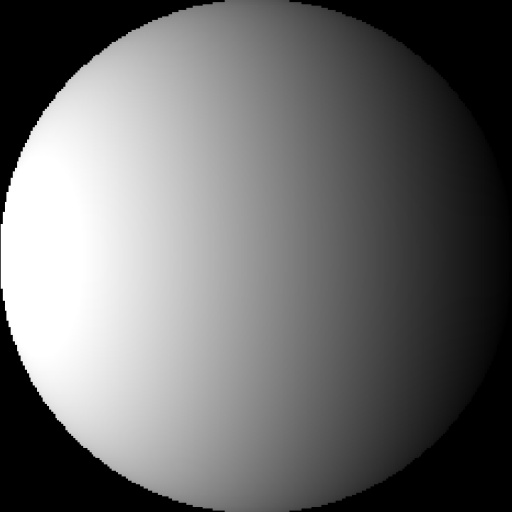}
\hspace{2pt}
\includegraphics[width=0.12\columnwidth]{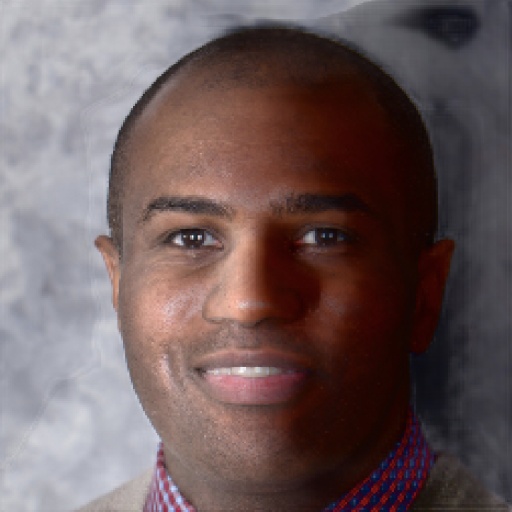}
\includegraphics[width=0.12\columnwidth]{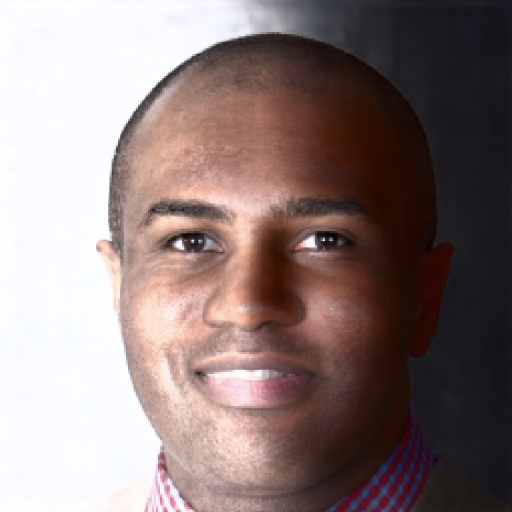}
\includegraphics[width=0.12\columnwidth]{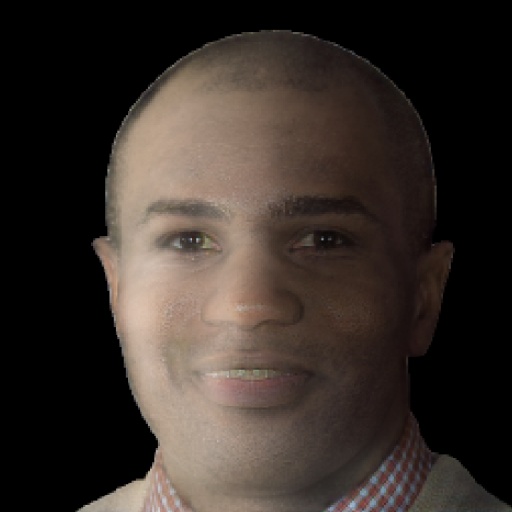}
\includegraphics[width=0.12\columnwidth]{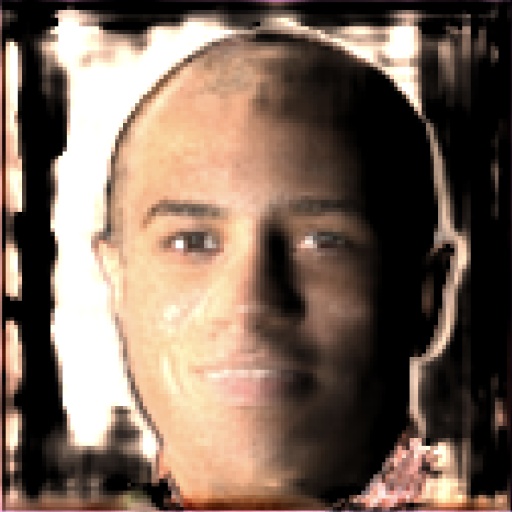}
\hspace{2pt}
\includegraphics[width=0.06\columnwidth]{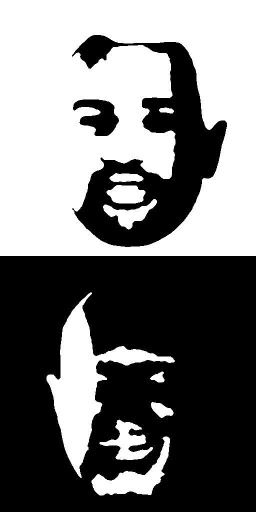}
\includegraphics[width=0.12\columnwidth]{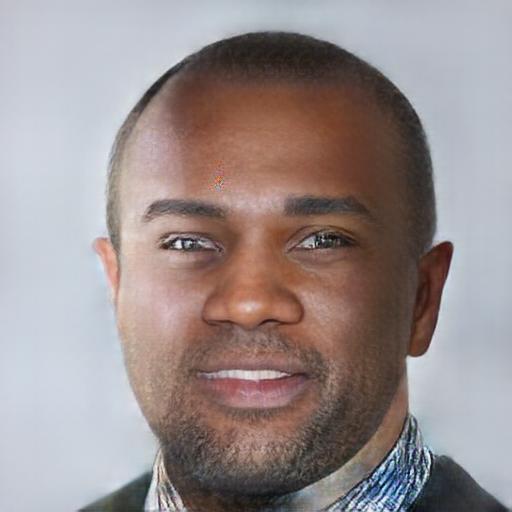}
\\

\makebox[0.12\columnwidth]{Original}
\makebox[0.12\columnwidth]{Target}
\hspace{2pt}
\makebox[0.12\columnwidth]{\citet{HouZSBT021}}
\makebox[0.12\columnwidth]{\citet{zhou2019deep}}
\makebox[0.12\columnwidth]{\citet{sun2019single}}
\makebox[0.12\columnwidth]{\citet{sengupta2018sfsnet}}
\hspace{2pt}
\makebox[0.06\columnwidth]{Masks}
\makebox[0.12\columnwidth]{Ours}
\\
}
\caption{Comparison of light and shadow editing with previous methods. The masks used in our method are computed automatically from the user-specified light directions. }
\label{fig:exp_light_compare}
\end{figure*}

Our model can also be used for fine-grained light and shadow editing through the modification of the binary light and shadow masks. As the examples show in Fig.~\ref{fig:exp_light}, it is convenient to add or remove shadow/light to the desired area with simple mask modification operations. Our model is feed forward, so it can be easily implemented for interactive editing. In Fig.~\ref{fig:exp_light_compare}, we also compare the images generated using our model and a few existing portrait relighting methods. Our method automatically generates shadow and light masks (second-to-last column), which are binary images, and feed them into our neural network model. We observe that our method can yield visually pleasing results comparable to the previous methods, while providing users with a direct way to edit light and shadow. 

In our implementation, we apply a simple method to extract light and shadow masks from the user-specified light image (second column in Fig.~\ref{fig:exp_light_compare}) and the original portrait images, and then generate the final masks through the \textit{AND}/\textit{OR} Boolean operations. Note that it is also possible to build 3D face models \citep{Tran_2018_CVPR,Wu_2019_CVPR} from which high-quality light and shadow masks can be generated. We leave it as a future work.

\subsubsection{Portrait Artworks}
\begin{figure}[t!]
\centering
{\footnotesize
\includegraphics[width=0.192\columnwidth]{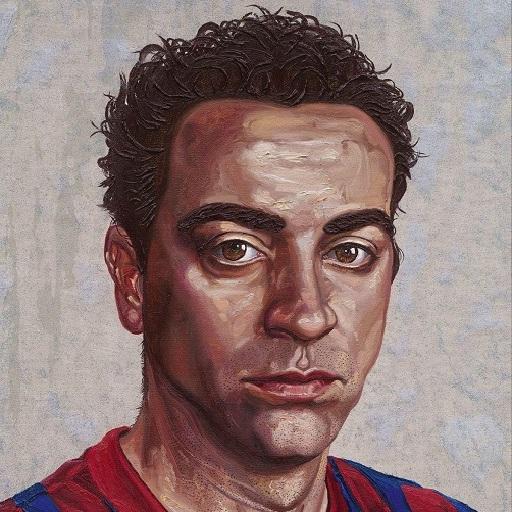}
\includegraphics[width=0.192\columnwidth]{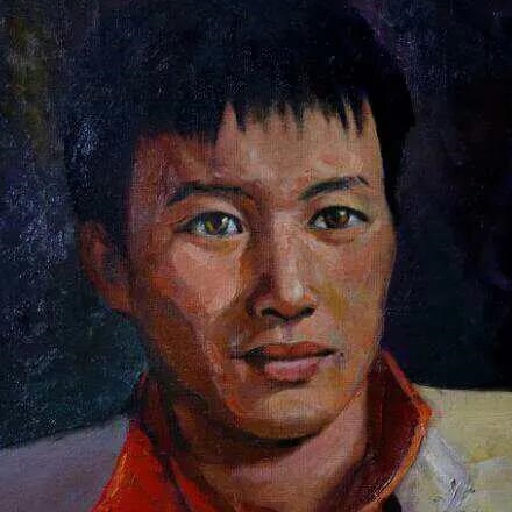}
\includegraphics[width=0.192\columnwidth]{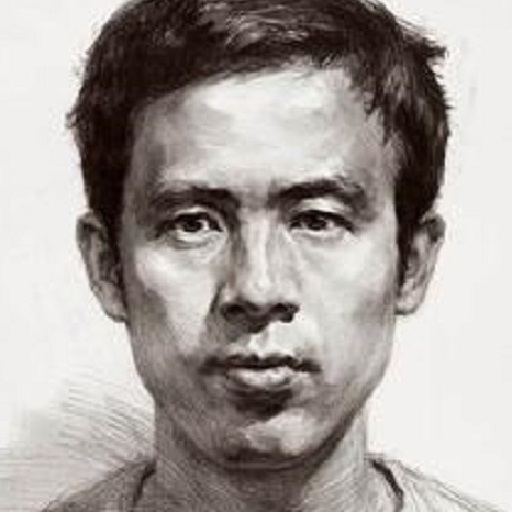}
\includegraphics[width=0.192\columnwidth]{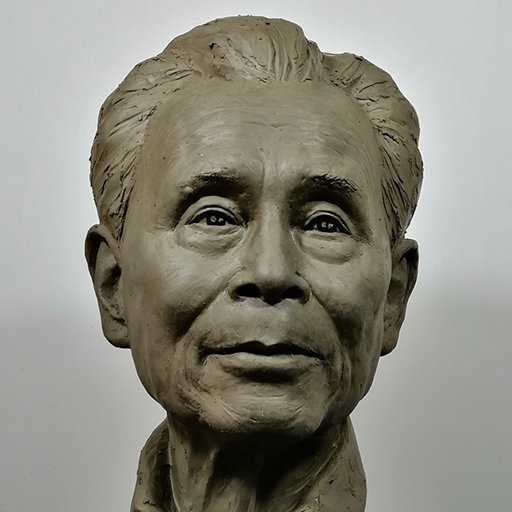}
\includegraphics[width=0.192\columnwidth]{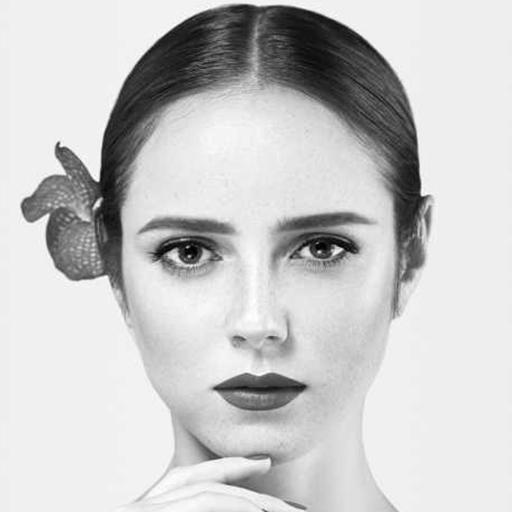}\\
\vspace{2pt}
\includegraphics[width=0.192\columnwidth]{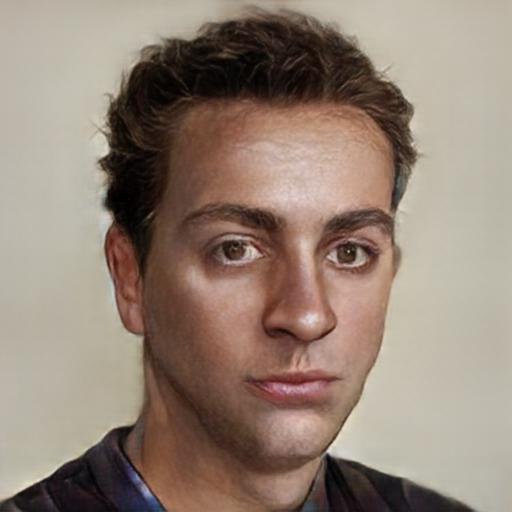}
\includegraphics[width=0.192\columnwidth]{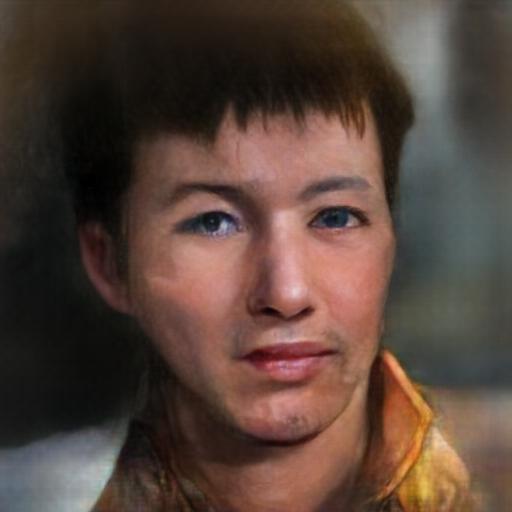}
\includegraphics[width=0.192\columnwidth]{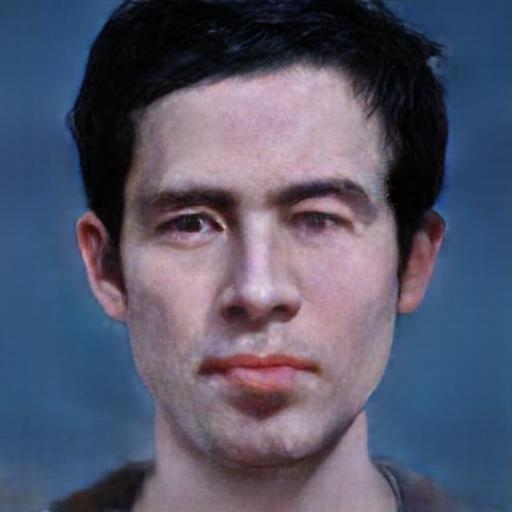}
\includegraphics[width=0.192\columnwidth]{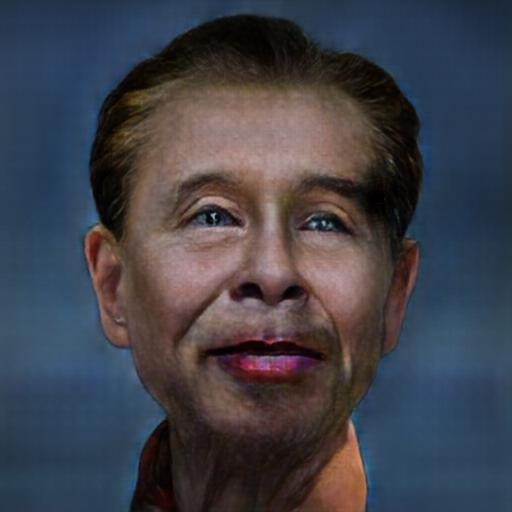}
\includegraphics[width=0.192\columnwidth]{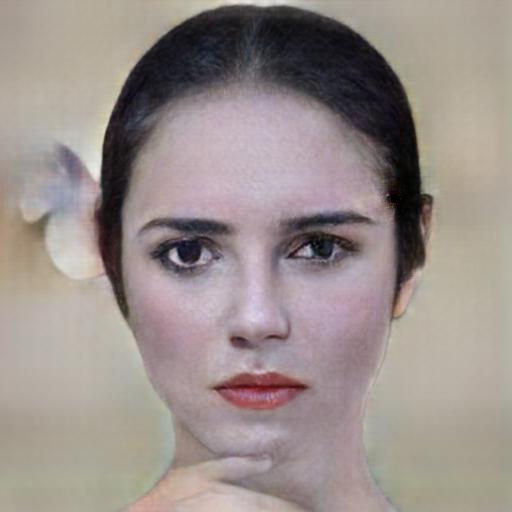}\\
\makebox[0.192\columnwidth]{(a)}
\makebox[0.192\columnwidth]{(b)}
\makebox[0.192\columnwidth]{(c)}
\makebox[0.192\columnwidth]{(d)}
\makebox[0.192\columnwidth]{(e)}\\
}
\caption{Our method can improve photo-realism of various types of portrait artworks. The upper row shows the input images and the bottom row is the generated results. (a) and (b) oil paintings; (c) pencil sketch; (d) sculpture; (e) colorization of black-and-white photo.}
\label{fig:artwork}
\end{figure}

Though the main objective of our method is for portrait photo editing, we observe that our method can also be applied to various types of portrait artworks. As illustrated in Fig.~\ref{fig:artwork}, our model can improve photo-realism of oil paintings, pencil sketches, sculptures and black-and-white photos. We generate the results  in two simple steps: 1) extracting edge maps, colors and shadow/light masks from the original image (\S\ref{sec:model_input_sec}); and 2) editing colors via a palette.

\subsection{Ablation Study}

\subsubsection{Asymmetric Conditional GAN}
\textbf{Fine-grained color control.} We first conduct experiments to quantitatively analyze the performance of the proposed asymmetric conditional GAN for fine-grained color control. We denote by \textbf{AC-GAN} our model trained with the asymmetric conditional GAN, which takes color palette $\mI_{CP}$ as the generator input, and average color map $\mI_{C}$ as the discriminators input. Denote by \textbf{C-GAN} our model trained with the original conditional GAN, which takes $\mI_{CP}$ as both generator and discriminator inputs.

The test split is used for evaluation to compare the images generated by \textbf{AC-GAN} and \textbf{C-GAN}. Instead of the original color palettes, we use the color palette of a randomly selected image $\mP_{rand}$ as $\mI_{CP}$ to control image colors. Besides, we also use the manually annotated segmentation masks to generate color palette $\mO_{CP}$ for each synthesised image, following the steps described in \S\ref{sec:color_palette}. For each facial component, we compute the Euclidean distance between their corresponding average colors in $\mI_{CP}$ and $\mO_{CP}$. Besides, we also compare color distribution between the corresponding facial components in each synthesised image and $\mP_{rand}$. Similar to \citet{afifi2021histogan}, we adopt the average of KL divergence to measure the difference between RGB color histograms in our experiments.

From the average color distances shown in Table~\ref{tb:ablation_color}, we can see that colors of the images generated by \textbf{AC-GAN} are closer than those of \textbf{C-GAN} to the desired ones on the color palette for most of the facial components. Especially for lip, hair and eyes, users are usually more interested in editing their color. The face colors generated by \textbf{C-GAN} is slightly closer to the input for color control, which is probably because of the higher color variance of faces, so the value for face may be less representative. Since KL divergence treat distant and neighbouring columns in the RGB color histograms indifferently, which does not directly measure the color distance, so we use it as a reference to compare color distributions. The results in Table~\ref{tb:ablation_color} show that the color distributions generated by \textbf{AC-GAN} are closer to that of the real images for 3 out of the 5 components.

\begin{table}[ht!]
\centering
\caption{Color control using the original and asymmetric conditional GAN.}
\begin{small}
\begin{tabular}{lccccc}
\toprule
 & \textbf{Hair} &  \textbf{Skin} &  \textbf{Eyes} & \textbf{Lip} & \textbf{BG.}  \\
\midrule
\textbf{Avg. Color Dist. } & & & \\
\textbf{C-GAN} & 41.92 & \textbf{41.24} & 50.22 & 50.12 & 42.71 \\
\textbf{AC-GAN} & \textbf{41.20} & 41.72 & \textbf{49.94} & \textbf{47.88} & \textbf{40.44} \\
\midrule
\textbf{Color Hist. Dist.} & & & \\
\textbf{C-GAN} & 1.1103 & \textbf{0.7294} & 0.5773 & \textbf{0.8494} & 2.3353 \\
\textbf{AC-GAN} & \textbf{1.0933} & 0.7309 & \textbf{0.5763} & 0.8711 & \textbf{2.3039} \\
\bottomrule
\end{tabular}
\end{small}
\label{tb:ablation_color}
\end{table}

\noindent\textbf{Robustness to noisy edge maps.} The asymmetric conditional GAN is also possibly useful to improve the model robustness to noisy edge maps. To verify this, we apply the proposed edge map noising methods $n(\cdot)$ to the edge maps in the training split, the use the asymmetric and original conditional GAN to train the model. Similar to the experiments above, the models are denoted with \textbf{AC-GAN} and \textbf{C-GAN}, respective. \textbf{AC-GAN} takes $n(\mI_{E})$ as generator input and $\mI_{E}$ as discriminators input, while \textbf{C-GAN} takes $n(\mI_{E})$ as both generator and discriminators input. In addition, we also compare with the model trained with the original edge maps $\mI_{E}$ only, i.e., without applying the proposed edge map noising methods, which is denoted with \textbf{O-Edge}.

For the test split, we apply the different noising methods $n(\cdot)$ to the edge maps first, and then feed the same set of noisy edge maps into the models during inference. The average of the Structural Similarity Index (SSIM) \citep{wang2004image} between the original image and the synthesised images are computed to compare their robustness to the noises. 

As shown in Table~\ref{tb:ablation_edge}, \textbf{AC-GAN} consistently outperforms \textbf{O-Edge} and \textbf{C-GAN} in all evaluation scenarios, which effectively helps improve model robustness to the common edge map noises introduced by hand editing and edge map extraction models, including random remove, random shift and random lines. This again demonstrates the effectiveness of our proposed method in various editing tasks. 

\begin{table}[ht!]
\centering
\caption{Impact of the noisy edge maps on the model trained with the  conditional GAN and the proposed asymmetric conditional GAN. We use SSIM to measure the similarity between real and fake images. \textbf{O}, \textbf{RR}, \textbf{RS} and \textbf{RL} represent original edge map, random removal, random shift and random lines, respectively.}
\begin{tabular}{lcccc}
\toprule
 & \textbf{O} & \textbf{RR} &  \textbf{RS} &  \textbf{RL} \\
\midrule
\textbf{O-Edge} & 0.5930 & 0.5751 & 0.5895 & 0.5875 \\
\textbf{C-GAN} & 0.5974 & 0.5857 & 0.5939 & 0.5939 \\
\textbf{AC-GAN} & \textbf{0.6006} & \textbf{0.5896} & \textbf{0.5973} & \textbf{0.5974} \\
\bottomrule
\end{tabular}
\label{tb:ablation_edge}
\end{table}

\subsubsection{Region-Weighted Discriminators}
We also compare the quality of the generated images with and without the region-weighted discriminator $D$ introduced in \S\ref{sec:discriminators}. We use \textbf{MD} to denote the model trained with the original multi-scale discriminator ($D_G$ only), and use \textbf{RW-MD} to denote the proposed model that use region-weighted multi-scale discriminators ($D_G$ and $D_L$). Same as above, the training set is used to train \textbf{MD} and \textbf{RW-MD}, and the test set is used for evaluation.

\begin{figure}[htbp!]
\centering
{\footnotesize
\includegraphics[width=0.45\columnwidth]{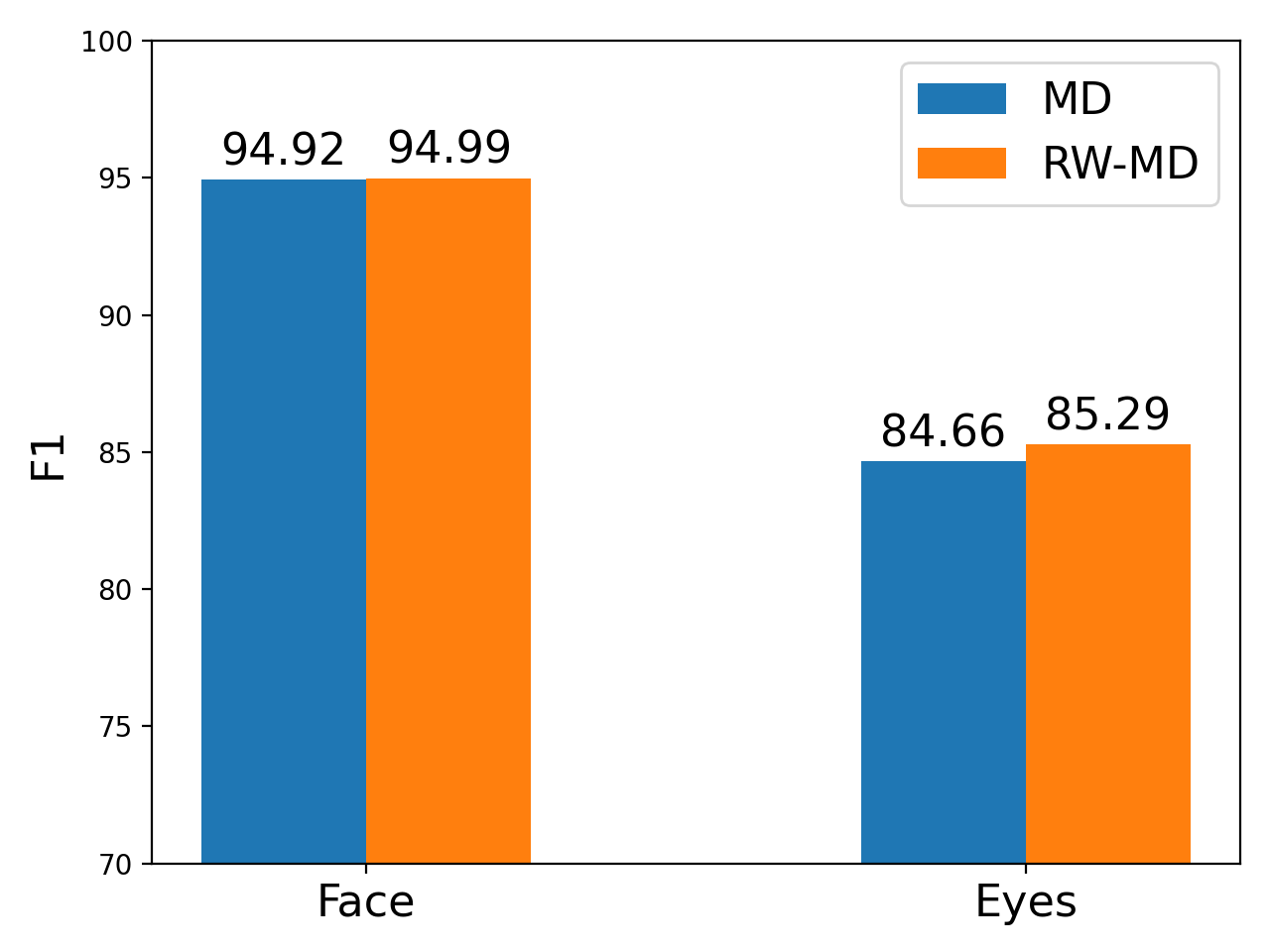}
}
\caption{Semantic segmentation F1 of the edited images generated with MD and RW-MD.}
\label{fig:rw_md}
\end{figure}

The results are shown in Figure~\ref{fig:rw_md}. Since the face and eye masks are used to give these regions higher weight when training \textbf{RW-MD}, we use the face parsing model trained by \citep{DBLPLee0W020} to annotate the fake images by identifying the corresponding segments and classifying their labels, which are then compared with the manual annotations. The average of semantic segmentation F1 is used to quantify the reconstruction quality. From the results we can see, the average F1 scores of \textbf{RW-MD} are higher than \textbf{MD} for both face and eyes, where the improvements of eyes are more obvious.

\begin{figure}[htbp!]
\centering
{\footnotesize
\makebox[0.12\columnwidth][c]{Original}
\makebox[0.12\columnwidth]{}
\makebox[0.12\columnwidth]{}
\makebox[0.12\columnwidth]{}
\makebox[0.12\columnwidth]{}
\vspace{2pt}
\\
\includegraphics[width=0.12\columnwidth]{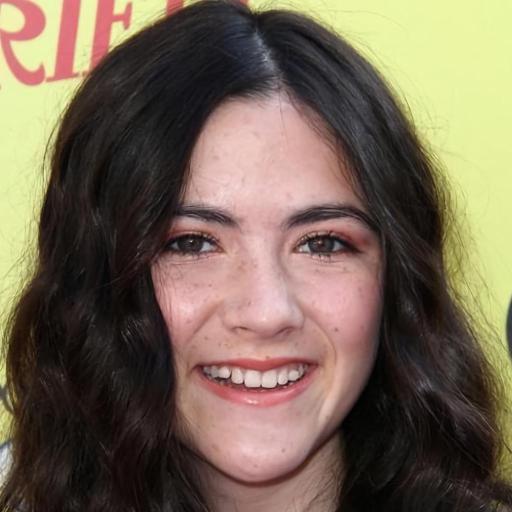}
\includegraphics[width=0.12\columnwidth]{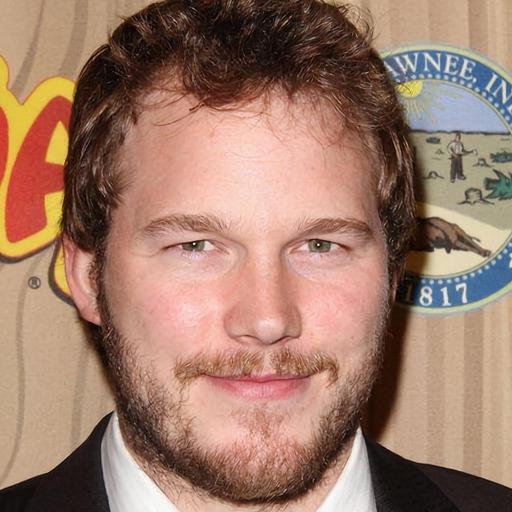}
\includegraphics[width=0.12\columnwidth]{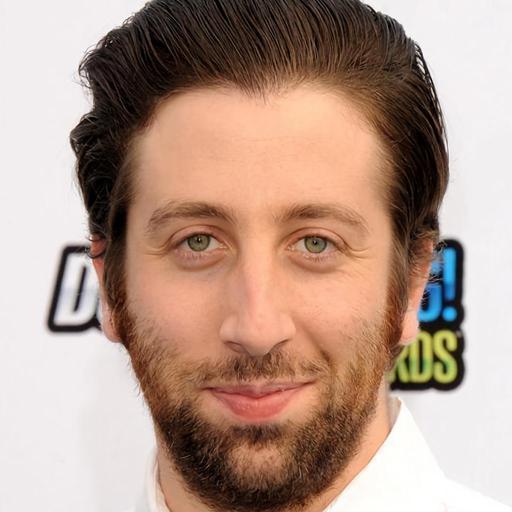}
\includegraphics[width=0.12\columnwidth]{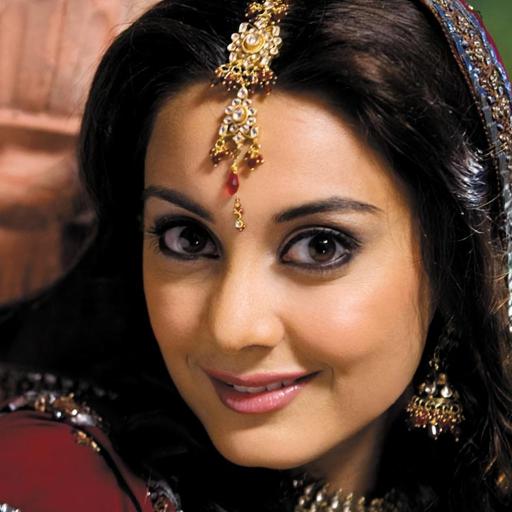}
\makebox[0.12\columnwidth]{}
\\
\makebox[0.12\columnwidth][c]{Editing}
\makebox[0.12\columnwidth]{}
\makebox[0.12\columnwidth]{}
\makebox[0.12\columnwidth]{}
\makebox[0.12\columnwidth]{}
\vspace{2pt}
\\
\includegraphics[width=0.12\columnwidth]{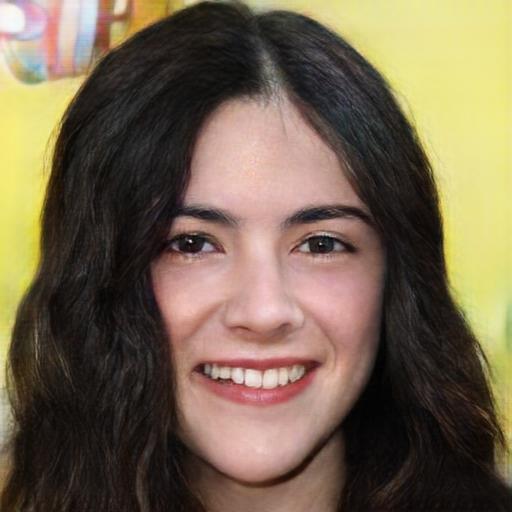}
\includegraphics[width=0.12\columnwidth]{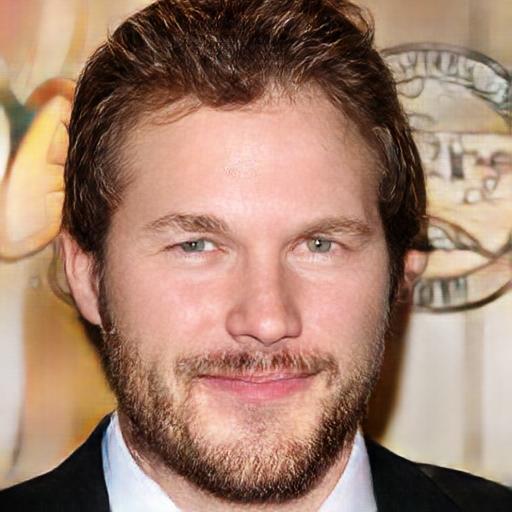}
\includegraphics[width=0.12\columnwidth]{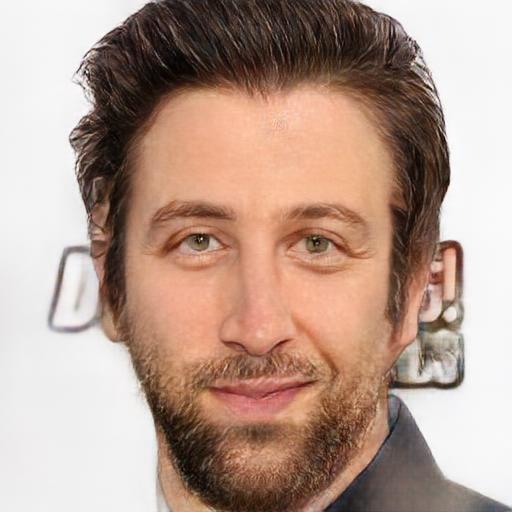}
\includegraphics[width=0.12\columnwidth]{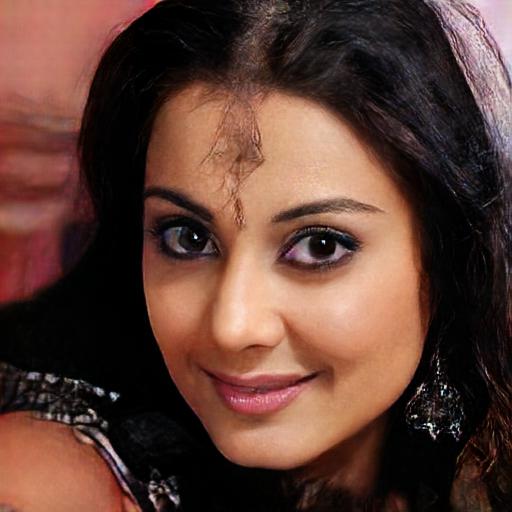}
\includegraphics[width=0.12\columnwidth]{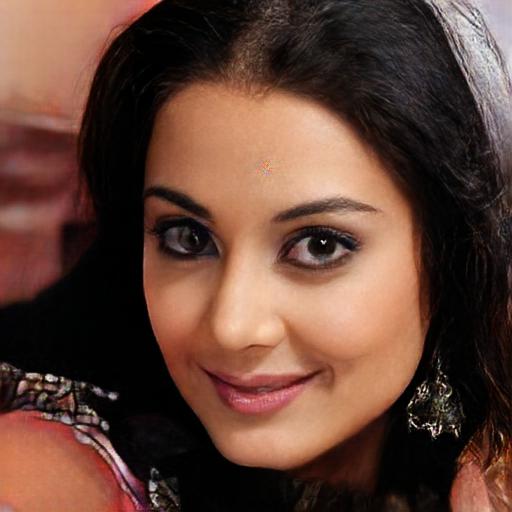}
\\
\makebox[0.12\columnwidth]{(a)}
\makebox[0.12\columnwidth]{(b)}
\makebox[0.12\columnwidth]{(c)}
\makebox[0.12\columnwidth]{(d)}
\makebox[0.12\columnwidth]{(e)}
}
\caption{Limitations. (a) Some fine details, such as freckles, are missing; (b)-(c) Background are changed. (d) missing jewelry; (e) generated image after erasing the jewelry in edge maps for (d), i.e., geometry editing.}
\label{fig:limitations}
\end{figure}

\section{Discussions \& Future Work}

We presented a new method for portrait image editing. Using a single neural network model, our method allows the user to edit colors, lights, shadows as well as geometries in an easy and intuitive manner. Our method can also generate visually pleasing images from hand-drawn sketches.

While we have demonstrated high quality editing results, our approach is still subject to a few limitations that can be addressed in follow-up work.
First, we observe our neural network model often removes small facial features, such as freckles in the results (Fig~\ref{fig:limitations}(a)), since our edge map extraction method usually ignores them. One possible solution is to provide an additional mask and annotate training data to model those small features. Second, our model focuses on foreground objects, thereby has difficulty in colorizing background patterns (Fig~\ref{fig:limitations}(b),(c)). This is because of the large variety in background patterns, which are more difficult for the model to learn. Inspired by \citet{yang2020deep}, one can add foreground masks so that the model generates portraits only, while keeping the original background unchanged. Third, our model cannot generate jewelry and headgear
that are not available in the training data. For example, the jewelry in Fig~\ref{fig:limitations}(d) is missing in the generated image. This problem can be solved by providing more relevant training data. Fourth, with the help of color palette, our method enables fine-grained color editing in batches. However, when applying our model to videos, we observe that the generated results often have flickering artifacts due to lack of consideration of temporal coherence. It is interesting to extend our method for fine-grained video color editing with video generation models. 

Though we focus on portrait image editing in the paper, we believe the proposed asymmetric conditional GAN (AC-GAN) is a general framework that can be applied to other controllable generation tasks. For example, AC-GAN can deal with the situations when the controlling signal is over simplified or abstracted for model to learn (e.g., auto-encoder encoded representation of edges), or when there is too few training data. We will explore along this direction in the near future.

\bibliography{main}

\vfill

\end{document}